\documentclass[titlepage]{amsart}
	\usepackage[margin=1.2in]{geometry}

\usepackage{graphicx}
\usepackage[english]{babel}
\usepackage{float}
\usepackage{todonotes}
\usepackage[export]{adjustbox}
\usepackage{amsmath,amssymb,latexsym,amsthm} 
\usepackage[foot]{amsaddr}

\theoremstyle{definition}
\newtheorem{teo}{Theorem}[section]

\newtheorem{alg}[teo]{Algorithm}
\newtheorem{defi}[teo]{Definition}

\newcommand{\ev}{\operatorname{ev}}

\title{Collision free motion planning on a wedge of circles}

\begin{document}
\author{Elif Sensoy}
\address[Elif Sensoy, Author]{Wright College,
 4300 N. Narragansett Avenue, Chicago, IL 60634 USA}
\email{esensoy@student.ccc.edu}

\address[Hellen Colman, Advisor]{Department of Mathematics, Wright College,
 4300 N. Narragansett Avenue, Chicago, IL 60634 USA}
\email{hcolman@ccc.edu}

\begin{abstract}
We exhibit an algorithm with continuous instructions for two robots moving without collisions on a track shaped as a wedge of three circles.
We show that the topological complexity of the configuration space associated with this problem is $3$. The topological complexity is a homotopy invariant that 
can be thought of as the minimum number of continuous instructions required to describe the movement of the robots between any initial configuration to any final one without collisions. The algorithm presented is optimal in the sense that it requires exactly $3$ continuous instructions.

\end{abstract}
\nocite{*}
\maketitle

\section{Introduction}

It is an important problem in robotics to coordinate the
movements of  vehicles  along a system of tracks without collisions. This problem was originally studied by Ghrist in \cite{Ghrist} by constructing  the configuration space $C^n(G)$ of
$n$ particles on a graph $G$,  given by
\[ C^n(G) = \{ (x_1, \ldots, x_n)\in G^n\,|\, \text{$x_i\neq
    x_j$ for $i\neq j$} \}.  \]

Every collision-free movement between two configurations of $n$ points
on the graph $G$ corresponds to a path in the configuration space $C^n(G)$. The
\emph{motion planning problem} consists of finding a function that
assigns to any pair of points in the configuration space, a path between them.

Mathematically, we can formalize this idea with the aid of the evaluation fibration. For any path-connected space $X,$ let $P(X)$ denote the space of continuous paths in $X$. The {\em evaluation fibration} $\ev: P(X)\to X\times X$  sends a path $\alpha$ to its endpoints:  $\ev(\alpha)=(\alpha(0),\alpha(1)).$  A section $s$ of this fibration takes a pair of points as input and gives a path between those points as output.  This section can be viewed as a rule

$$s: X\times X\to P(X)$$
which assigns to each pair of initial and final positions, a path between them in $X$. If $s$ is continuous, we have that any small perturbation in the initial and final positions will lead to only small variations between the assigned paths. Farber \cite{Farber1} proved that it is impossible to find a section that is continuous over the whole domain $X\times X$ if space $X$ is not contractible. 
Motivated by this, Farber introduced the topological
complexity of a space, see  \cite{survey} for a more recent survey on the subject.  The topological complexity of $X$, denoted by $TC(X)$, is a measure of this inability to find a continuous section over the whole domain.  The topological complexity of a space is invariant under homotopy.

We provide an explicit algorithm for two robots moving in a wedge of three circles $\Gamma$, which is optimal in the sense that the motion planning is performed with the minimal number of instabilities given by the topological complexity, $TC(C^2(\Gamma))$.

The organization of the paper is as follows. In the second section, we describe the statement of the motion planning problem in our case and give a throughout description of the configuration space. In the third section, we introduce the topological complexity.

Our approach consists of presenting a  comprehensive construction of the configuration space for the case of two robots moving on a wedge of circles and then building a deformation retract of this space into a simpler space that we call the {\em network}. We dedicate section 4 to this construction. We also describe here some homeomorphic spaces to the network that will make possible the calculation of the topological complexity in the next section. 

In section 5, we use the homotopy invariance to calculate the topological complexity of the configuration space, which provides us with the minimal number of instructions for our algorithm.

We exhibit an algorithm with a minimal number of instructions in section 6. This is the main section of this article. We give a detailed description of the instructions in the configuration space as well as in the physical space. We show how these instructions work in a sample case of two robots moving from a particular initial configuration to a final one.

This paper is the result of an undergraduate research project at Wilbur Wright College supervised by Professor Hellen Colman.

\section{Motion Planning Problem}
We consider two robots confined to move on a graph given by three loops joined at a point, where a collision between robots is not allowed. 
This restricted movement of the robots in tracks is quite common in applications of motion planning and can be viewed as an instance of the problem of simultaneous control of two objects avoiding collisions with each other.

Our problem involves constructing a program that allows two robots to move from their initial positions to their final ones without collisions. We want our algorithm to be robust, in the sense that small variations in the measuring of initial and final positions should lead to just small variations in those paths.

We will now define some terms that we will use throughout the paper. A {\em robot} is a mechanical system capable of moving autonomously.  The {\em physical space} is the physical environment in which the robots move.

 In our setting, the robots are two points $(A,B)$ and the physical space $\Gamma$ is a wedge of three circles, $\Gamma= \bigvee_3 S^1$.

%%%%%%%%%%%%%%%%%%%%%%%%%%%%%%

\subsection{Configuration Space}
Although the motion planning problem is defined in the physical space where the robots move, we will study another auxiliary space where it will be easier to design the motion: the configuration space. A {\em state} of the system is a specification of the positions of all robots, and the {\em configuration space} is the space of all possible states.

Motion planning problems involve navigation of the system through the configuration space in order to achieve a certain objective. The problem of finding individual paths for each robot in the physical space from initial to final positions translates into finding a single path in the configuration space from initial to final states.

\begin{figure}[h]
\caption{Physical space $\Gamma=\bigvee_3 S^1$}
\label{wedge}
\centering
\includegraphics[height=4cm]{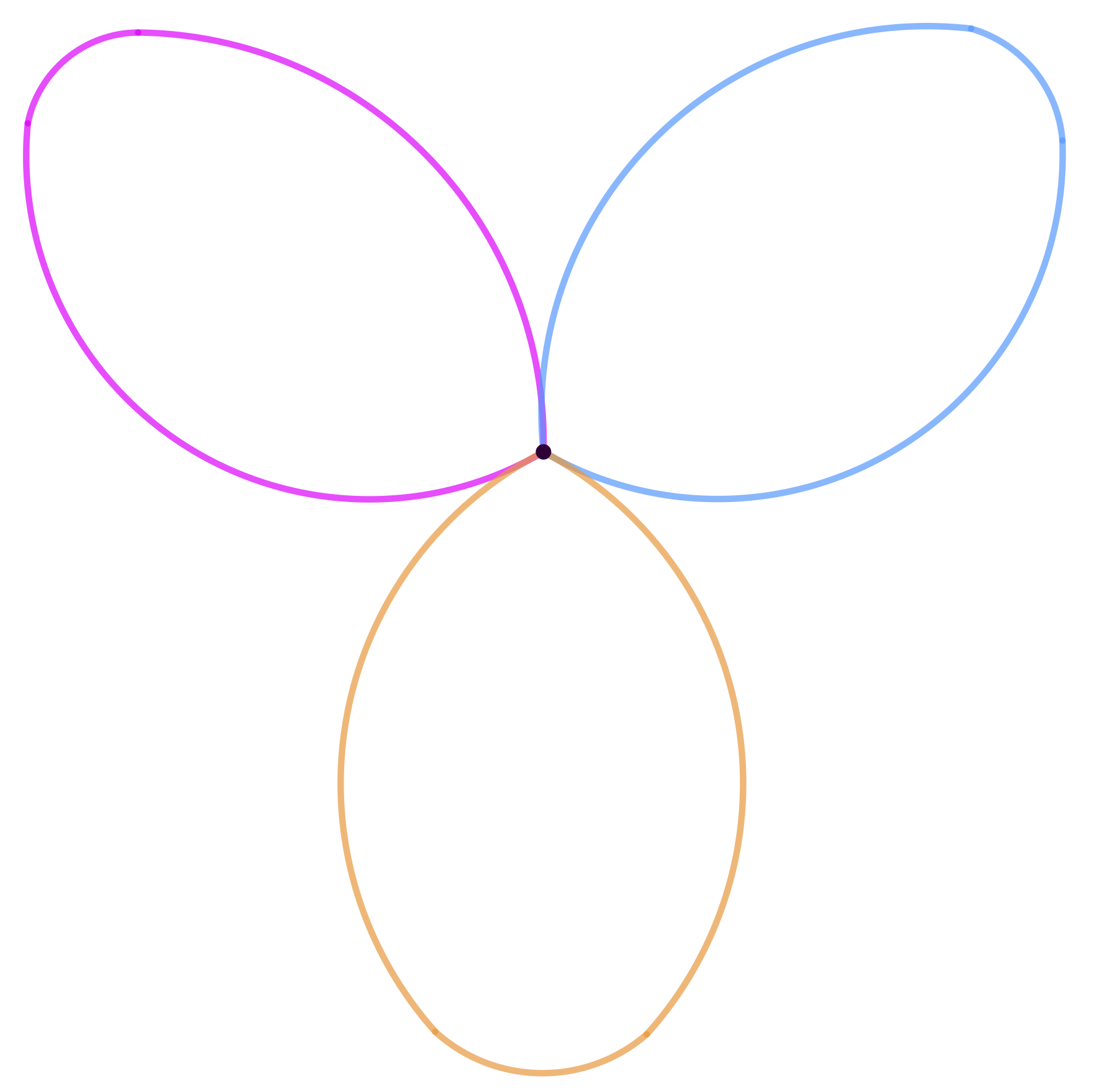}
\end{figure}

Let $(A,B)$ be two robots moving in a graph $\Gamma$ that consists of three circles joined at a point. The physical space is a wedge of circles as shown in figure \ref{wedge}.

 In order to construct the configuration space $X$ of two robots moving on a track $\Gamma$, $X=C^2(\Gamma)$, first we find the Cartesian product of  $\Gamma\times \Gamma$ of all states and then exclude the diagonal $\Delta$ of collision states. Formally, the configuration space is

$$X=C^2(\Gamma)=\Gamma\times \Gamma-\Delta=\big( \bigvee_3 S^1\times\bigvee_3 S^1\big)-\Delta$$

\begin{figure}[H]
\caption{Cartesian product $(\bigvee_3 S^1)\times(\bigvee_3 S^1)$}
\label{9tori}
\centering
\includegraphics[height=6cm]{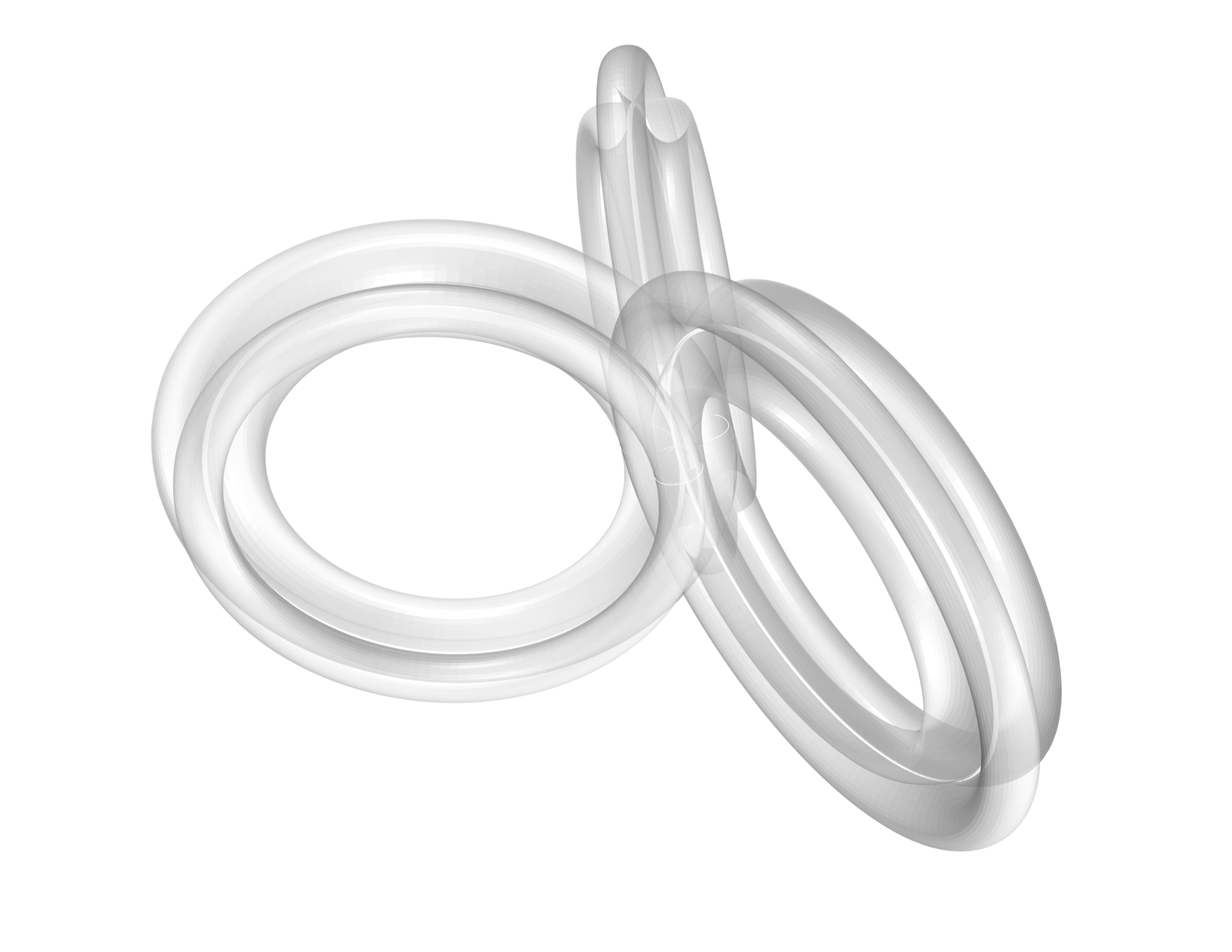}
\end{figure}

 The space $\Gamma\times \Gamma$ consists of three groups of three tori joined by a wedge of three circles. The circles in this wedge are meridians of the three tori in each group, and each group is a union of three tori with a common parallel circle. See figure \ref{9tori}.

The product $\Gamma\times \Gamma$ consists of nine tori in total and is considerably complicated to visualize since even though the dimension of the space is two, it is not possible to embed it in $R^3$. 

To remove the diagonal, we take into consideration that the collision points are the states where the two robots are in the same circle in  $\Gamma$: only one of the tori in each group would contain the entire diagonal circle. Once the diagonal is removed, each of these three tori will be cut open across its diagonal circle and homeomorphic to a cylinder. The other two tori in each group will be missing one point after the diagonal is removed.

\begin{figure}[H]
\caption{Interval to circle and square to torus}
\label{circletorus}
\centering
\includegraphics[height=6cm]{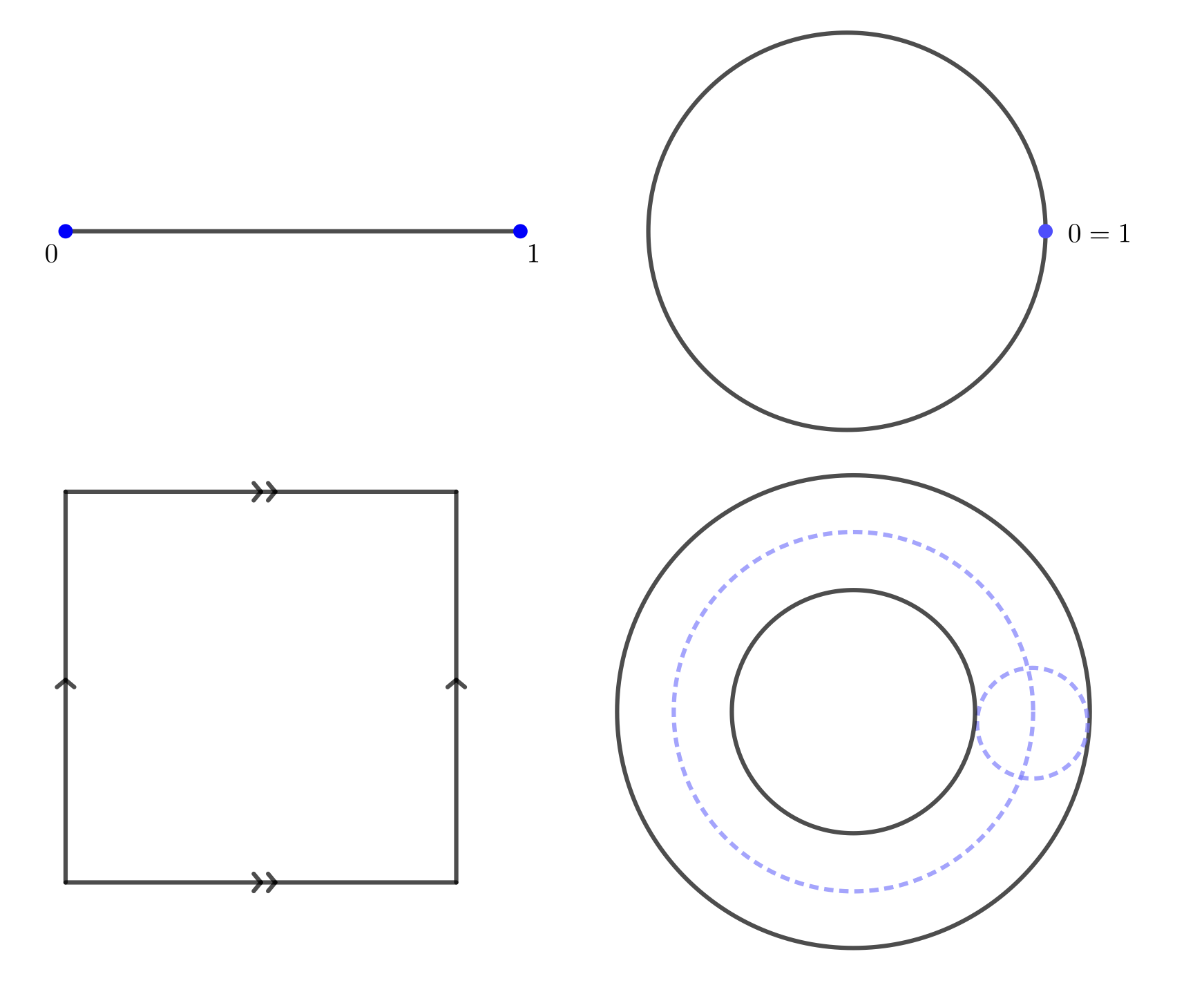}
\end{figure}

The configuration space is then composed of six tori and three cylinders all joined in an intricate way. Our approach will be to construct a flat representation of this space that will allow us to manipulate the space and understand it better. In this representation, each of the circles in the wedge  are considered segments with the extreme points identified and each torus corresponds to a square with the edges identified as in the figure \ref{circletorus}.

\begin{figure}[H]
\caption{Flat Representation of  $\Gamma\times \Gamma$}
\label{flat}
\centering
\includegraphics[height=5cm]{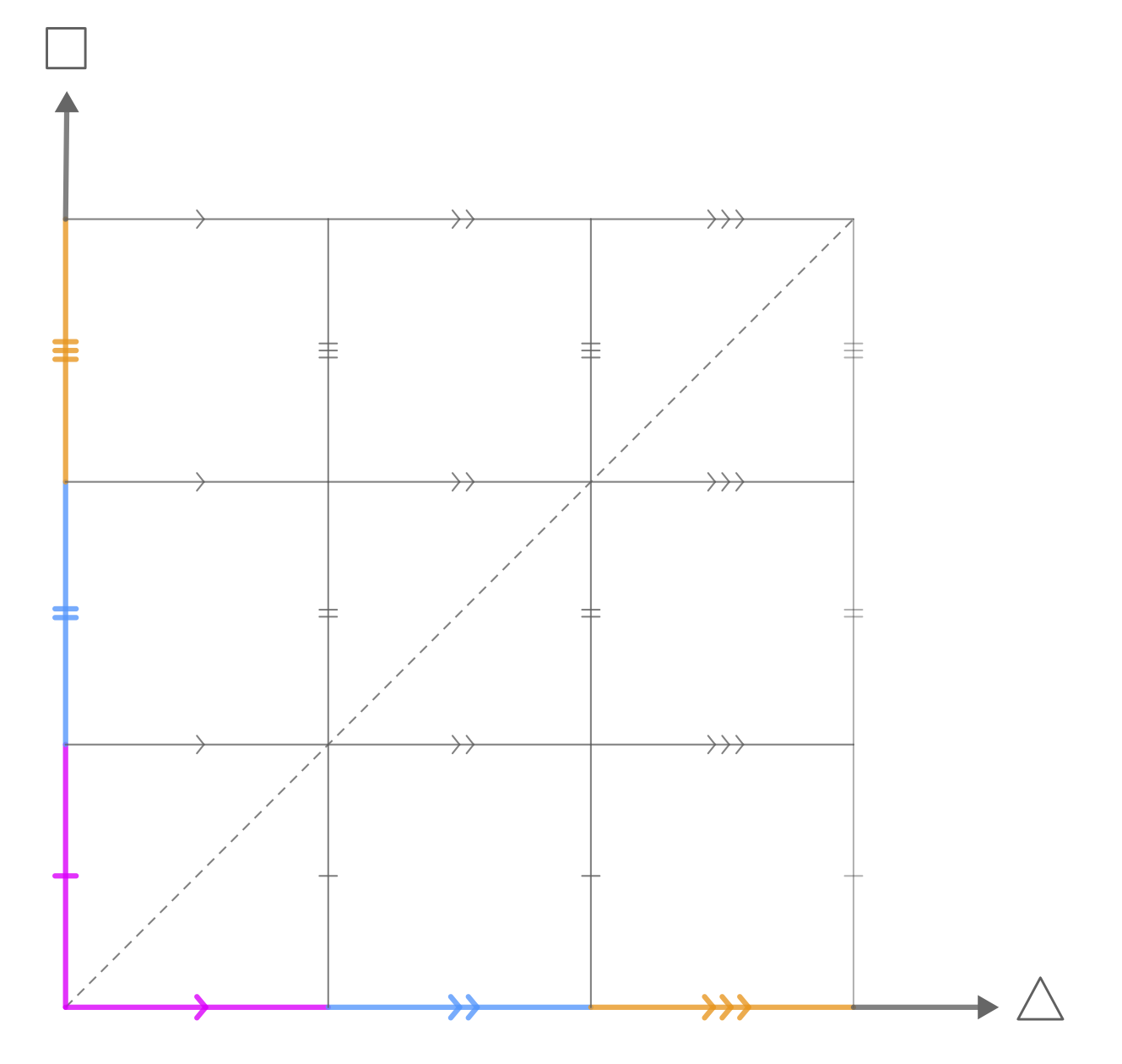}
\end{figure}

 To track the robots in this representation, we will use the Cartesian system of coordinates, where the $x$-axis represents the positions of the first robot $A$ and the $y$-axis represents the positions of the second robot $B$. Each state represents the combined positions of the two robots $(A,B)$ in $\Gamma\times \Gamma$. Alternatively, we will also use a triangle shape $\triangle$ for the robot $A$ and a square shape $\square$ for the robot $B$. Figure \ref{flat} depicts the flat representation of  $\Gamma\times \Gamma$.
 
\begin{figure}[H]
\caption{Flat representation of $\Gamma\times \Gamma-\Delta$}
\label{parallelogram}
\centering
\includegraphics[height=5cm]{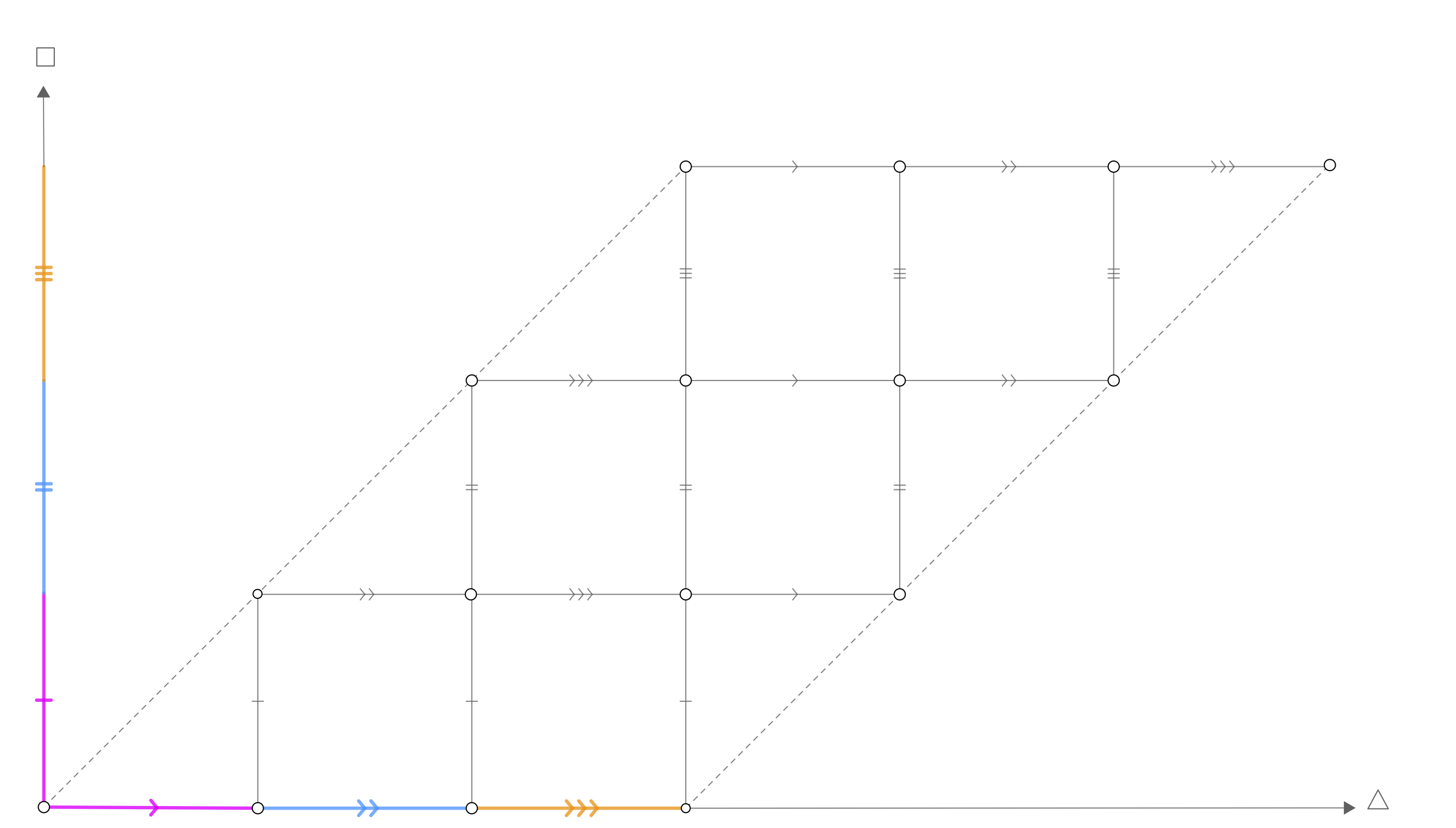}
\end{figure}

The next step is to subtract the diagonal $\Delta$ from the Cartesian product of $\Gamma\times \Gamma$ to obtain the configuration space $X=\Gamma\times \Gamma-\Delta$. Our configuration space $X$ consists of six squares with vertices removed and with the edges identified;  and three pairs of triangles, each pair with one edge identified. The squares correspond to those configurations where the two robots are on different circles in $\Gamma$, and the triangles correspond to the positions of robots in the same circle. Therefore, the configuration space $X$ consists of  six squares and six triangles with the identifications are shown in figure \ref{parallelogram}. 

\section{Topological Complexity}
The notion of  topological complexity of the configuration space $X$, $TC(X)$, was introduced by Farber in \cite{Farber1}. If $X$ is the configuration space of robots moving in a physical space $\Gamma$, then the topological complexity is an invariant that measures the navigational complexity of robots moving in $\Gamma$. This number can be defined as the minimum number of {\em continuous rules} required to describe the movement between any two points in $X$. 
A motion planning algorithm takes a pair of points as input, the initial and the final states, and returns a continuous path as output between them. In other words, a motion planning algorithm is a section $s: X \times  X \to P X$ of the evaluation map $\ev : P X \to X \times X$. That is, the section $s$ receives as input the initial and final positions of the robots $(A_i, B_i)$ and  $(A_f, B_f)$ and returns as output a path $\alpha: I\to X$ such that $\alpha(0)=(A_i, B_i)$ and $\alpha(1)=(A_f, B_f)$.

As we stated before, we would like a {\em robust} motion planning algorithm. Unfortunately, most navigation plans are discontinuous.  Farber  showed that a continuous navigation plan exists if and only if $X$ is contractible and for non-contractible spaces considered navigation plans that are continuous only when restricted to subsets of the whole Cartesian product $X\times X$. The {\em topological complexity} is the minimal $k$ for which $X \times X$ can be covered by $k$ open subsets over each of which the section is continuous. These open sets are called {\em domains of continuity}.

In the same paper, Farber proved that the topological complexity is invariant under homotopy.
\begin{teo}\cite{Farber1}\label{FarberHomotopy}
If $X$ has the same type of homotopy as $Y$, then $TC(X)=TC(Y)$.
\end{teo}
This result will allow us to find the topological complexity of our configuration space by deforming $X$ into a simpler space with the same type of homotopy and then calculate its topological complexity.

%%%%%%%%%%%%%%%%%%%%%%%%%%%%%%%%%%%%%%%%%%

\section{Homotopy of the Configuration Space}
As previously mentioned, topological complexity is the same across spaces with the same type of homotopy. Finding another space $N$ that is homotopically equivalent to $X$ will reduce the complexity of the configuration space and allow us to work on a relatively simpler space, which will be of great use not only to calculate the topological complexity but also to describe the final algorithm.

\subsection{The network N}\label{sectionHomotopy}
We construct a homotopy $$H:X\times I\rightarrow X$$ that deforms the whole space $X$ into a graph, which we call network $N$. In other words, $H_{0}(X)=X$ and $H_{1}(X)=N$, where $N$ is a graph.

\begin{figure}[H]
\caption{The network N}
\label{network}
\centering
\includegraphics[height=6cm]{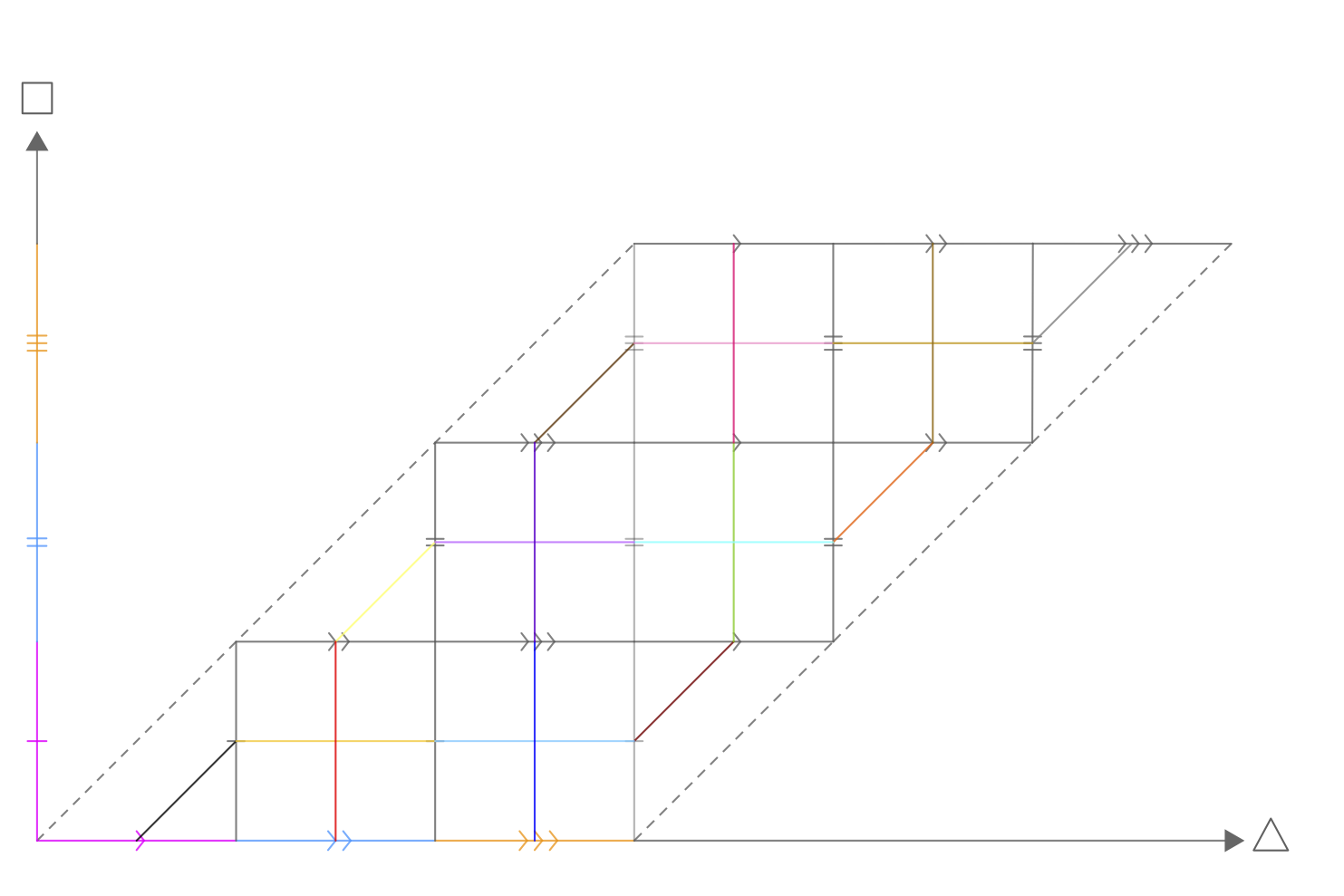}
\end{figure}

 The network $N\subset X$ consists of horizontal, vertical and diagonal segments with the identifications showed in figure \ref{network}.

 We refer to the network parts in squares as {\em cross segments} and in triangles as {\em diagonal segments}. 
 
\begin{figure}[H]
\caption{Cross segments}
\label{cross}
\centering
\includegraphics[height=3cm]{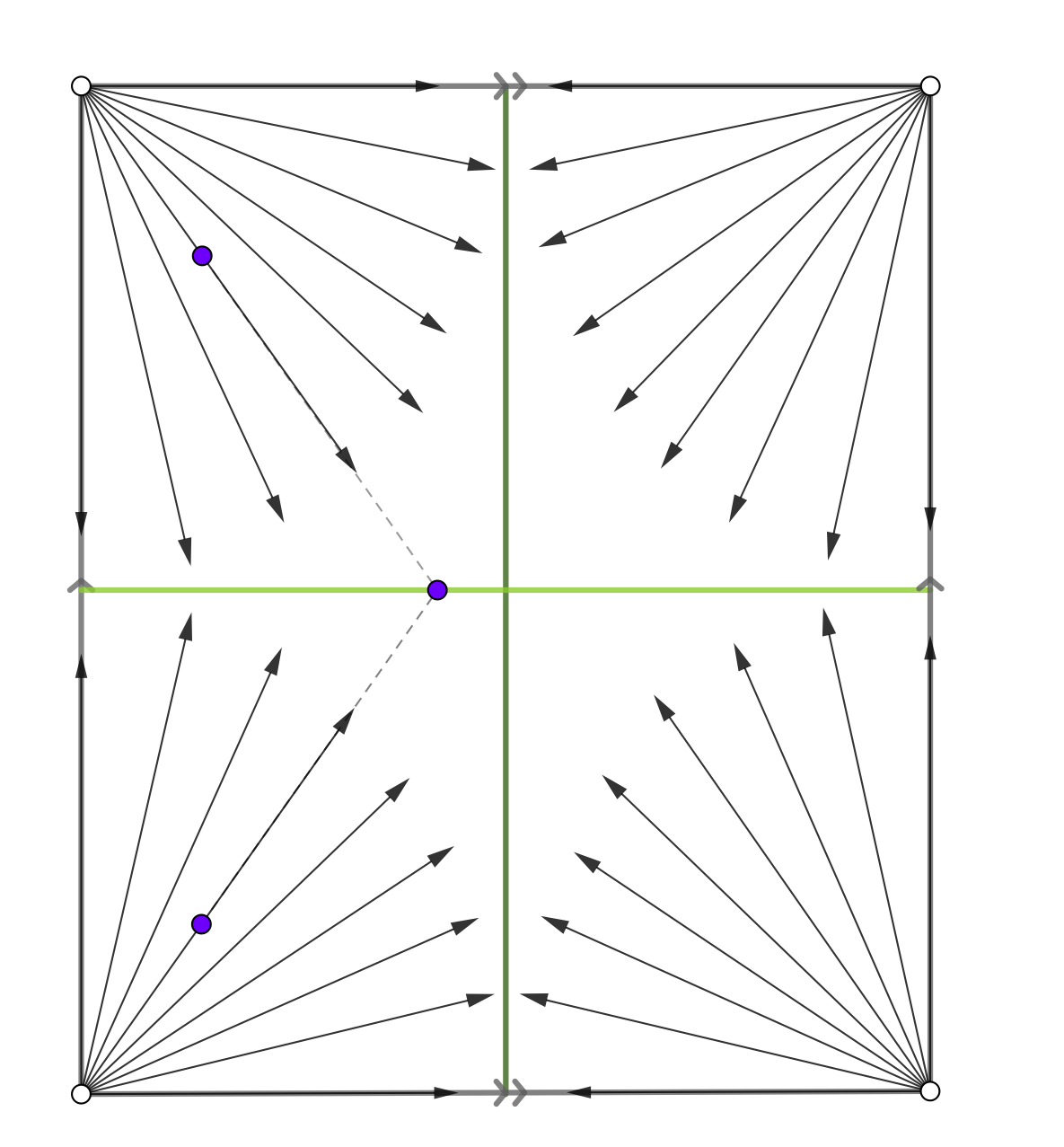}
\end{figure}

\begin{figure}[H]
\caption{Diagonal segments}
\label{diagonal}
\centering
\includegraphics[height=4cm]{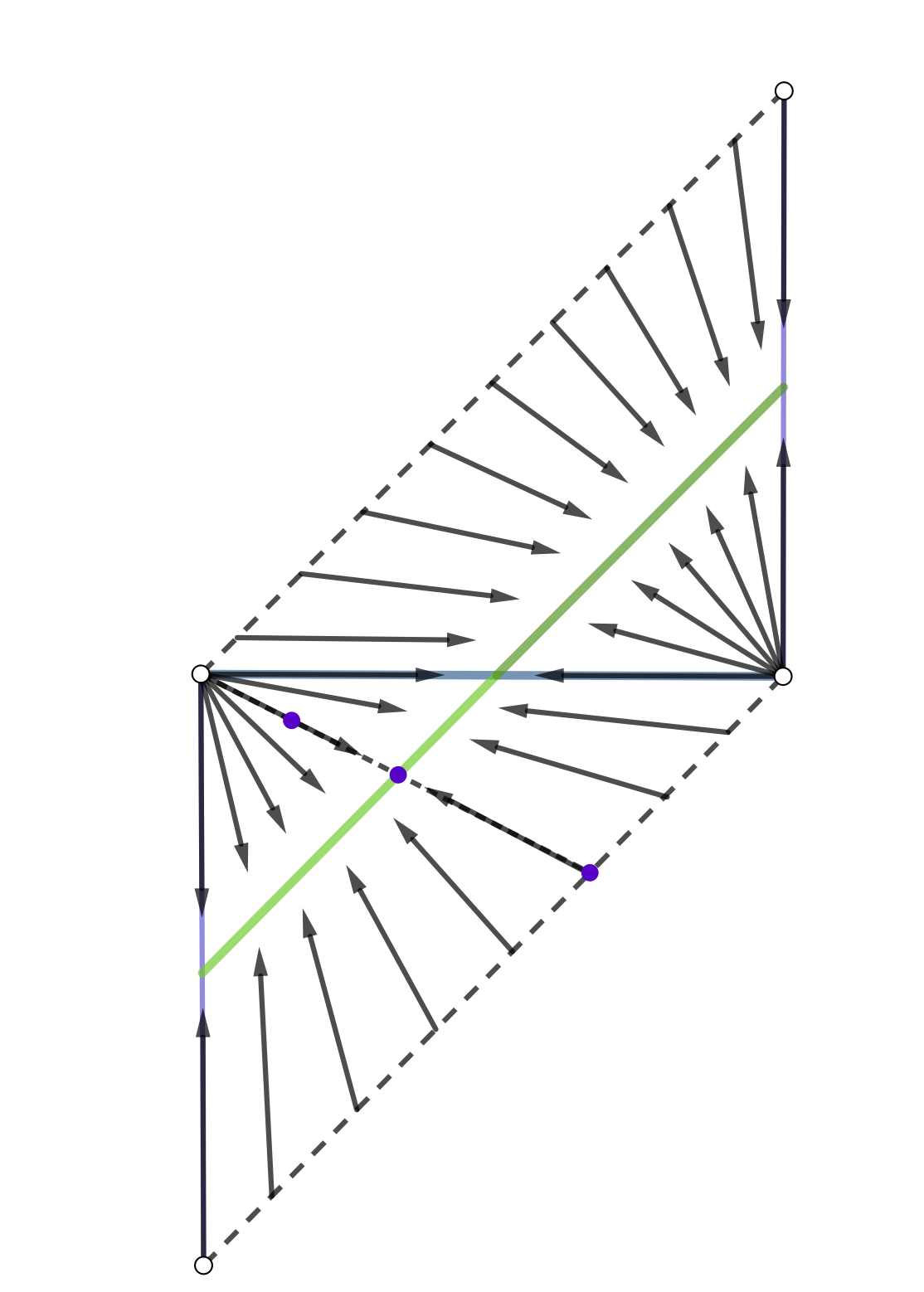}
\end{figure}

 The deformation of squares into cross segments is illustrated in  figure \ref{cross} and the triangles into diagonal segments in figure \ref{diagonal}. 

\begin{figure}[H]
\caption{Homotopy traces in $X$}
\label{traces}
\centering
\includegraphics[height=6cm]{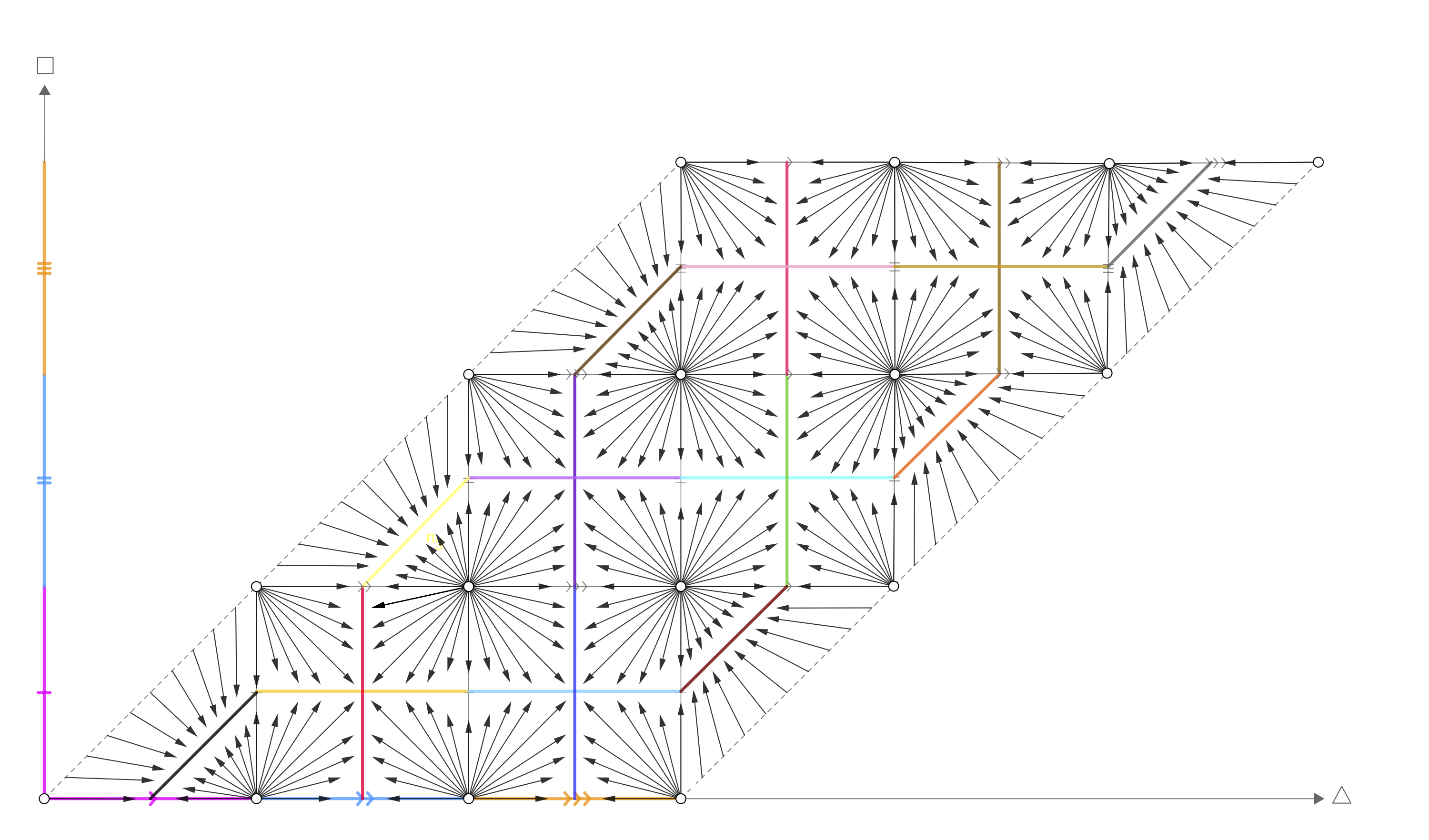}
\end{figure}

The traces $H_t(x)$ of the homotopy $H$ for all $x\in X$ are shown in figure \ref{traces}. These traces will be useful to transit in and out of the network $N$ in the configuration space $X$.

%%%%%%%%%%%%%%%%%%%%%%%%%%%

 \subsection{The chain $C$}
 Following the identification of sides in the flat representation of $X$, we can observe that the network $N$ is homeomorphic to a chain of circles. 

\begin{figure}[H]
\caption{Torus minus a point into a figure eight}
\label{figure8}
\centering
\includegraphics[height=4cm]{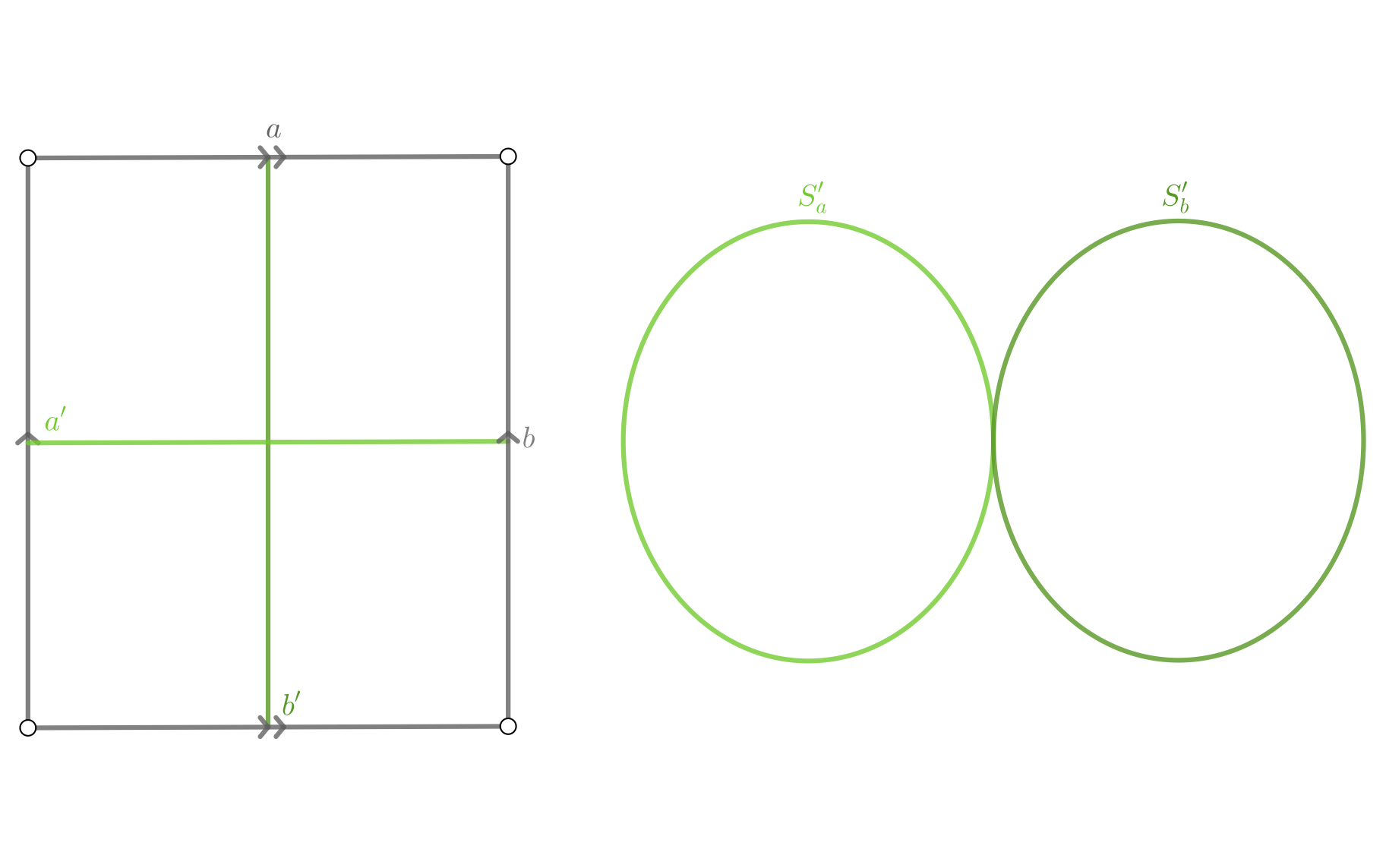}
\end{figure}

Observe that the points along the diagonal are removed from our product space, forming tori with a removed point. Each torus minus a point is deformed into the cross segments in the corresponding square, namely a figure eight. See figure \ref{figure8}.

\begin{figure}[H]
\caption{Cylinder into a circle}
\label{cylinder}
\centering
\includegraphics[height=5cm]{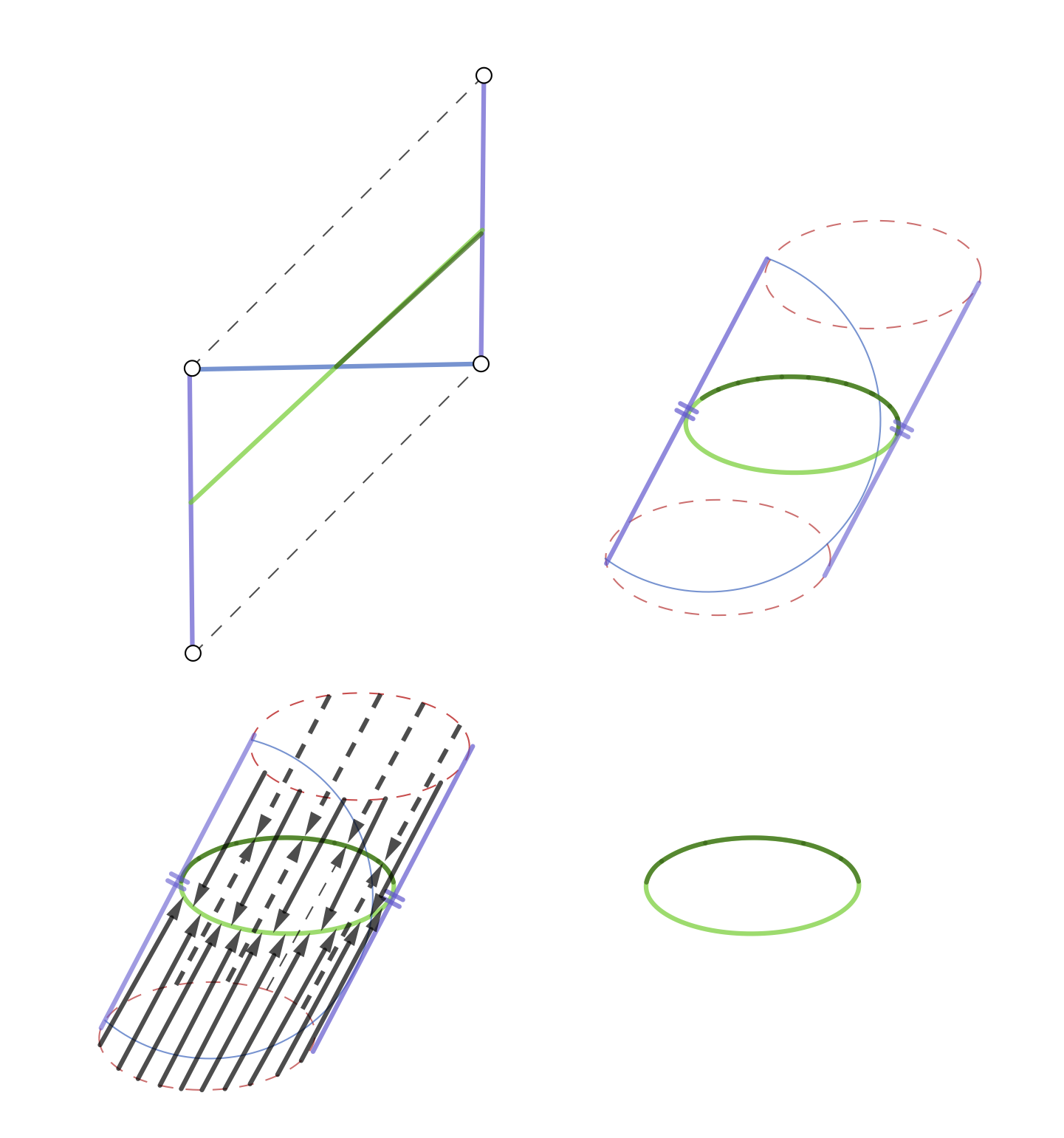}
\end{figure}

As we glue the two triangle parts along their identified edge, we obtain a parallelogram with two sides identified. Once that we glue these sides of the parallelogram together, we obtain a cylinder. Each cylinder is deformed into the diagonal segments in the corresponding triangles,  namely a circle. See  figure \ref{cylinder}.

We obtained a figure eight from each of the six squares in $X$ and a circle from each of the three parallelograms. In total, $12$ circles $+ 3$ circles $=15$ circles. We will call this union of circles the chain $C$. We have that the network $N$ is homeomorphic to the chain $C$.

\begin{figure}[H]
\caption{Labelling segments in the network $N$}
\label{names}
\centering
\includegraphics[height=6cm]{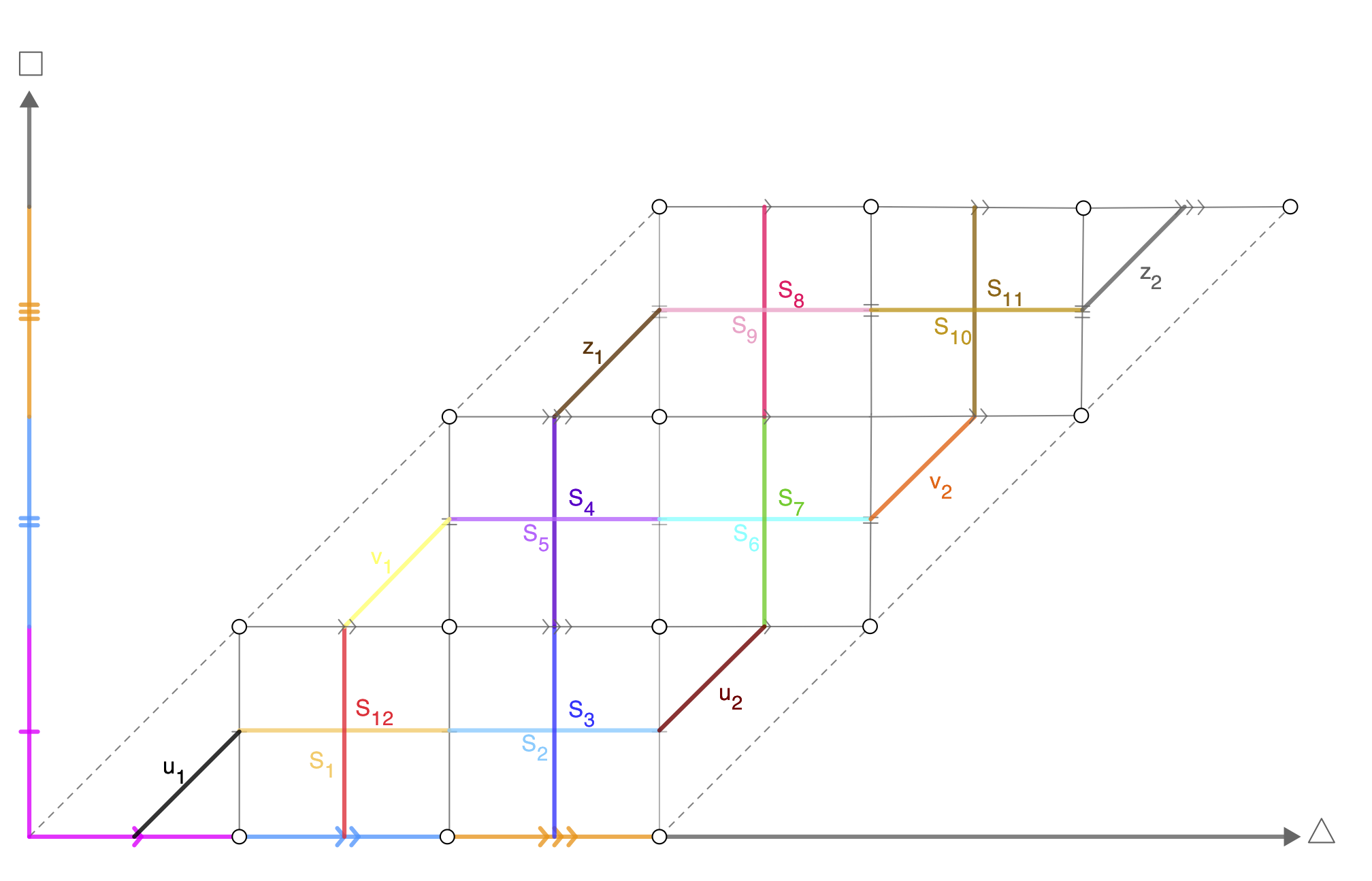}
\end{figure}

We use the following naming system for the cross and diagonal segments in $N$. The cross segments are ${S_1,S_2,...S_{12}}$ and diagonal segments are ${u_1,u_2,v_1,v_1,v_2,z_1,z_2}$. See figure \ref{names}.

\begin{figure}[H]
\caption{Chain of $15$ circles}
\label{chain}
\centering
\includegraphics[height=5cm]{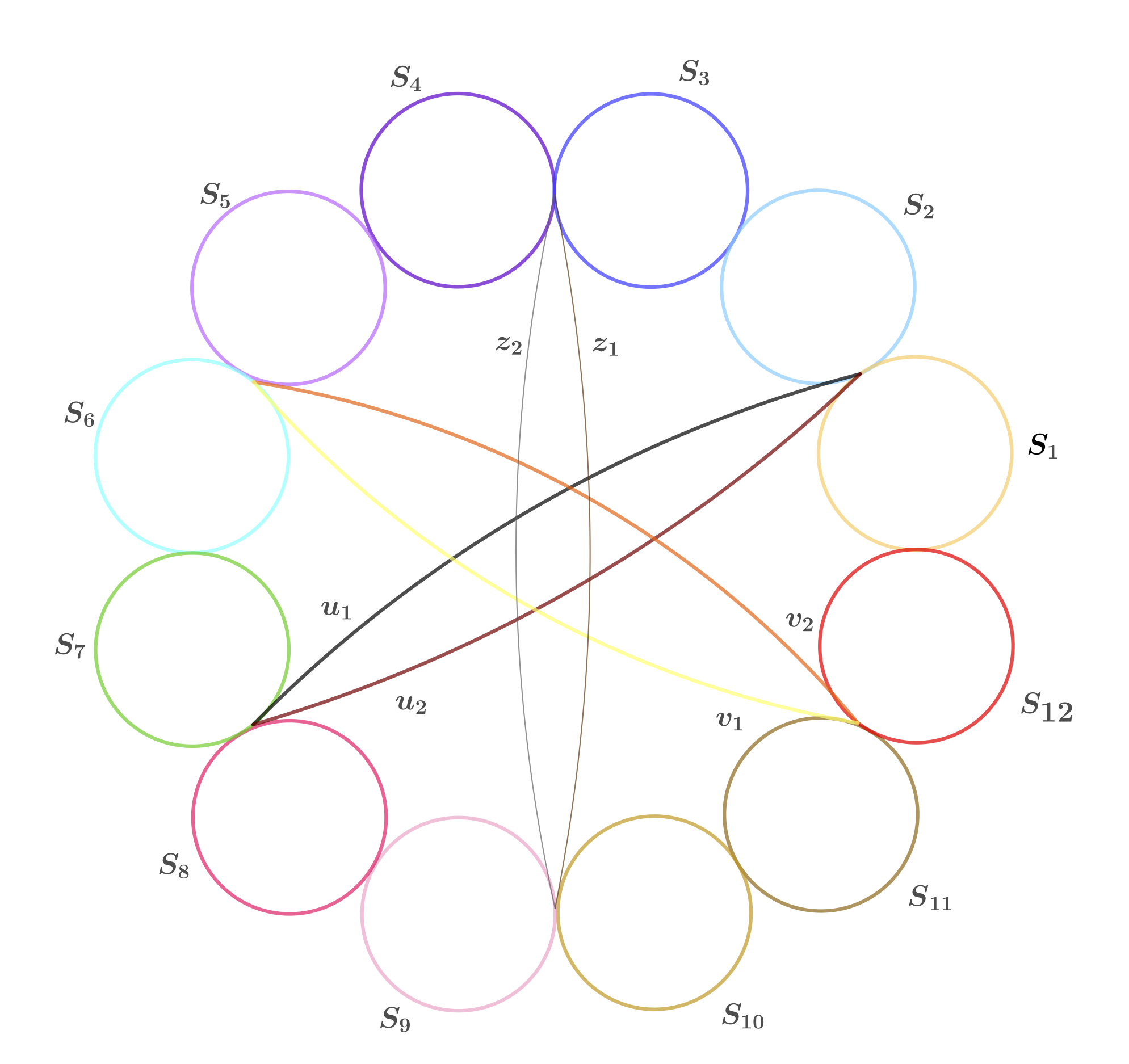}
\end{figure}

Then we draw the chain of $15$ circles accordingly, as reflected in figure \ref{chain}.

\begin{figure}[H]
\caption{Exterior semicircles in the chain $C$}
\label{exteriorC}
\centering
\includegraphics[height=5cm]{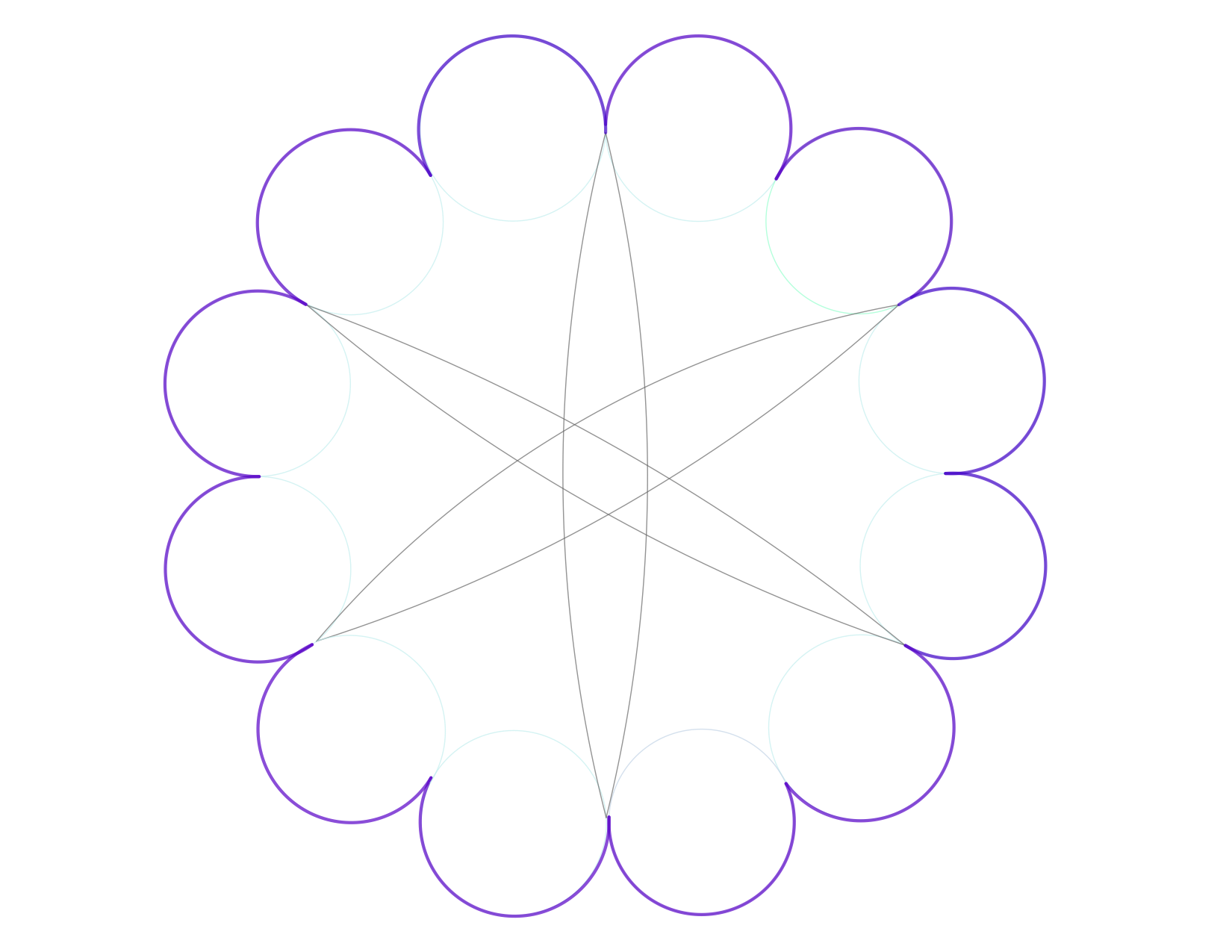}
\end{figure}

\begin{figure}[H]
\caption{Exterior semicircles in the network $N$}
\label{exteriorN}
\centering
\includegraphics[height=6cm]{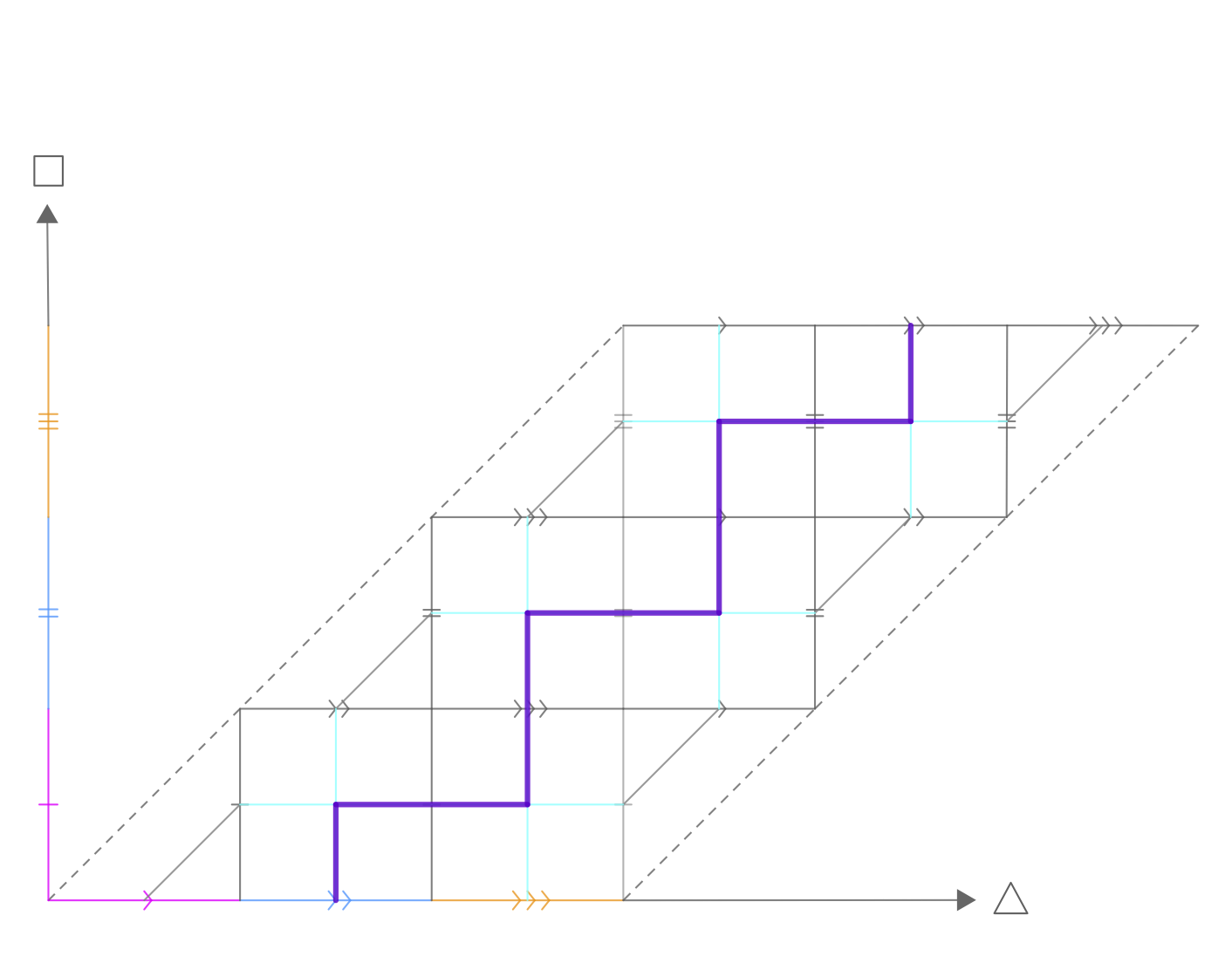}
\end{figure}

We will refer to the outer circles in the chain $C$ as {\em border circles} and inner circles as {\em connecting circles}. The outermost semicircles forming the border circles will be called {\em exterior semicircles} and the innermost ones will be the {\em interior semicircles}. In the  construction of  our motion planning algorithm, these exterior semicircles will have a distinguished role. The exterior semicircles in $C$ and in $N$ are shown in figures \ref{exteriorC} and \ref{exteriorN}. 

\begin{figure}[H]
 \caption{Horizontal and vertical circles in $C$}
\label{hv}
\centering
\includegraphics[height=6cm]{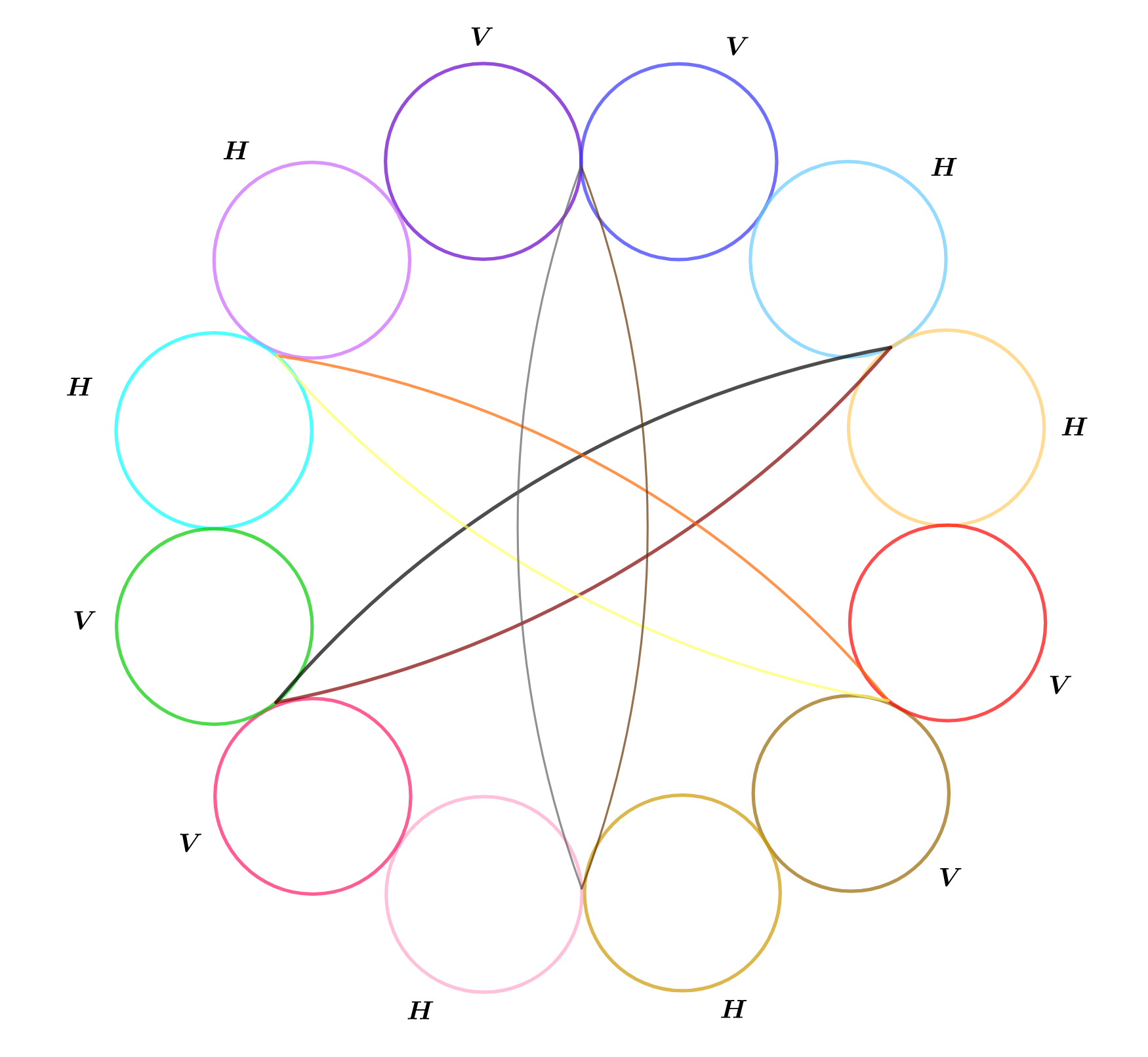}
\end{figure}

The chain circles in $C$ that correspond to vertical cross segments in the network $N$ are called {\em vertical circles}, and the circles that correspond to horizontal cross segments in $N$ are {\em horizontal circles}. Vertical circles correspond to positions in the physical space at which robot $A$ is at a pole while robot $B$ is at any place other than a pole in a different circle. Similarly, horizontal circles are the positions in $\Gamma$ at which robot $B$ is at a pole while robot $A$ is at a position other than a pole in a different circle. Vertical and horizontal circles in $C$ are shown in figure \ref{hv}. 

%%%%%%%%%%%%%%%%%%%%%%%%%%%%%%%%%%%%

\subsection{The bouquet of circles $B$}
\begin{figure}[H]
\caption{A Circle into a Teardrop}
\label{drop}
\centering
\includegraphics[height=5cm]{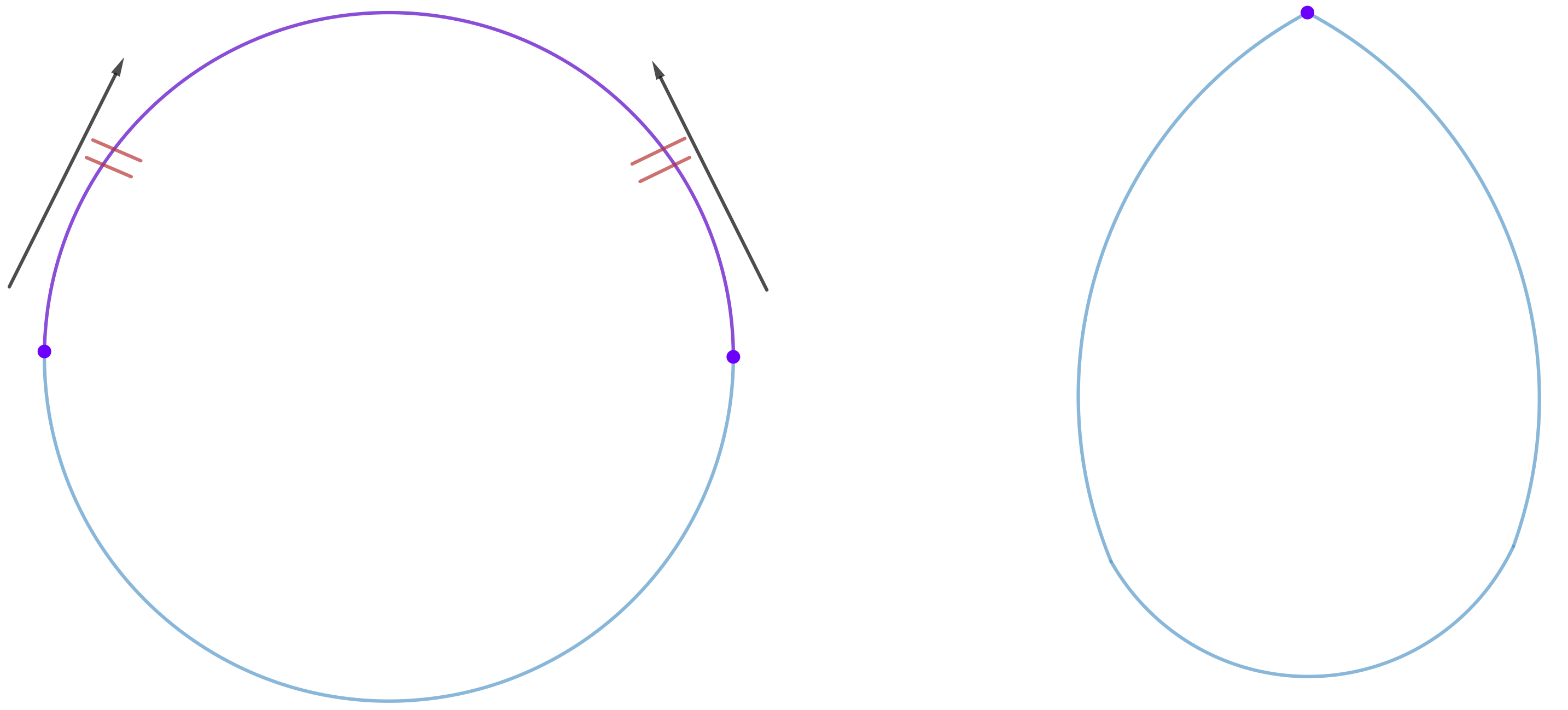}
\end{figure}

Now, we will find a homotopy equivalent space to the chain $C$. We see that each circle in the chain has two distinguished points that connect it to other circles. When we compress an interior semicircle into these points through homotopy, the interior semicircle contracts to a point, and the exterior semicircle forms a teardrop shape, as shown in the figure \ref{drop}. 

\begin{figure}[H]
\caption{Border circles into 13 teardrops}
\label{drop12}
\includegraphics[height=3.5cm]{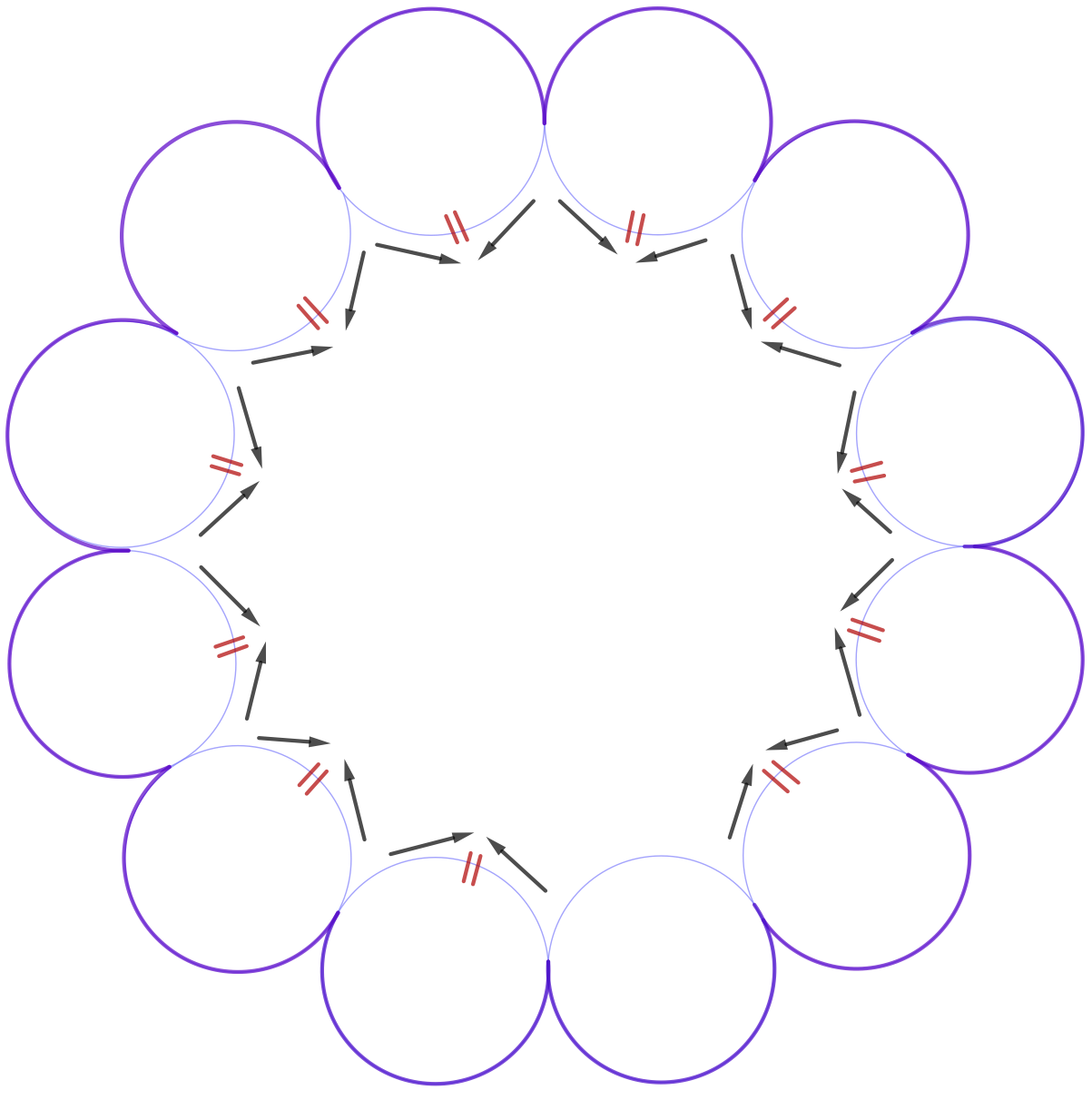}
\includegraphics[height=3.5cm]{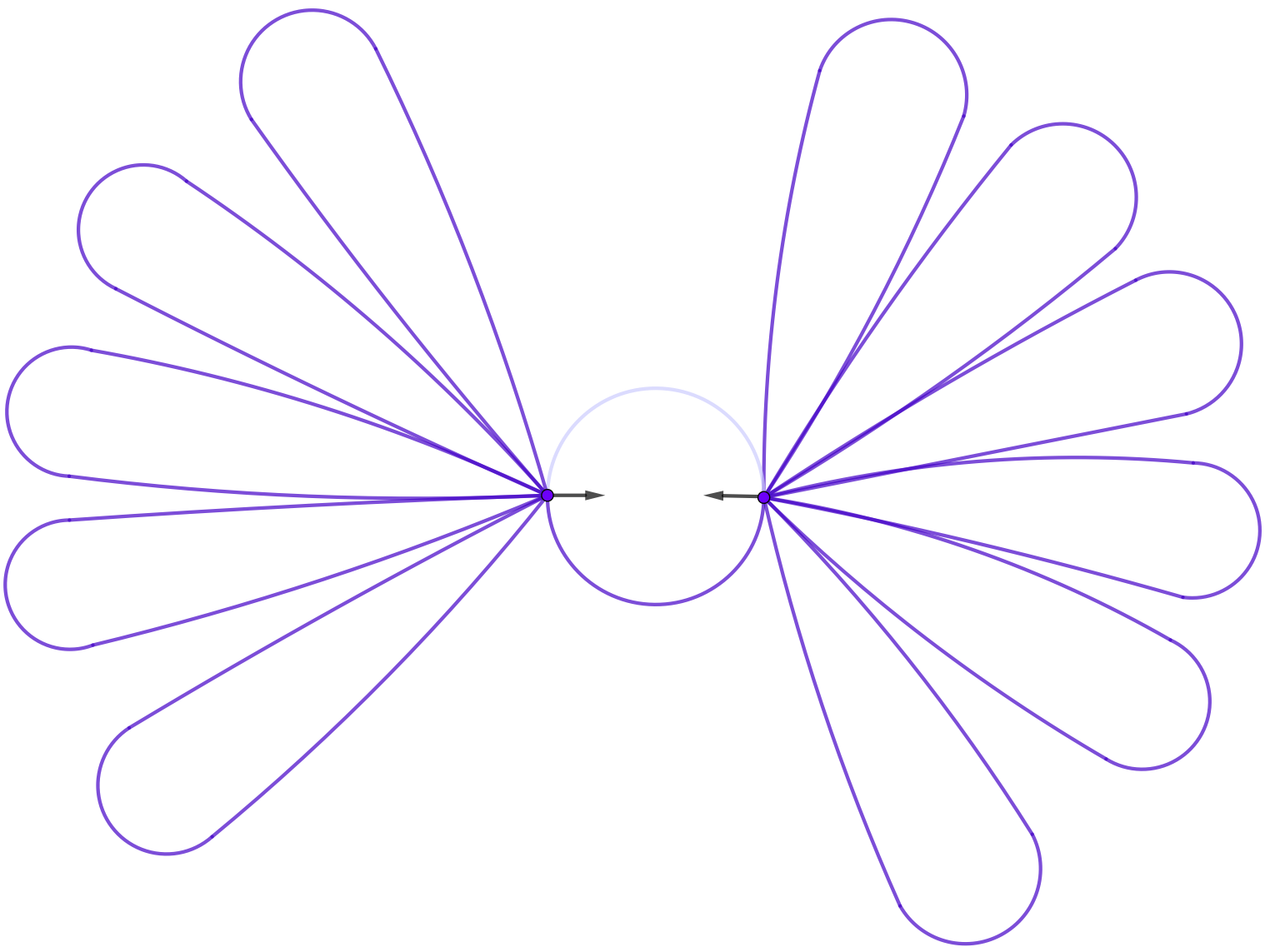}
\includegraphics[height=3.5cm]{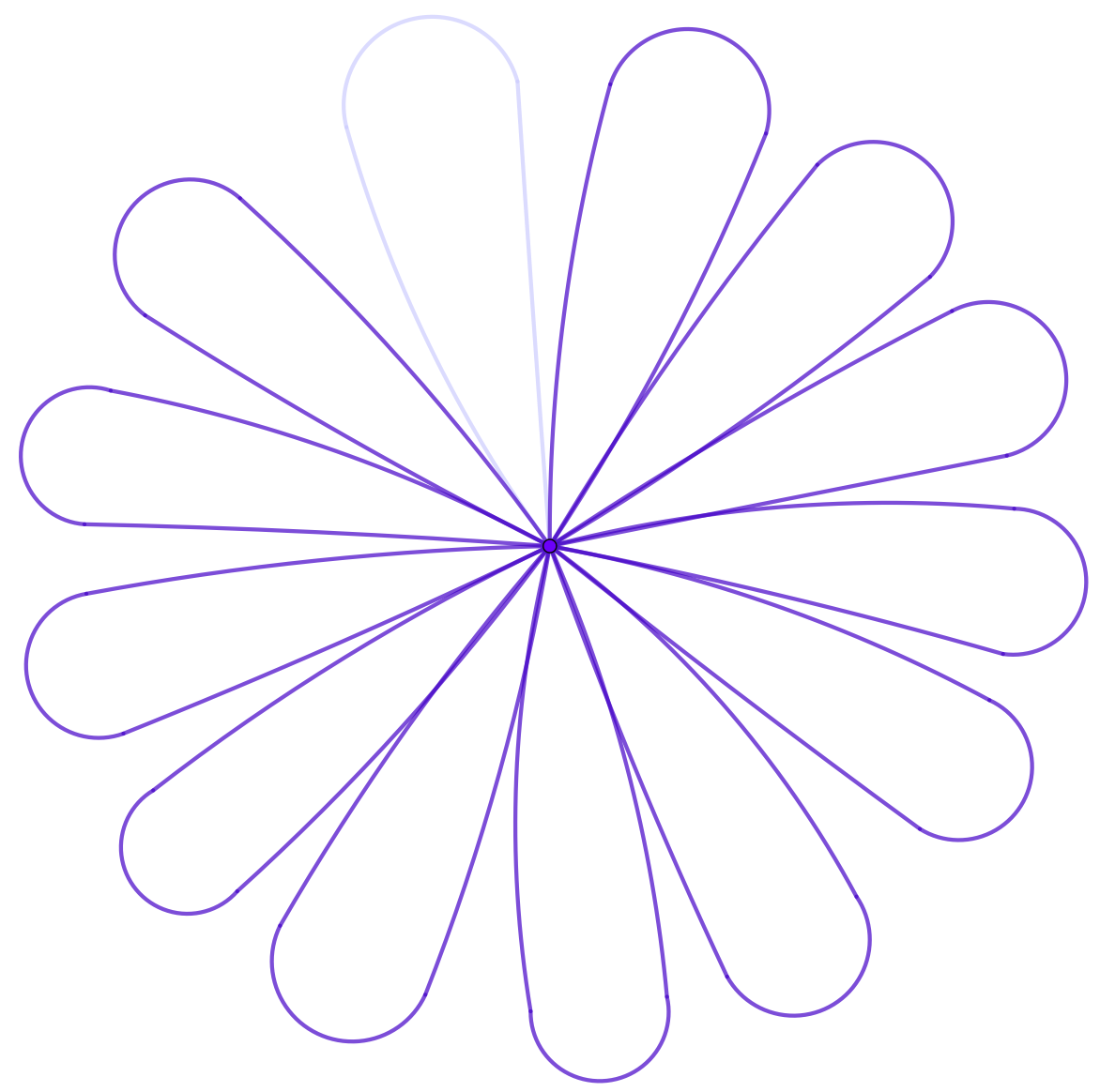}
\end{figure}

As we apply the operation to the entire chain $C$, all interior semicircles except one will collapse into a point. The process will follow as these $11$ interior semicircles collapse one by one forming eleven teardrops until they reach the last circle. At the final step, the two distinguished points will collapse into one forming two more teardrops. Hence, the total is $13$ teardrops. We can observe this occurring in the figure \ref{drop12}.

As stated before, the  connecting circles in the chain $C$ consist of two different semicircles joined by their endpoints. When we identified these endpoints, each connecting circle forms two teardrops.

\begin{figure}[H]
\caption{Bouquet of $19$  Circles}
\label{bouquet}
\centering
\includegraphics[height=6cm]{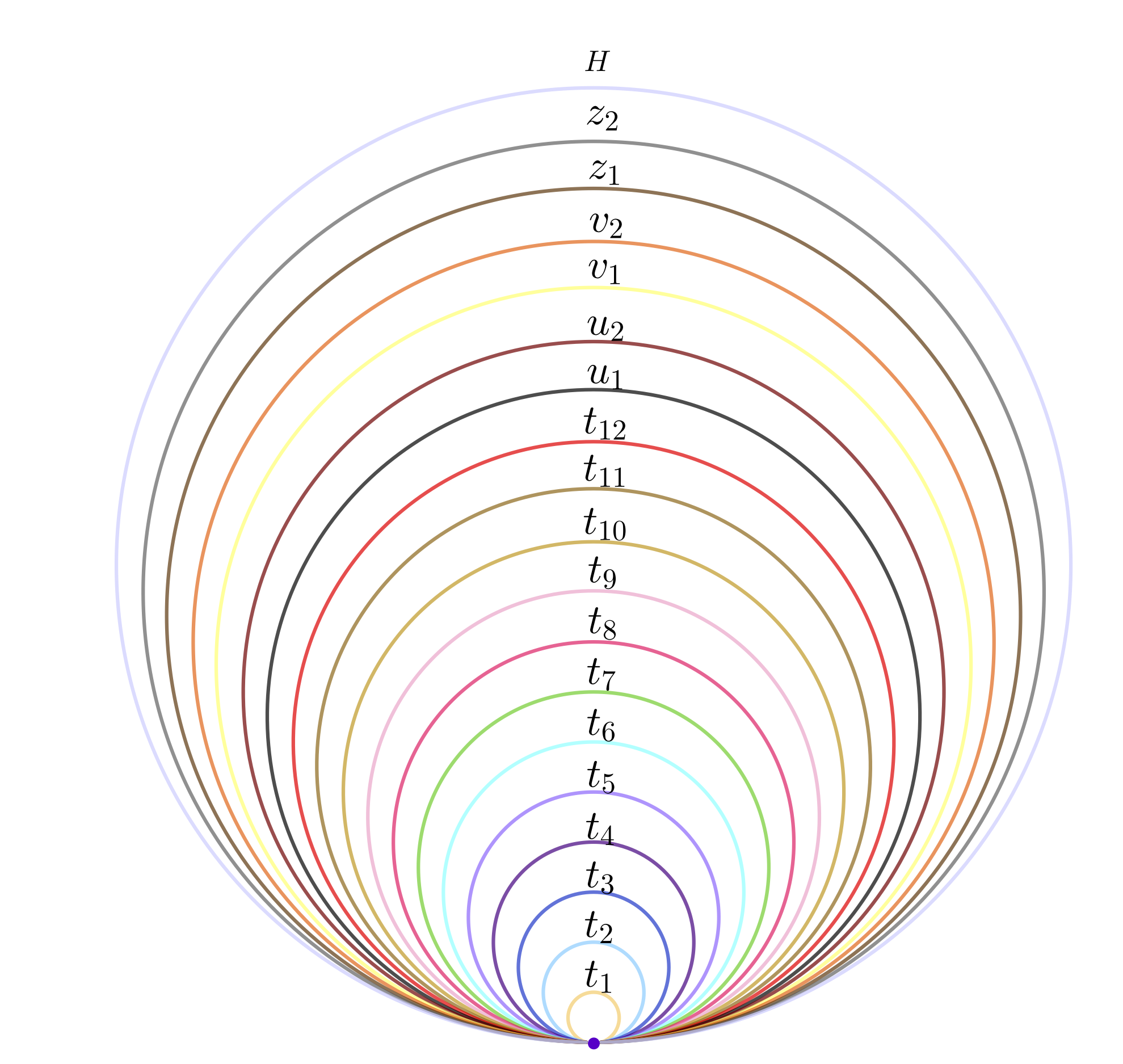}
\end{figure}

Therefore, we have $6$ teardrops from connecting circles and $13$ from border circles, forming a total of $19$ teardrops joined at one point, namely, a bouquet of $19$ circles, $B=\bigvee_{19}S^1$. See figure \ref{bouquet}. Observe that the connecting semicircles in $C$ lie on the outer rim of $B$. 

We have that the configuration space is homotopic to a bouquet of $19$ circles:
$$X\simeq N \cong C \simeq B$$
where $\simeq$ means homotopy equivalence and $\cong$ means homeomorphism.

%%%%%%%%%%%%%%%%%%%%%%%%%%%%%%%%%%%%%

\section{Topological Complexity of $X$}
In order to find the number of instructions for our motion planning algorithm, we consider  the topological complexity of the configuration space, $TC(X)$. 

Farber calculated the topological complexity of a graph $G$ based on their first Betti number. The first Betti number of a graph $G$ with $n$ vertices, $m$ edges and $k$ connected components equals $ b_1(G)=m-n+k$.

\begin{teo}\cite{Farber2}
Let $G$ be a connected graph, then 

\begin{equation}
  TC(G) =
    \begin{cases}
      1 & \text{if $b_1(G) = 0$}\\
      2 & \text{if $b_1(G) = 1$}\\
      3 & \text{if $b_1(G) \geq 2$}
    \end{cases}       
\end{equation}

Here $b_1(G)$ denotes the first Betti number of $G$. 

\end{teo}

We have shown that the configuration space $X$ is homotopy equivalent to $N$. Then we showed that the network $N$ is homeomorphic to a chain of circles $C$, which is homotopic to a wedge of $19$ circles $B$. We know that the topological complexity of $X$ will be equal to that of $B$ by theorem \ref{FarberHomotopy}. Since the first Betti number of a bouquet of nineteen circles is $b_1=19-1+1\geq 2 $, we have that $TC(B)=3$ and therefore, $TC(X)=3$.

We conclude that for two robots moving on a wedge of three circles, any motion planning algorithm will require at least three continuous instructions.

\section{Motion Planning Algorithm}
Our goal is to give an explicit description of  three continuous instructions in the physical space where the robots motion takes place. 

We will start by building the algorithm in the configuration space and then translate these instructions to the physical space by using all the constructions we have shown previously. 
Recall that any position in the physical space $\Gamma$ correspond to a state in the configuration space $X$. Our first step will be to move any state in $X$ to the corresponding state in the network $N$ following the traces of the homotopy shown in section \ref{sectionHomotopy}.

The following definitions of specific positions and states in $\Gamma$ and $X$ will help to describe our algorithm.

\begin{defi}
The {\em vertex} is the center point of the physical space at which the three circles intersect and the {\em poles} are the antipodal points to the vertex in each circle in the physical space $\Gamma$. 
\end{defi}

The states in the network $N$ correspond to positions of the robots  in $\Gamma$ where at least one robot is at a pole and the other at a different circle or both robots are at antipodal positions  in the same circle.

\begin{defi}
The {\em nodes} in the chain $C$ are the intersection points of the circles. The nodes where three circles intersect will be called {\em j-points}.
\end{defi}

\begin{figure}[H]
\caption{Cross states in $N$ and  $C$}
\label{CSN}
\centering
\includegraphics[height=5cm]{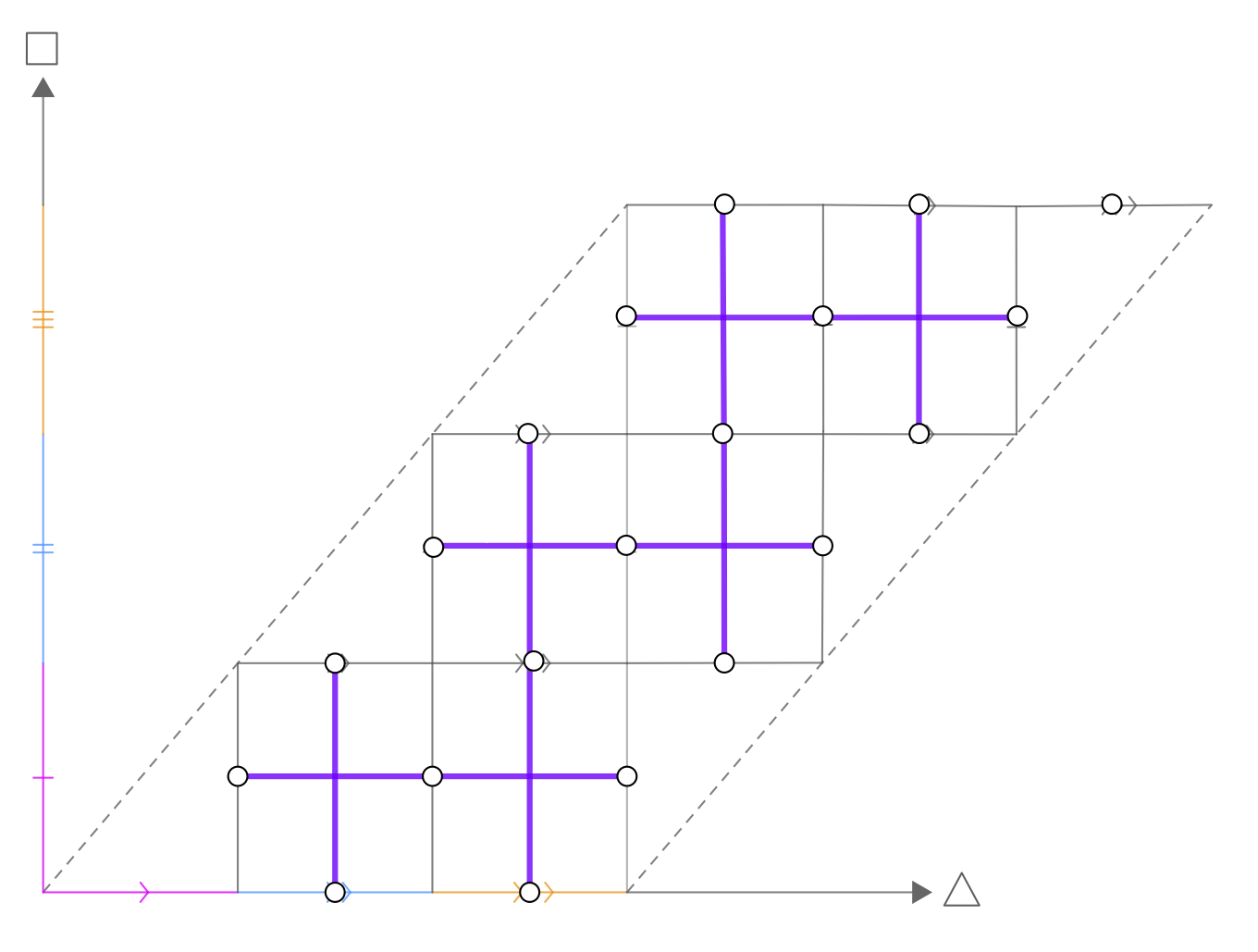}
\includegraphics[height=5cm]{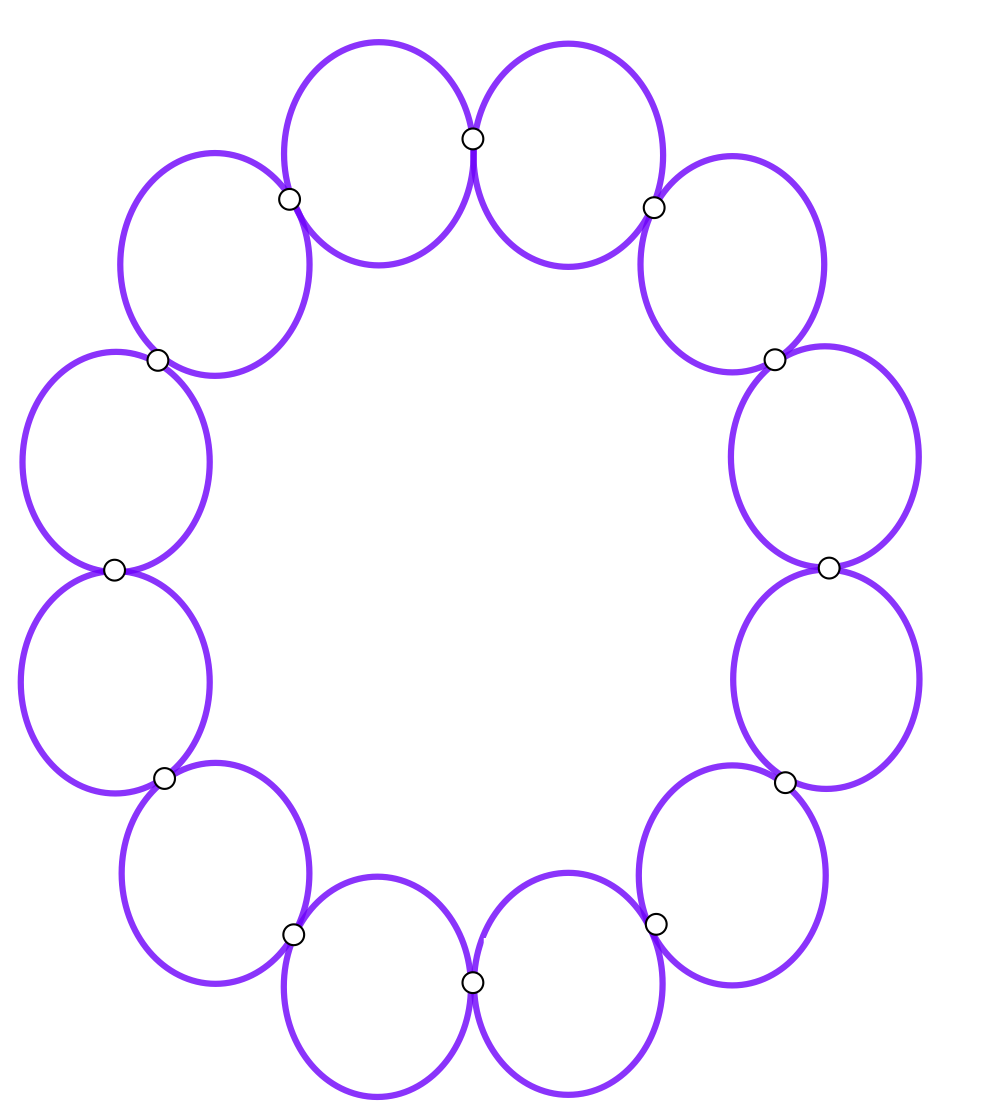}
\end{figure}

\begin{figure}[H]
\caption{An example of cross state in $\Gamma$}
\label{CSC}
\centering
\includegraphics[height=4cm]{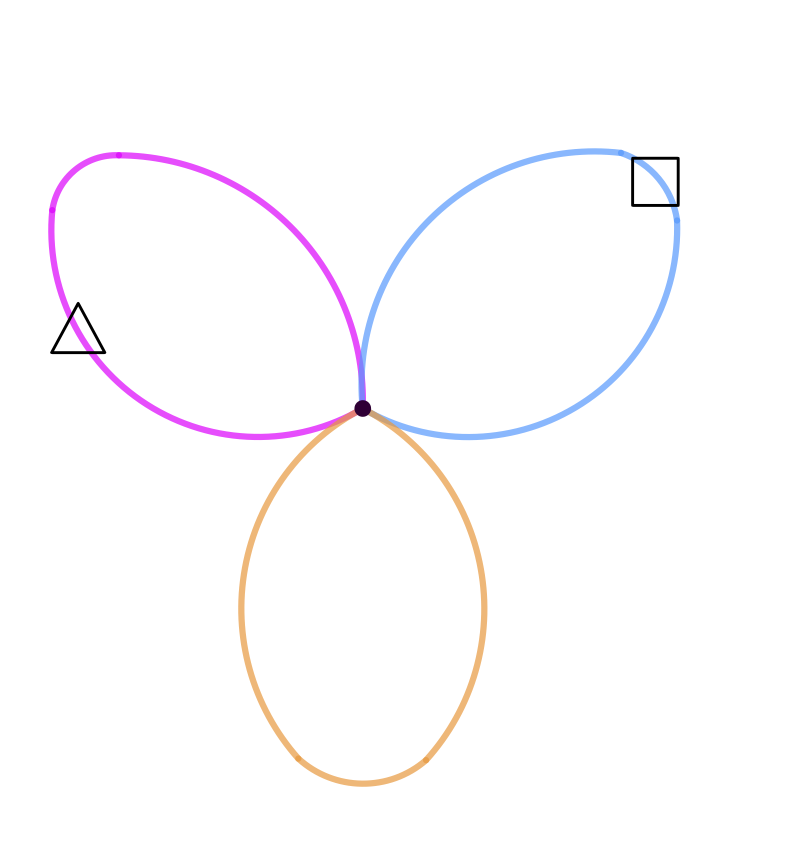}
\end{figure}

\begin{defi}
When a state is located in the interior of a cross segment in $X$, we will say it is a {\em cross state}. In the physical space $\Gamma$, cross states translate as the position of at least one robot being at the pole and the other at any of the other two circles but not at the vertex. A cross state in the chain $C$ will correspond to a point in a border circle that is not a j-point. See figures \ref{CSN} and \ref{CSC}. 
\end{defi}

 \begin{defi}

A {\em cross center} is a cross state  where two robots are at the intersection of horizontal and vertical network segments. A cross center corresponds to the positions  in which both robots are at a pole in different circles in $\Gamma$ . We can observe this state in the chain $C$ at the nodes that are not j-points. See figure \ref{CC}.

 \begin{figure}[H]
\caption{Cross center states in $N$ and $C$}
\label{CC}
\centering
\includegraphics[height=5cm]{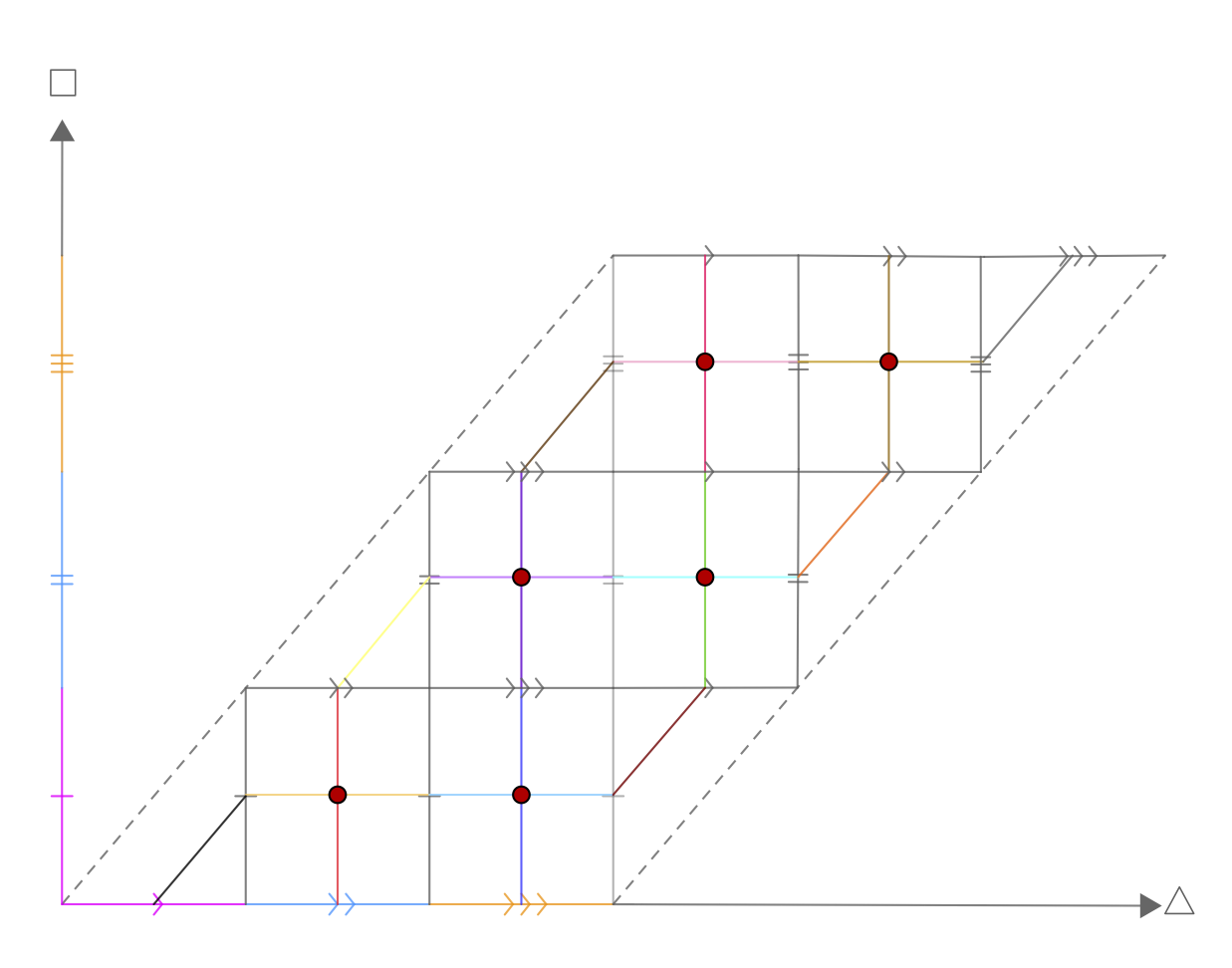}
\includegraphics[height=5cm]{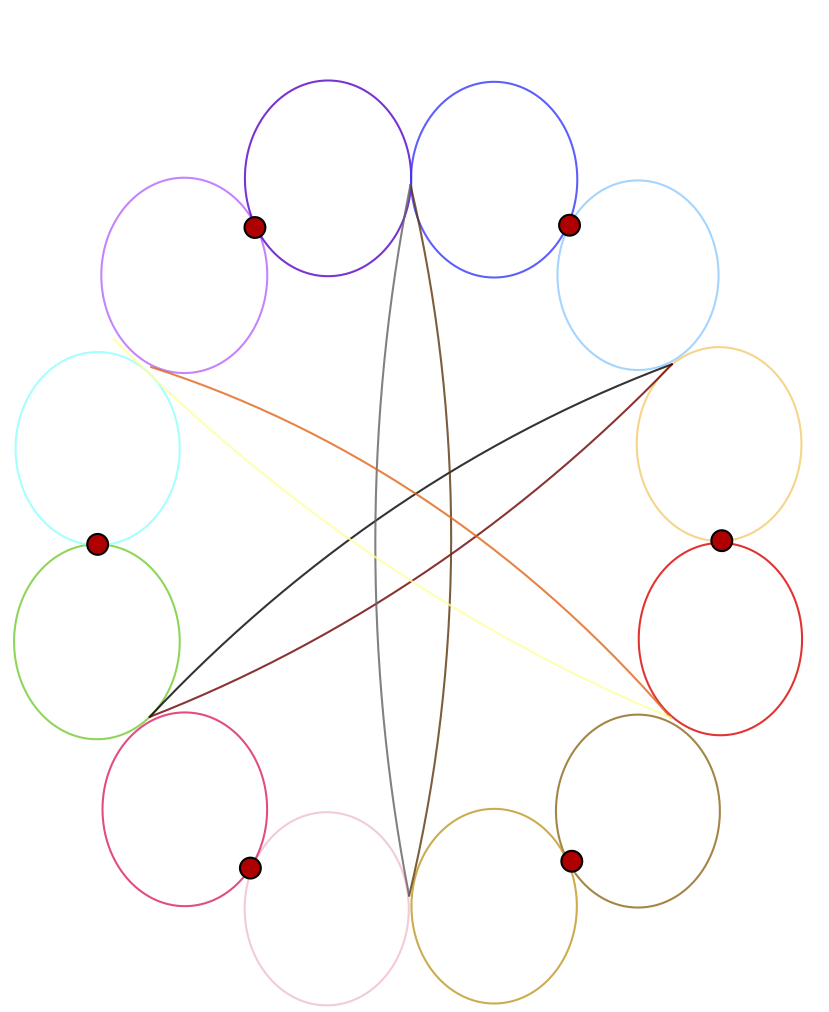}
\end{figure}

 \begin{figure}[H]
\caption{An example of cross center state in $\Gamma$}
\label{CC}
\centering
\includegraphics[height=4cm]{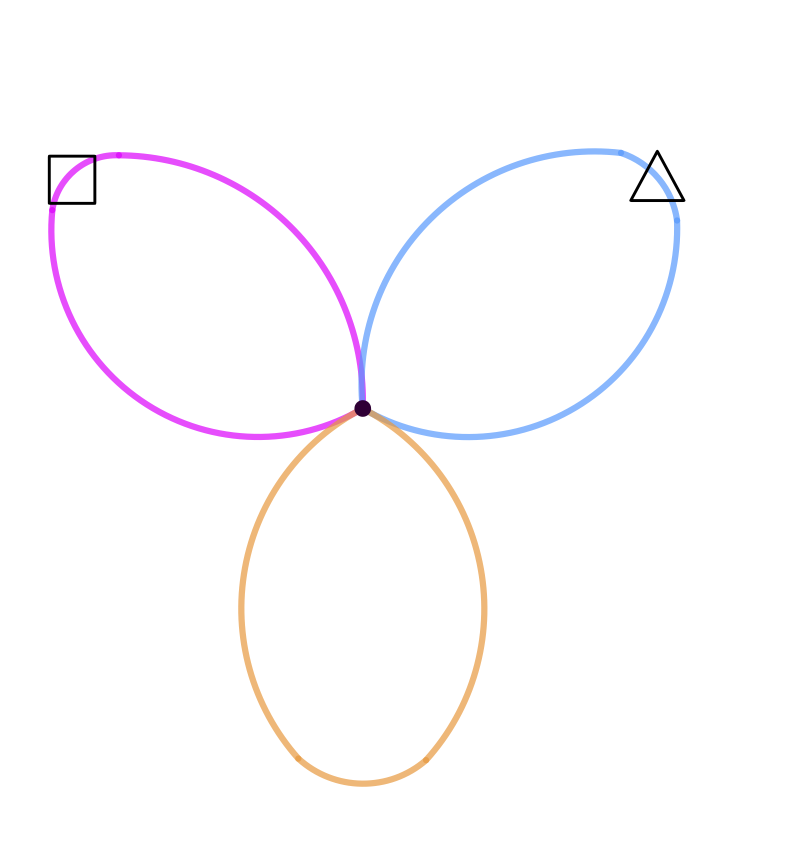}
\end{figure}

\end{defi}

\begin{defi}
When the state is in the interior of a diagonal segment in the network $N$, we will say that the state is a {\em diagonal state}. In $\Gamma$, this state will correspond to antipodal positions in any circle, but none is at the vertex. Diagonal states correspond to  points in connecting circles with the exception of j-points. See figures \ref{DSN} and \ref{DSC}.

\begin{figure}[H]
\caption{Diagonal states in $N$ and $C$}
\label{DSN}
\centering
\includegraphics[height=5cm]{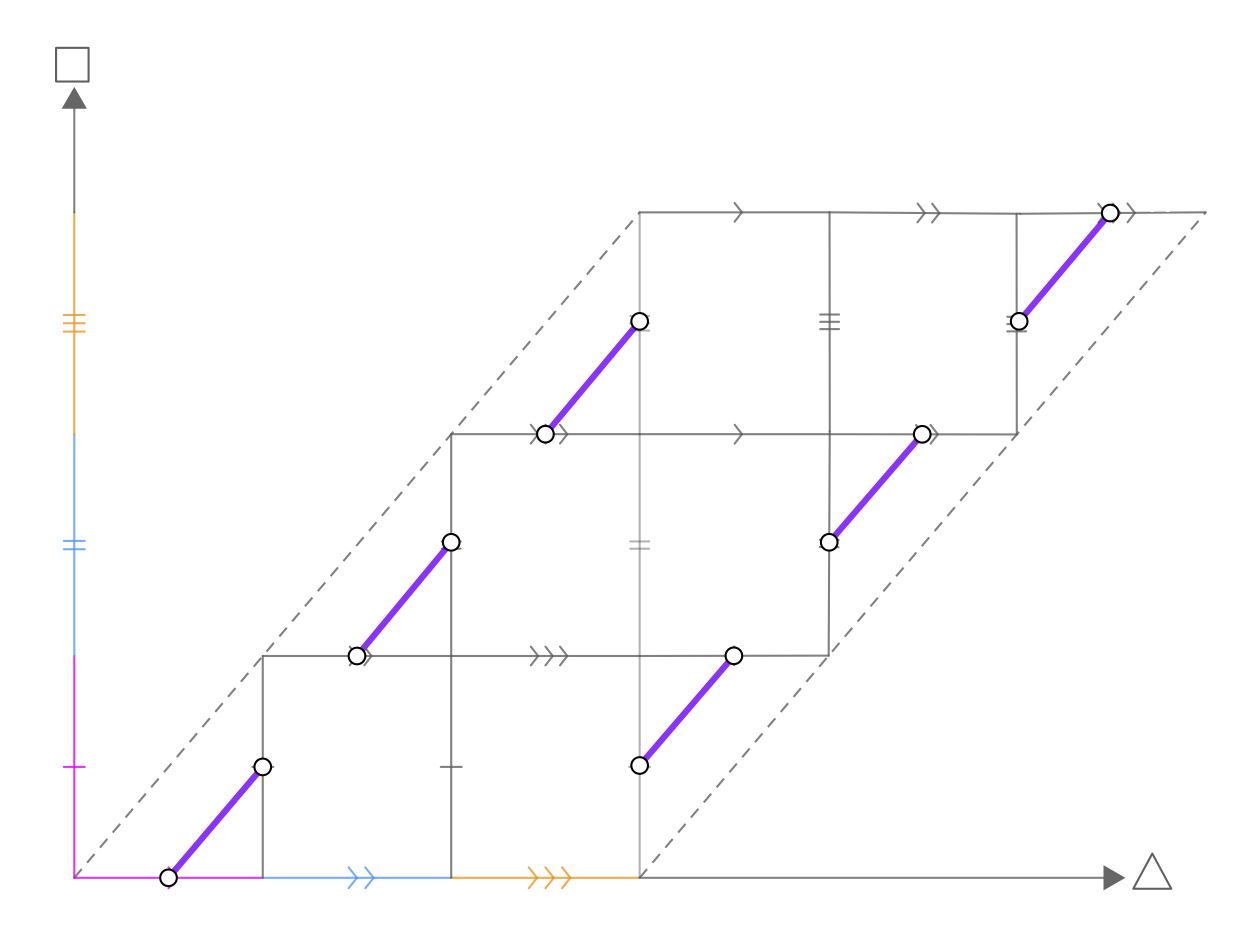}
\includegraphics[height=5cm]{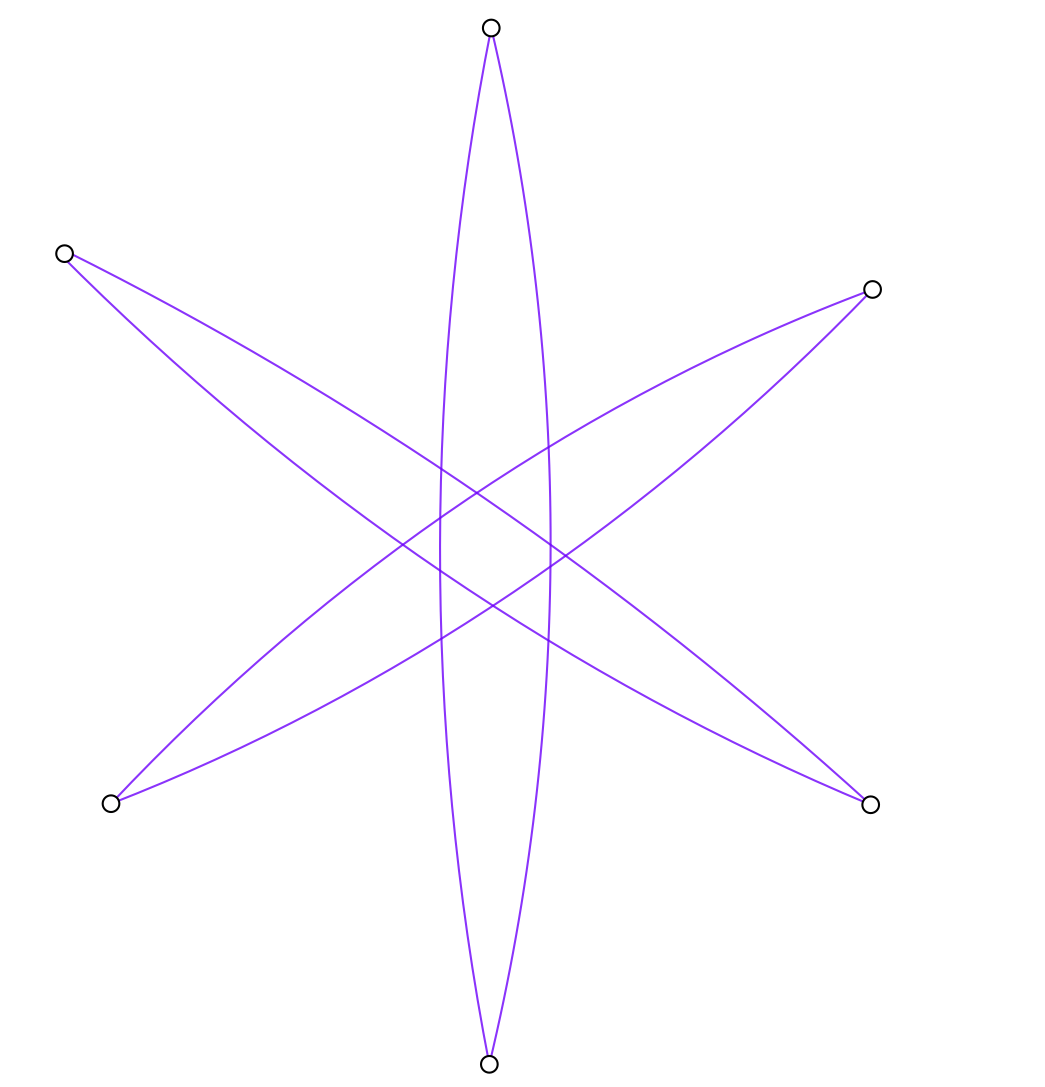}
\end{figure}

\begin{figure}[H]
\caption{An example of diagonal state in $\Gamma$}
\label{DSC}
\centering
\includegraphics[height=4cm]{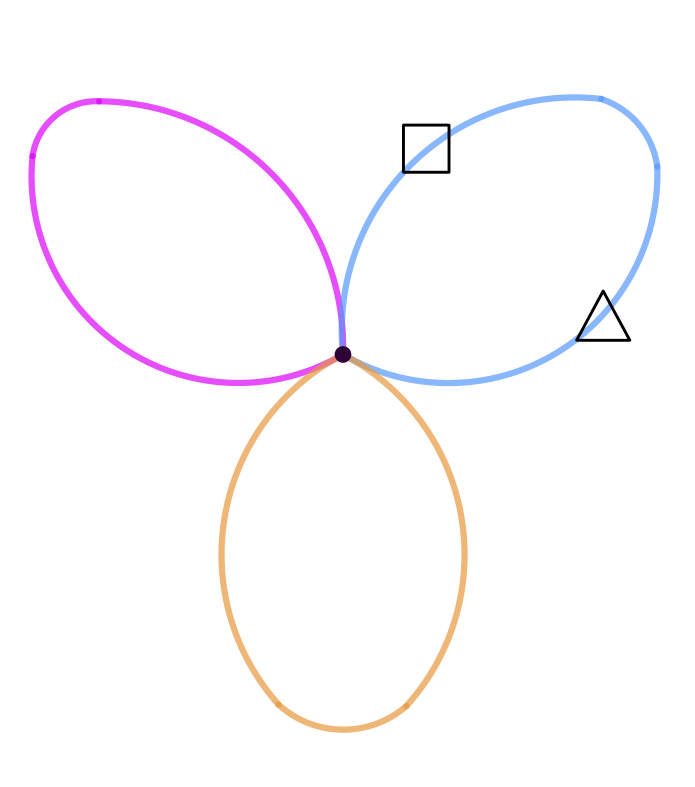}
\end{figure}

\end{defi}

We observe that j-states are states  in $N$ that are  neither diagonal nor  cross states.  There are two types of j-states: {\em vertical} and {\em horizontal}. Vertical j-states are those located between a diagonal segment and a vertical cross segment, whereas horizontal j-states are found between a diagonal segment and a horizontal cross segment.  
  
\begin{figure}[H]
\caption{ j-states in $N$ and $\Gamma$}
\label{jN}
\centering
\includegraphics[height=5cm]{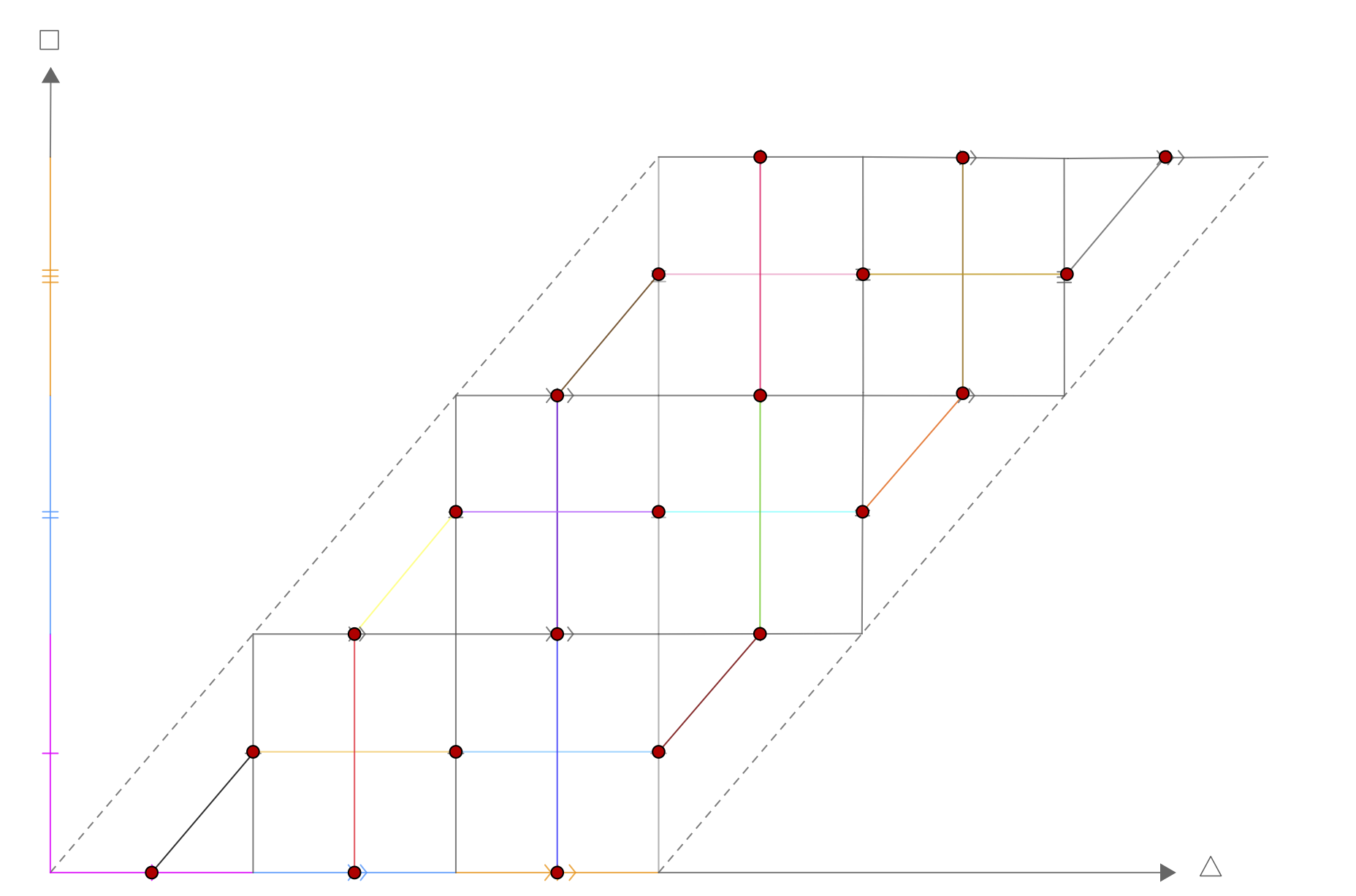}
\includegraphics[height=5cm]{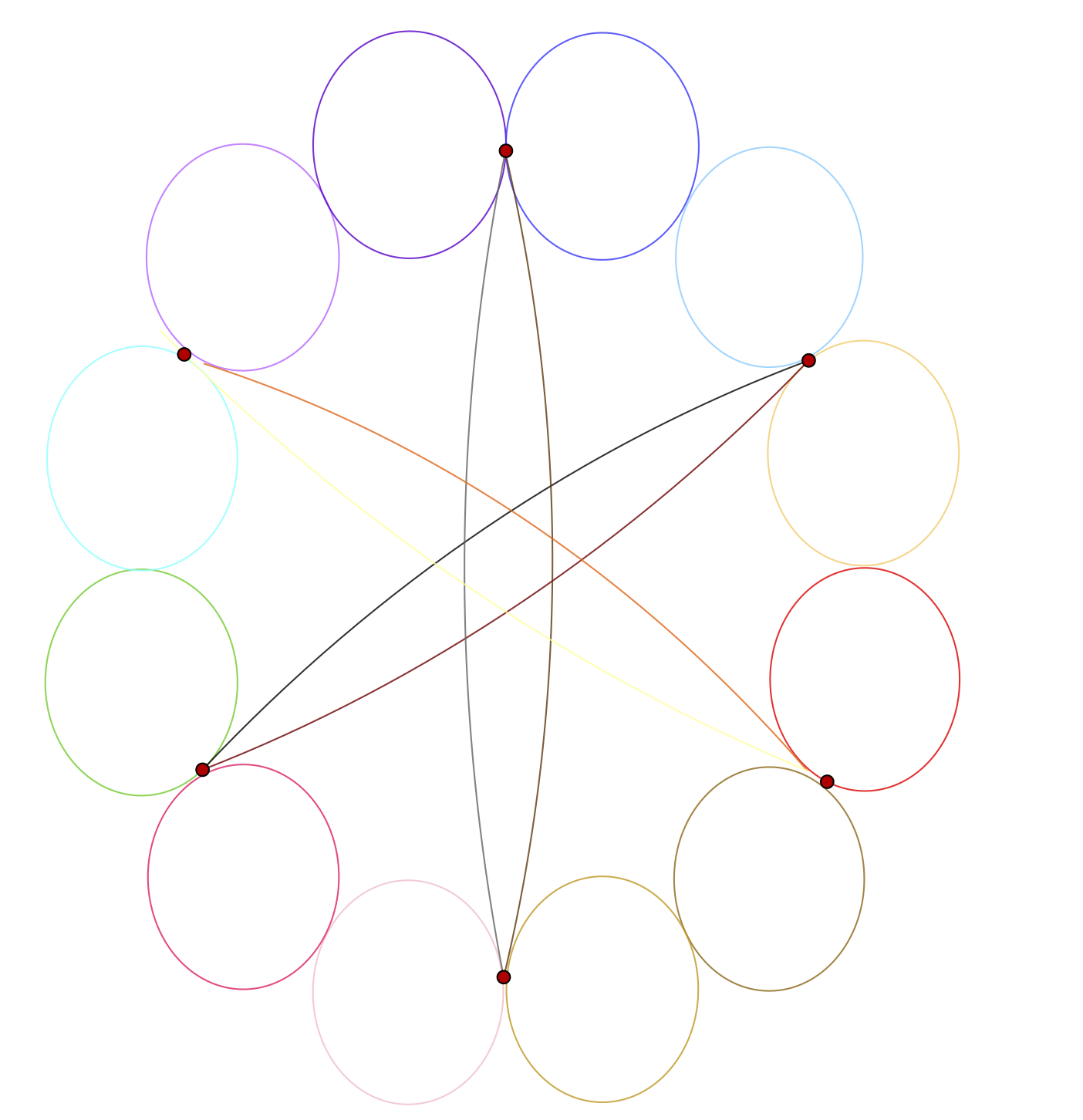}
\end{figure}

\begin{figure}[H]
\caption{An example of j-state in $\Gamma$}
\label{jC}
\centering
\includegraphics[height=4cm]{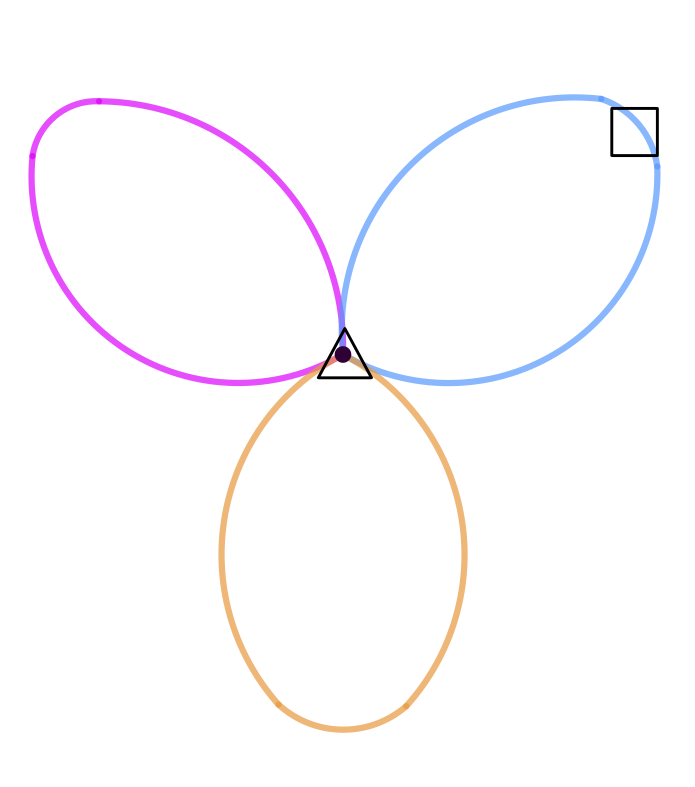}
\end{figure}

A j-state occurs when one robot in the physical space $\Gamma$ is at a center while the other is at a pole. In a vertical j-state, robot $B$ is at the vertex, whereas in a horizontal j-state, robot $A$ is at the vertex while the other robot is at a pole.  In the chain $C$, they correspond to the intersection points of border  and connecting circles. Vertical j-point are between a connecting circle and two vertical circles, and a horizontal j-point is between a connecting circle and two horizontal circles. See figure \ref{jN} and \ref{jC}.

\begin{figure}[H]
\caption{Counterclockwise direction in N and $C$}
\label{fig:counterclockwise}
\centering
\includegraphics[height=5cm]{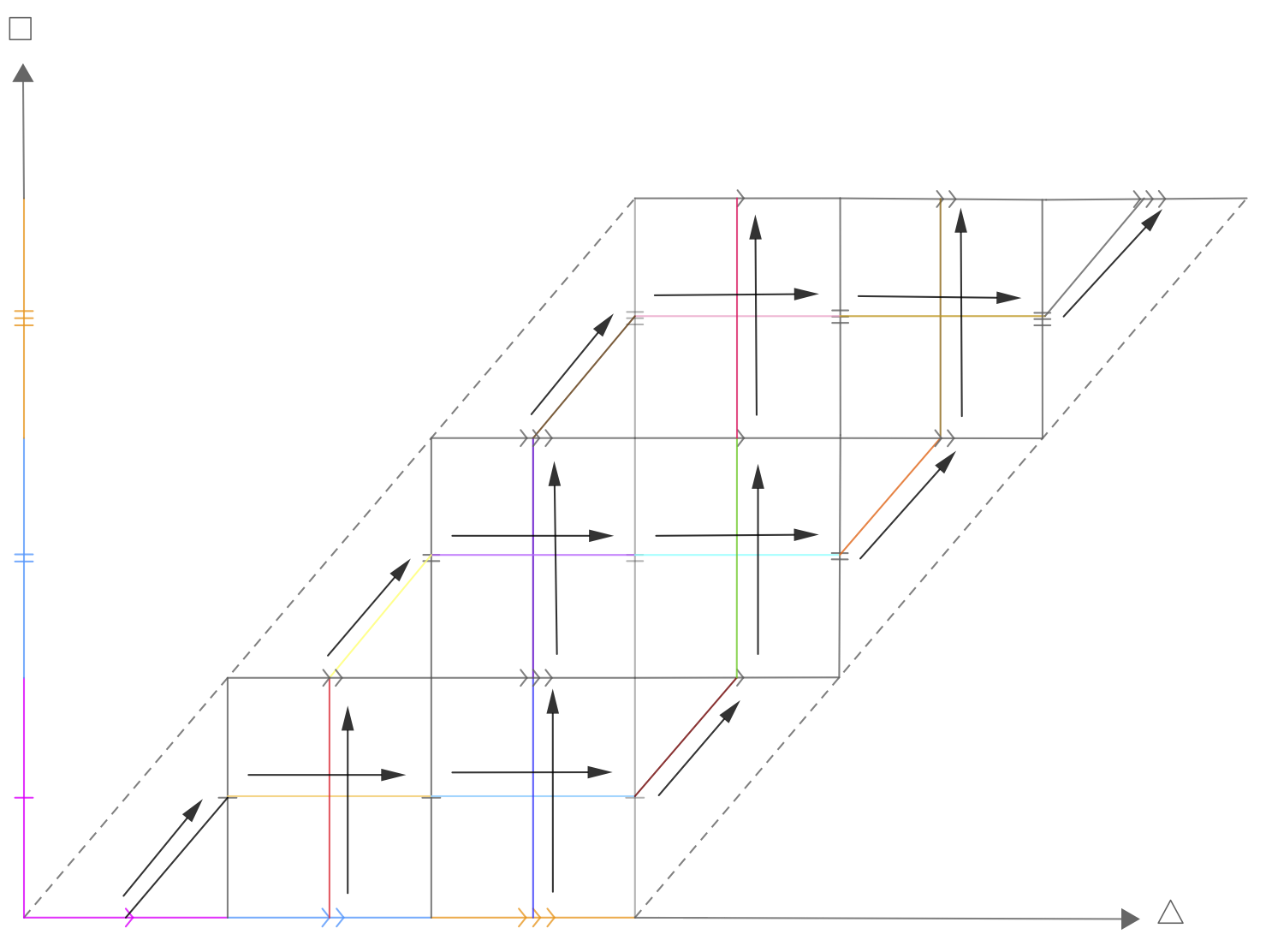}
\includegraphics[height=5cm]{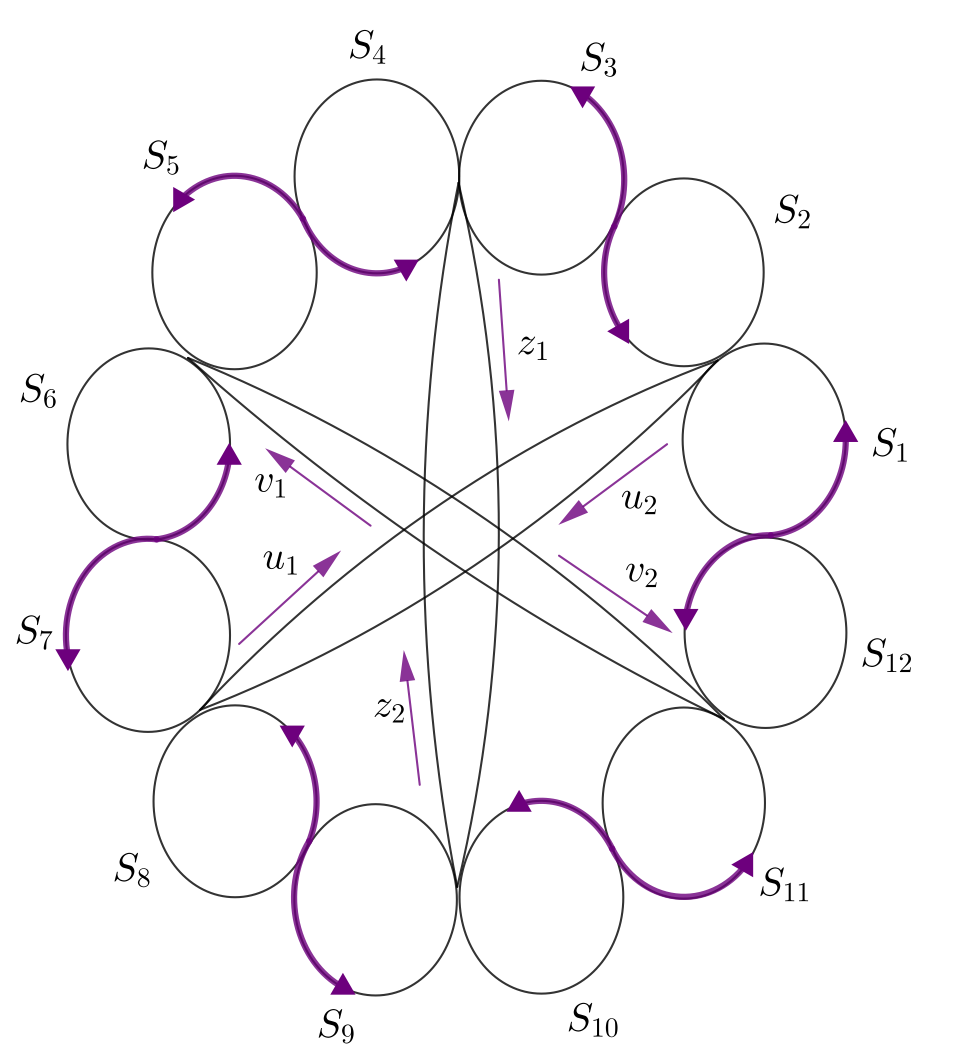}
\end{figure}

\begin{figure}[H]
\caption{Counterclockwise direction in $\Gamma$}
\label{fig:chaindire}
\centering
\includegraphics[height=4cm]{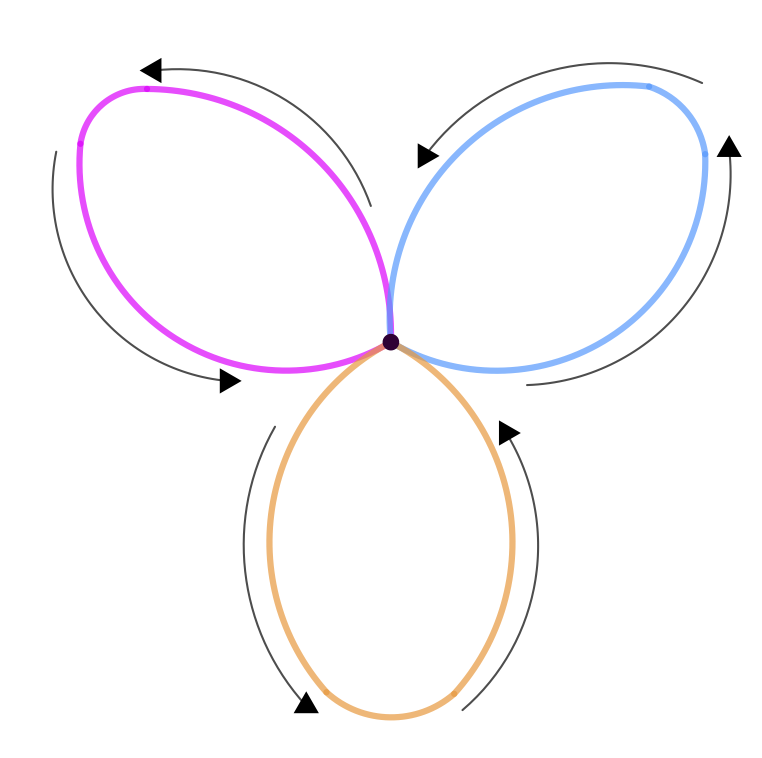}
\end{figure}

 In the network $N$ the counterclockwise orientation is given by the orientation in $X$: in $x$-axis from left to right and in $y$-axis upwards. This orientation in the physical space and the chain $C$ is illustrated in the figure \ref{fig:counterclockwise} and \ref{fig:chaindire}.
 
%%%%%%%%%%%%%%%%%%%%%%%%

\subsection{Algorithm in the configuration space}

In this section, we will describe the instructions needed to move a initial state  $P_i$ in $X$ to the final state $P_f$.

Recall that the first step of the algorithm consists of  projecting the initial and final states $P_i$ and $P_f$
into the network $N$ following the traces of the homotopy $H:X\times I\rightarrow X$ as in figure \ref{traces}. We will denote the projections $P^*_i= H_1(P_i)$ and $P^*_f= H_1(P_f)$ .
  
Next, we will construct the algorithm in the network $N$ to move from $P^*_i$ to $P^*_f$. The following definitions will help to describe the algorithm in the network $N$ and chain $C$.

 \begin{defi} The {\em initial node} ${CN}_i$ is the node in the chain $C$ corresponding to the cross center or j-state in $N$ at which the robots are initially located in N. If  $P^*_i$ is a cross state, then the initial chain node ${CN}_i$ is the cross center of that cross state. If  $P^*_i$ is a diagonal state, then the initial chain node ${CN}_i$ is the closest j-point in the counterclockwise direction. If $P^*_i$ is a node itself, then ${CN}_i=P^*_i$

Similarly,  the {\em final node} ${CN}_f$ is the node in the chain $C$ corresponding to the final state $P^*_f$.
\end{defi}

\begin{defi} The {\em zigzag circle} is the union of the exterior semicircles in the chain $C$. See figure \ref{exteriorC} and \ref{exteriorN}. 
\end{defi}

\begin{figure}[H]
\caption{Zigzag circle in $N$ and $C$}
\centering
\label{zigzagNC}
\includegraphics[height=5cm]{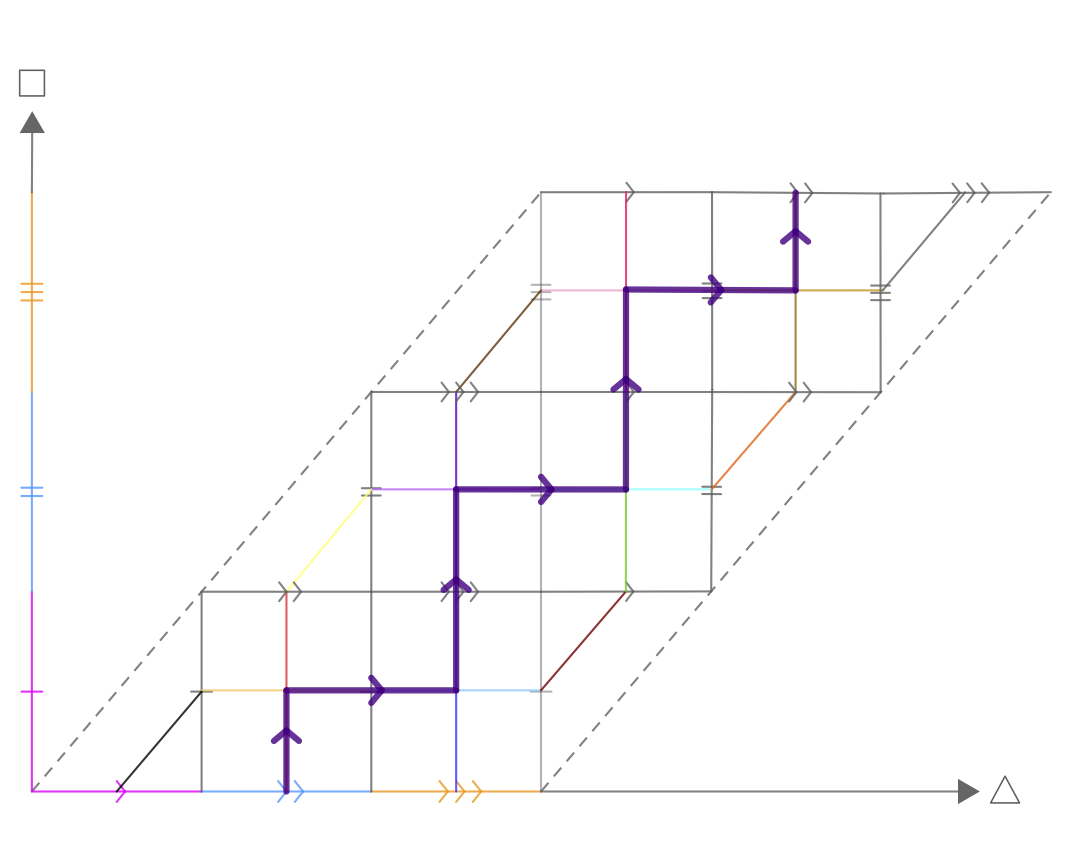}
\includegraphics[height=5cm]{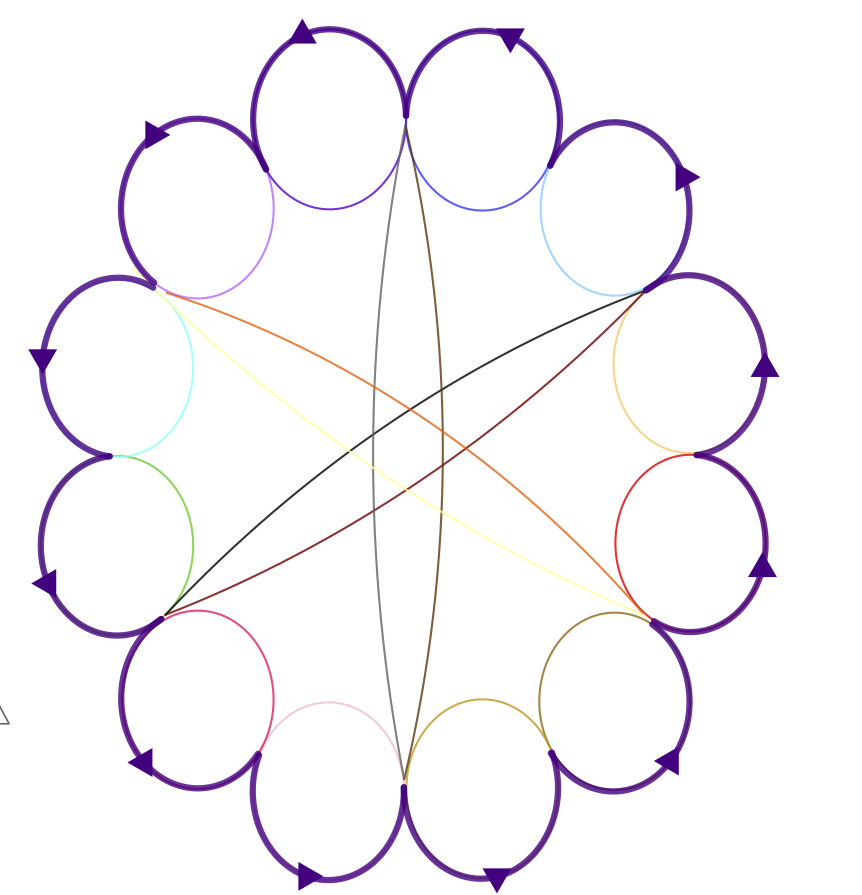}
\end{figure}

In figure \ref{zigzagNC} we show the zigzag circle in both network $N$ and chain $C$ with the counterclockwise orientation.

\begin{figure}[H]
\caption{Example of zigzag circle positions in $\Gamma$}
\centering
\label{halftwo}
\includegraphics[height=4cm]{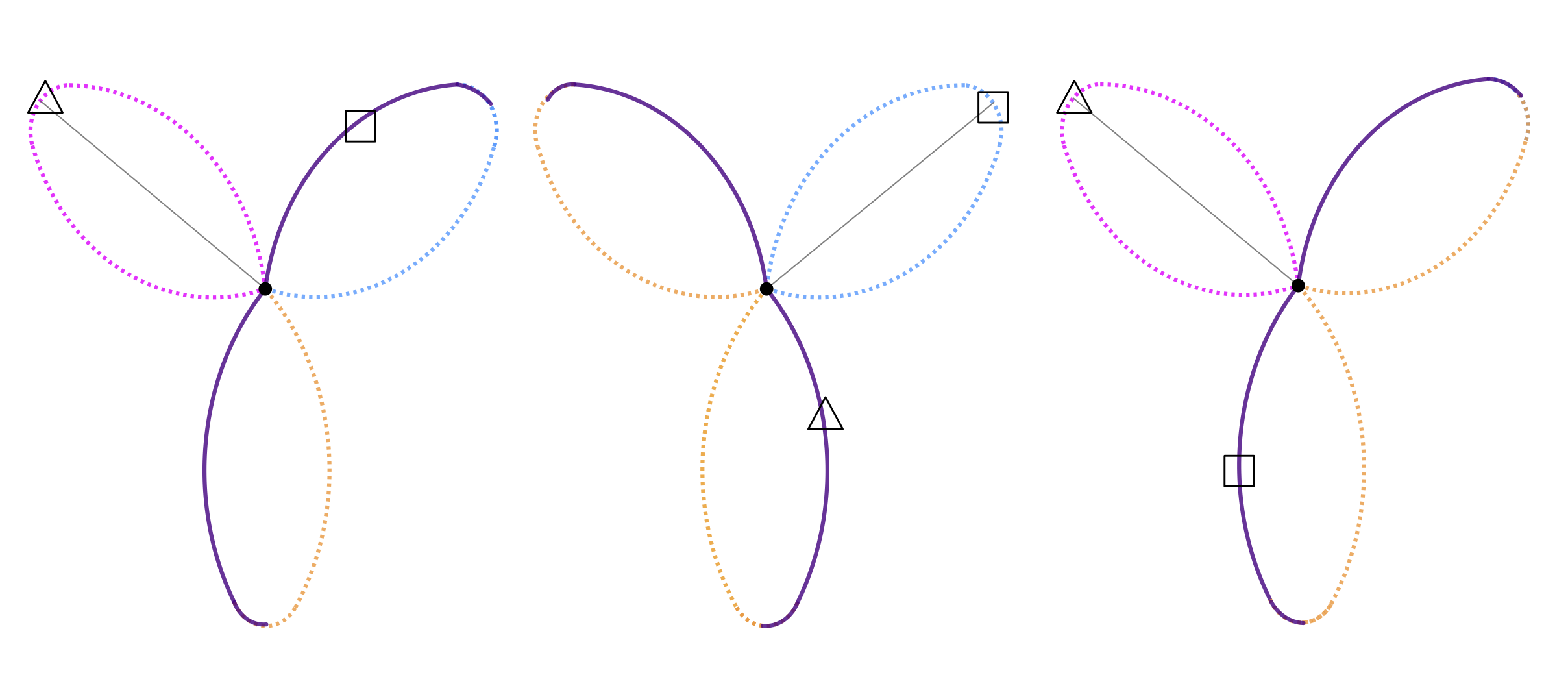}
\end{figure}

Observe that the zigzag circle positions that are not nodes, correspond in $\Gamma$  to configurations in which the robot that is not at a pole or vertex is positioned at one of the two innermost semicircles opposite to the robot at the pole. We show sample positions of robots in the zigzag circle in figure \ref{halftwo}. 

Once that the states are nodes, the idea of our algorithm will be to move robots from a node to another following the zigzag circle in counterclockwise direction. 
 
First we move the initial and final states $P_i^*$ and $P_f^*$ in N to initial and final nodes ${CN}_i$ and ${CN}_f$ following shortest path or counterclockwise shortest path as described before.

As a result of this step, the states are nodes in the network $N$. 
   
   Next, we move counterclockwise in zigzag circle from $CN_i$ to $CN_f$, which is the fundamental step of our algorithm. 
   
   Finally, we reverse the motion from  $CN_f$ to $P_f^* $.
    
\begin{alg}\label{N}
Therefore, the algorithm in the network $N$ is as follows. 

\begin{enumerate}
        \item {\textbf{First Step}} If $P_i^*$  is a cross state, move shortest path to initial chain node ${CN}_i$. If it is a diagonal state, move counterclockwise to  ${CN}_i$.  If $P_i^*$ is a node, stay motionless. Do the same with the final state to obtain ${CN}_f$.
	
	\item { \textbf{Second Step}} Move from ${CN}_i$ to ${CN}_f$ following the zigzag circle counterclockwise.   If ${CN}_i={CN}_f$, stay motionless.
	
	\item{ \textbf{Third Step}}  Reverse the movement in first step to obtain $P_f^*$. 
\end{enumerate}
\end{alg}

\begin{alg}
When transitioning to the chain $C$, we obtain the following motion planning algorithm:
		
\begin{enumerate}

\item {\textbf{First Step}} If $P_i^*$  is on a border circle, move shortest path to initial chain node ${CN}_i$. Otherwise, move counterclockwise to ${CN}_i$.  Do the same with the final state to obtain ${CN}_f$.

      \item { \textbf{Second Step}}  Move the robots from ${CN}_i$ to ${CN}_f$  counterclockwise through exterior circles in the chain $C$. If ${CN}_i={CN}_f$, stay motionless.
      
      \item{ \textbf{Third Step}}  Reverse the movement in first step to move back from ${CN}_f$ to $P_f^*$ . 

\end{enumerate}
\end{alg}

We can extend the range of the algorithm in the network and chain to all sets of points in the configuration space $X$ by adding at the beginning and end, the steps given by the traces of the homotopy.

\begin{alg}
Hence, the algorithm in the configuration space $X$ is:

\begin{enumerate}
	\item {\textbf{Preliminary Step}} Move from $P_i$ to $P_i^*$ following the traces of the homotopy. Do the same to move from $P_f$ to $P_f^*$.
		
	\item { \textbf{Main Step}} Move from $P_i^*$ to $P_f^*$ following the above algorithm \ref{N} described in $N$.
	
	\item{ \textbf{Final Step}}  Reverse the movement in preliminary step to move from $P_f^*$ to $P_f$.
	
	\end{enumerate}
\end{alg}

%%%%%%%%%%%%%%%%%%%%%%%%%%

\subsection{Algorithm in the physical space}

We deduce now the instructions in the physical space $\Gamma$. The goal will be first to move the robots such that they will be in a network position, then to translate the algorithm in N to the physical space.

The movement determined by the traces of the homotopy is translated to the physical space as follows: if the robots are in the same circle, move them away from each other until they reach an antipodal position and if they are in different circles, move them until one reaches a pole position. In both cases ratio of the robots' speeds are given by the slopes of the homotopy traces shown in figure \ref{traces}.

Once that initial and final positions are retracted to the network, we describe the movement within the network. The first step is to move any point in the network to a node. Then the movement will proceed between nodes.

The fundamental observation here is that the counterclockwise movement following the zigzag circle in the chain $C$ restricts the movement in the physical space as shown in figure \ref{halftwo}.

Recall that states in the network can be described as positions in which at least one robot is at a pole if the robots are in different circles and antipodal positions if the robots are in the same circle. If the robots are not already at a node, 
move the robot that is not at a pole to the pole following the shortest path if they are  at different circles and  move the robots simultaneously counterclockwise until one arrives at the vertex and the other at a pole if they are in the same circle.

\begin{defi}
We will say that the two robots are in a {\em pole-pole} position in $\Gamma$ if they are both at poles. Observe that these positions correspond to cross center states in the network $N$.
\end{defi}

\begin{defi}
We will say that the two robots are in a {\em pole-vertex} position in $\Gamma$ if one is at a pole and another is at the vertex. Observe that these positions correspond to j-states in the network $N$.
\end{defi}

Therefore, all nodes in the network can be described in the physical space as pole-pole or pole-vertex positions.

Now we will describe the movement in the physical space between these positions following the counterclockwise direction in the zigzag circle. We will call {\em allowed semicircles} to the two innermost semicircles opposite to the stationary robot as shown in figure \ref{halftwo}.

The following movements $m_{VP}$ and $m_{PV}$ describe the way to exit a pole-vertex or a pole-pole position. 
\begin{enumerate}
\item[$m_{VP}$] From pole-vertex to contiguous pole-pole: Move the robot that is at the vertex counterclockwise in the allowed semicircles until it reaches the pole. 
\item[$m_{PV}$] From pole-pole to contiguous pole-vertex: Move the only robot that can move counterclockwise in the allowed semicircles until it reaches the vertex. 
\end{enumerate}

\begin{defi}
A {\em zigzag movement from a pole-vertex to a pole-pole position} is a series of concatenated movements $m_{VP}m_{PV}m_{VP}\cdots m_{VP}$ as before where the first movement originates in the given pole-vertex position and the last movement finished at the given pole-pole position.
Similarly, a {\em zigzag movement from a pole-pole to a pole-vertex position} is a series of concatenated movements $m_{PV}m_{VP}\cdots m_{PV}$.
\end{defi}

\begin{alg}\label{physicalN}
The algorithm for network positions is as follows.
\begin{enumerate}
\item {\textbf{First Step}} If the two robots are at a pole-pole or pole-vertex position, stay motionless. Otherwise, if the robots are at different circles, move the robot that is not at a pole to the pole following the shortest path. If they are in the same circle, move the robots simultaneously counterclockwise until one arrives at the vertex and the other at a pole. Do the same for the final positions.  Let us call {\em initial pole position} and {\em final pole position} the output of this step.

\item{ \textbf{Second Step}} Move from the initial pole position to the final one following the zigzag movement.

 \item{ \textbf{Third Step}}  Reverse the movement in first step to go from the final pole position to the final position in the network.

\end{enumerate}
\end{alg}

\begin{alg} To extend the previous algorithm to all the possible positions in the physical space, we add the preliminary and final steps given by the traces of the homotopy.

\begin{enumerate}
	
	\item {\textbf{Preliminary Step}} If the initial position of the robots are in the same circle, move them away from each other until they reach an antipodal position and if they are in different circles, move them until one reaches a pole position. 
 Do the same with the final position. Let us call {\em initial network position} and {\em final network position} the output of this step
		
	\item { \textbf{Main Step}} Move from the initial to the final network position  as in algorithm \ref{physicalN}.
	
	\item{ \textbf{Final Step}}  Reverse the movement in preliminary step to move from the final network position to the final position.
	
	\end{enumerate}
\end{alg}

The above algorithm is discontinuous: if the initial or final configuration is a node, then a small perturbation of $P$ may lead to a different motion. Note that if, for instance, we restrict the algorithm to the states $P_i$ and $P_f$ that are not nodes the algorithm is continuous here.

We previously proved that $TC(X)$ is 3. Our objective then will be to exhibit the three domains of continuity.

We will restrict the algorithm to the pairs in the following three regions $U'$, $V'$, and $W'$ of the Cartesian product $X\times X$. 
   The first region $U'$ is the set of points $P_i$ and $P_f$, such that none of them is a node,  
   $$U'=\{(P_i, P_f) / P_i\not \in CN, P_f\notin CN \}.$$ 
   The second region $V'$ consists of the pair of points, initial and final, such that one of them is a node and the other is not, $$V'=\{(P_i, P_f) / (P_i\not \in CN,  P_f\in CN) \mbox{ or } (P_i\in CN,  P_f\not\in CN) \}.$$ 
   The third region $W'$ covers the rest of the Cartesian product, $$W'=\{(P_i, P_f) /  P_i\in CN,  P_f\in CN\} $$  where both initial and final points are nodes.

Let $U$, $V$, and $W$ be small open neighborhoods of $U'$, $V'$, and $W'$ respectively. These are our domains of continuity. 
\subsubsection{Sample case}

Now, we will show how our motion planning algorithm works in a sample case. 
%Observe that the initial and final configuration does not have a corresponding motion in chain $C$. 

\begin{figure}[H]
\caption{Preliminary Step in $X$:  $P_i\rightarrow P_i^*$, $P_f\rightarrow P_f^*$}
\centering
\includegraphics[height=6cm]{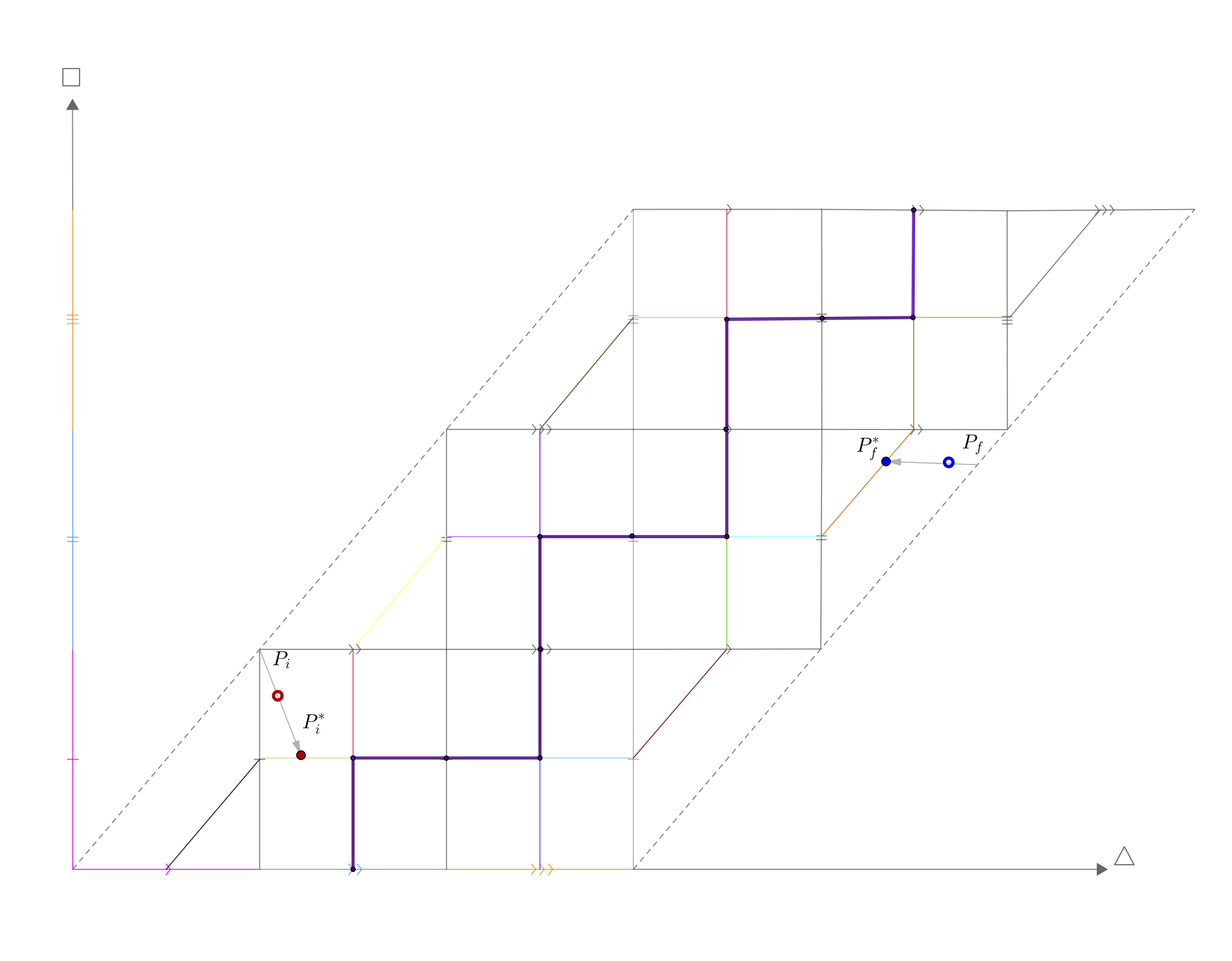}
\end{figure}

\begin{figure}[H]
\caption{Preliminary Step in $\Gamma$}
\centering
\includegraphics[height= 5cm]{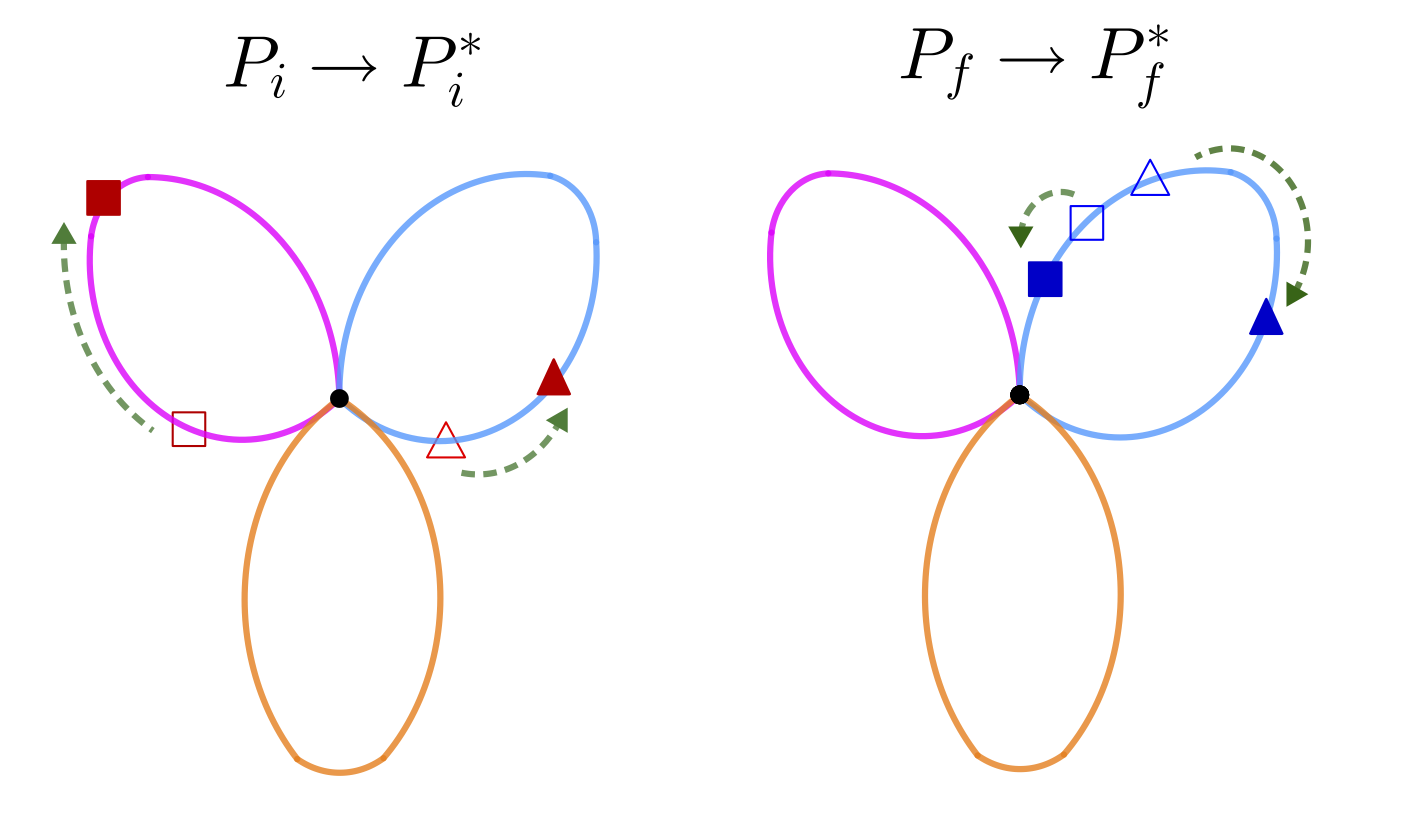}
\end{figure}

\begin{figure}[H]
\caption{First Step in $N$ and $C$: $P_i^*\rightarrow {CN}_i$,  $P_f^*\rightarrow {CN}_f$}
\centering
\includegraphics[height=6 cm]{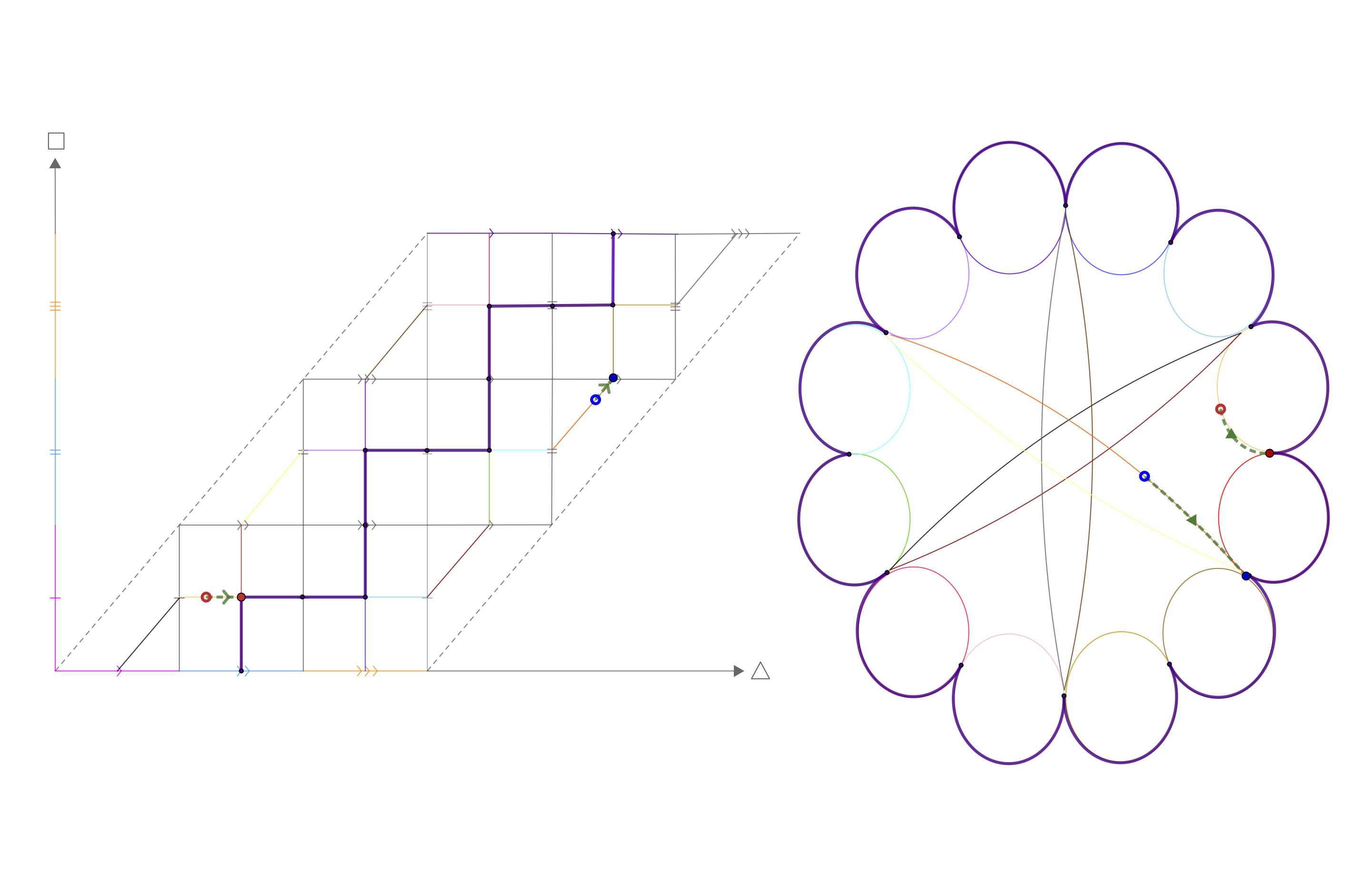}
\end{figure}

\begin{figure}[H]
\caption{First Step in $\Gamma$}
\centering
\includegraphics[height=5cm]{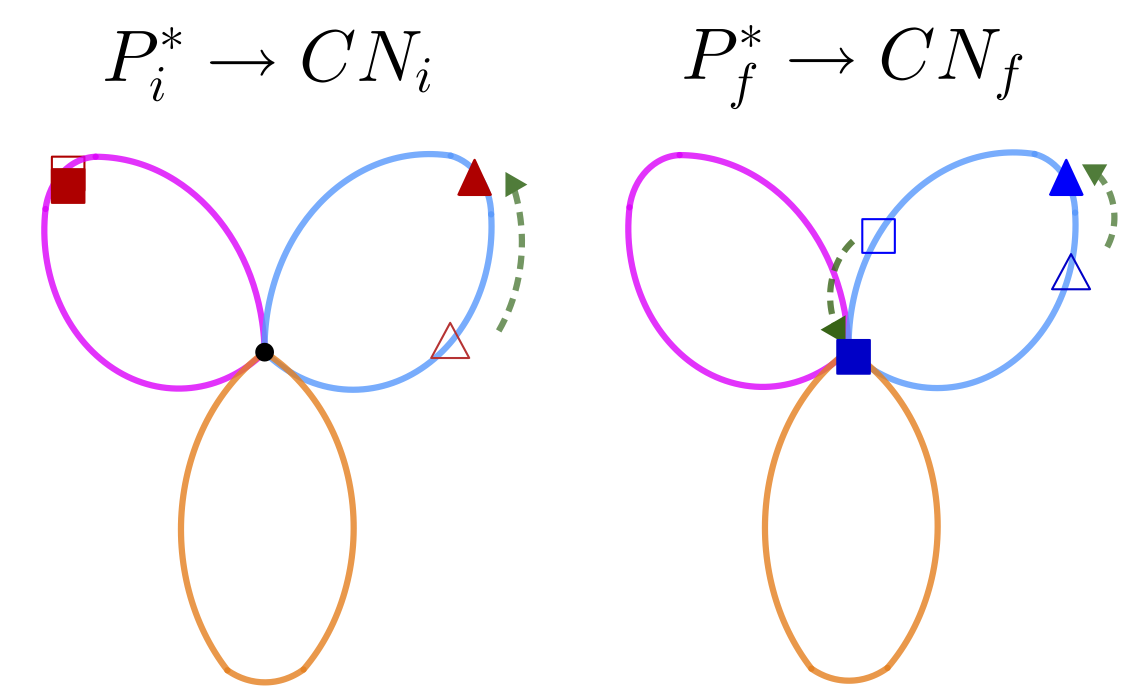}
\end{figure}

\begin{center}
\begin{figure}[H]
\caption{Second Step in $N$ and $C$: Move from ${CN}_i$ to ${CN}_f$}
\includegraphics[height=6cm]{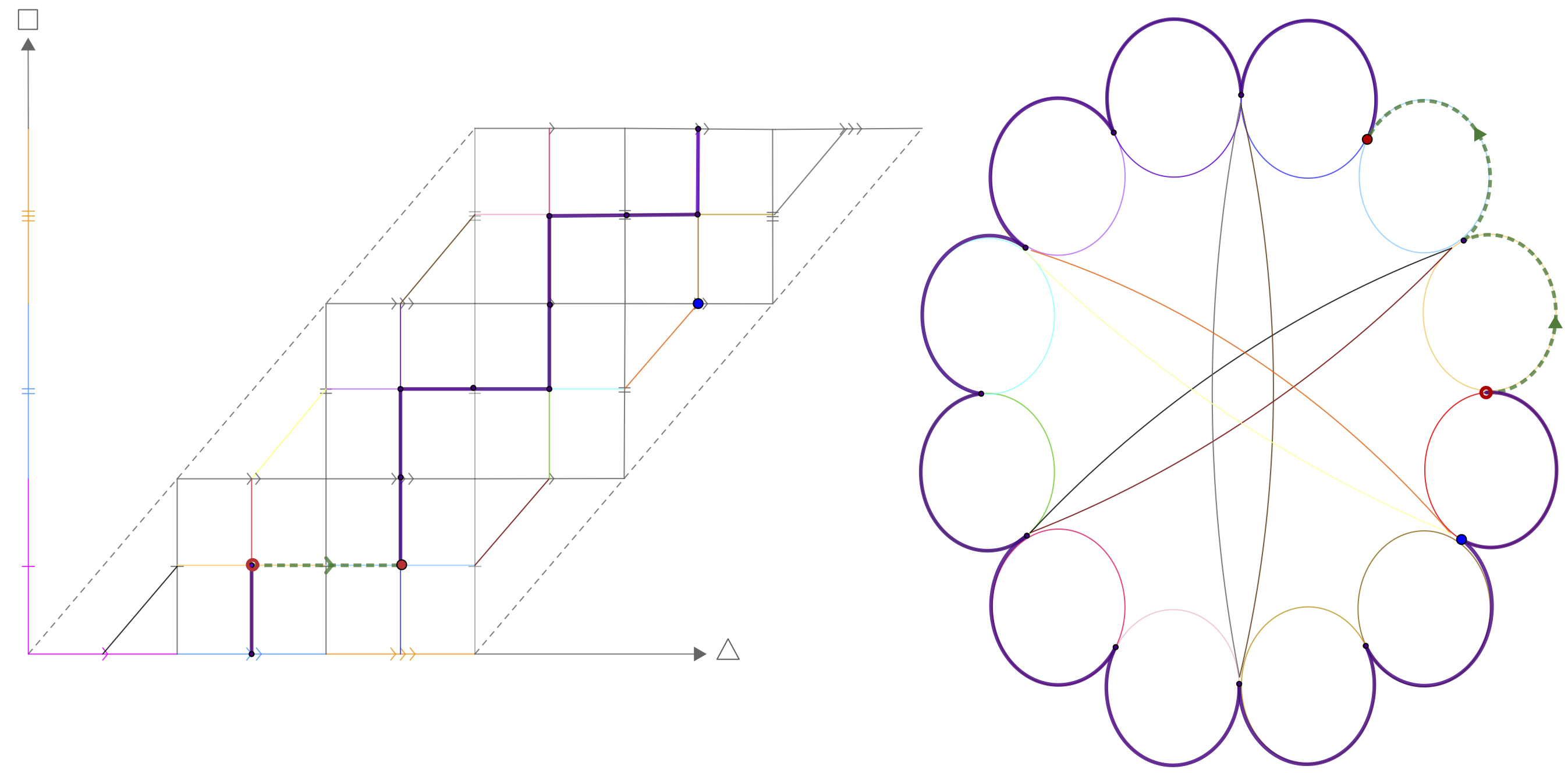}
\includegraphics[height=6cm]{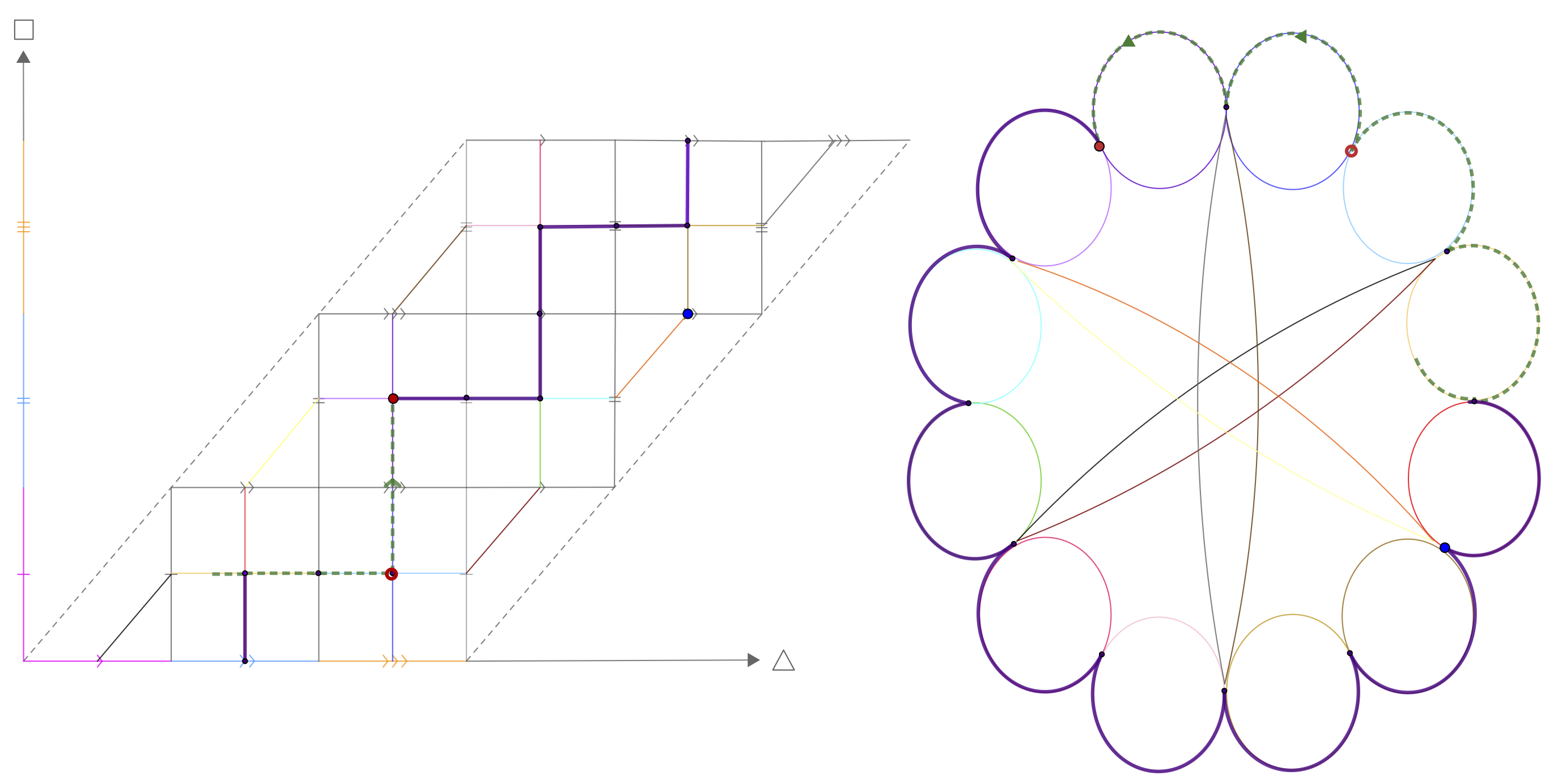}
\includegraphics[height=6cm]{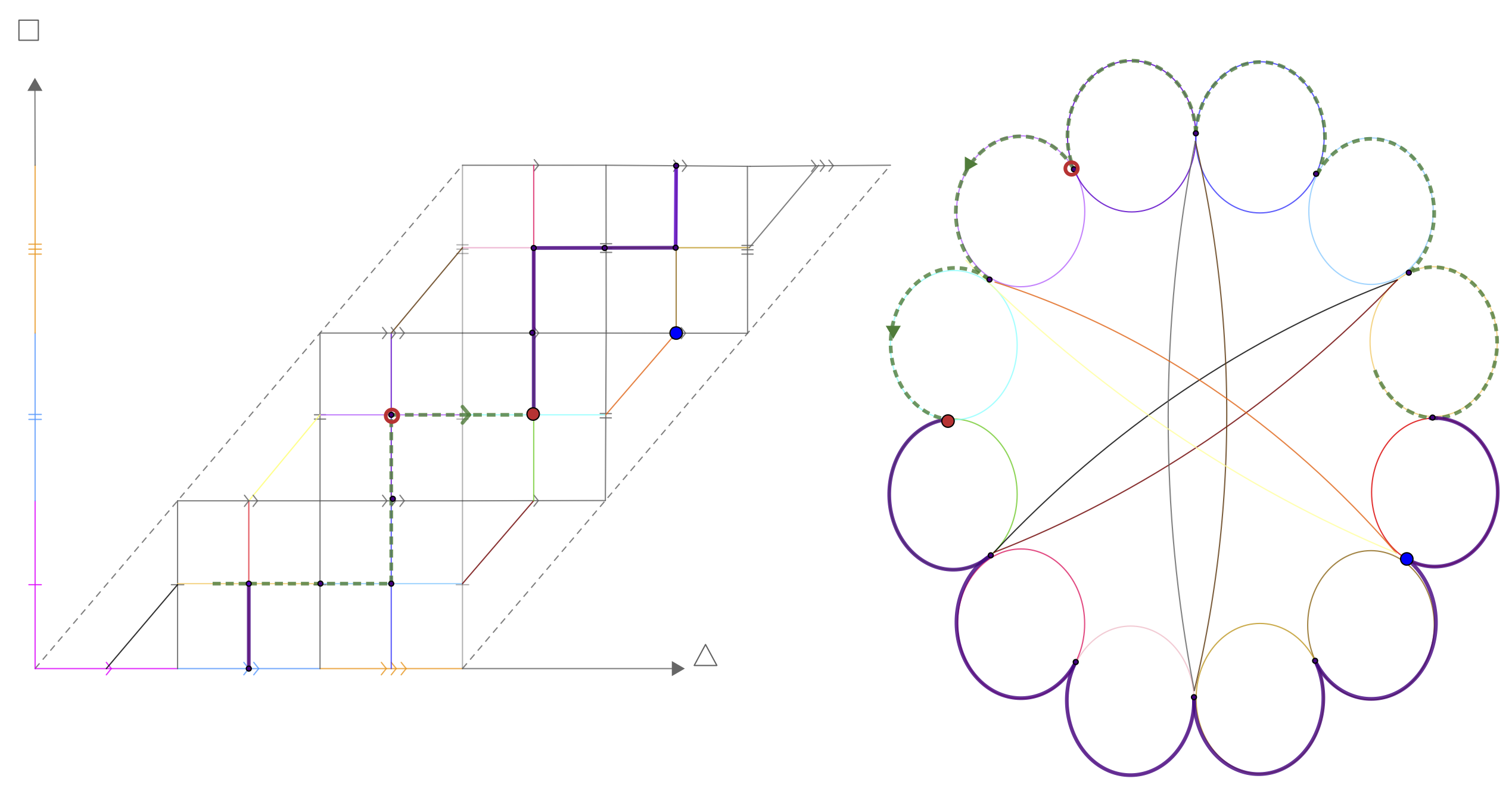}
\end{figure}
\end{center}

\begin{center}
\begin{figure}[H]
\includegraphics[height=6cm]{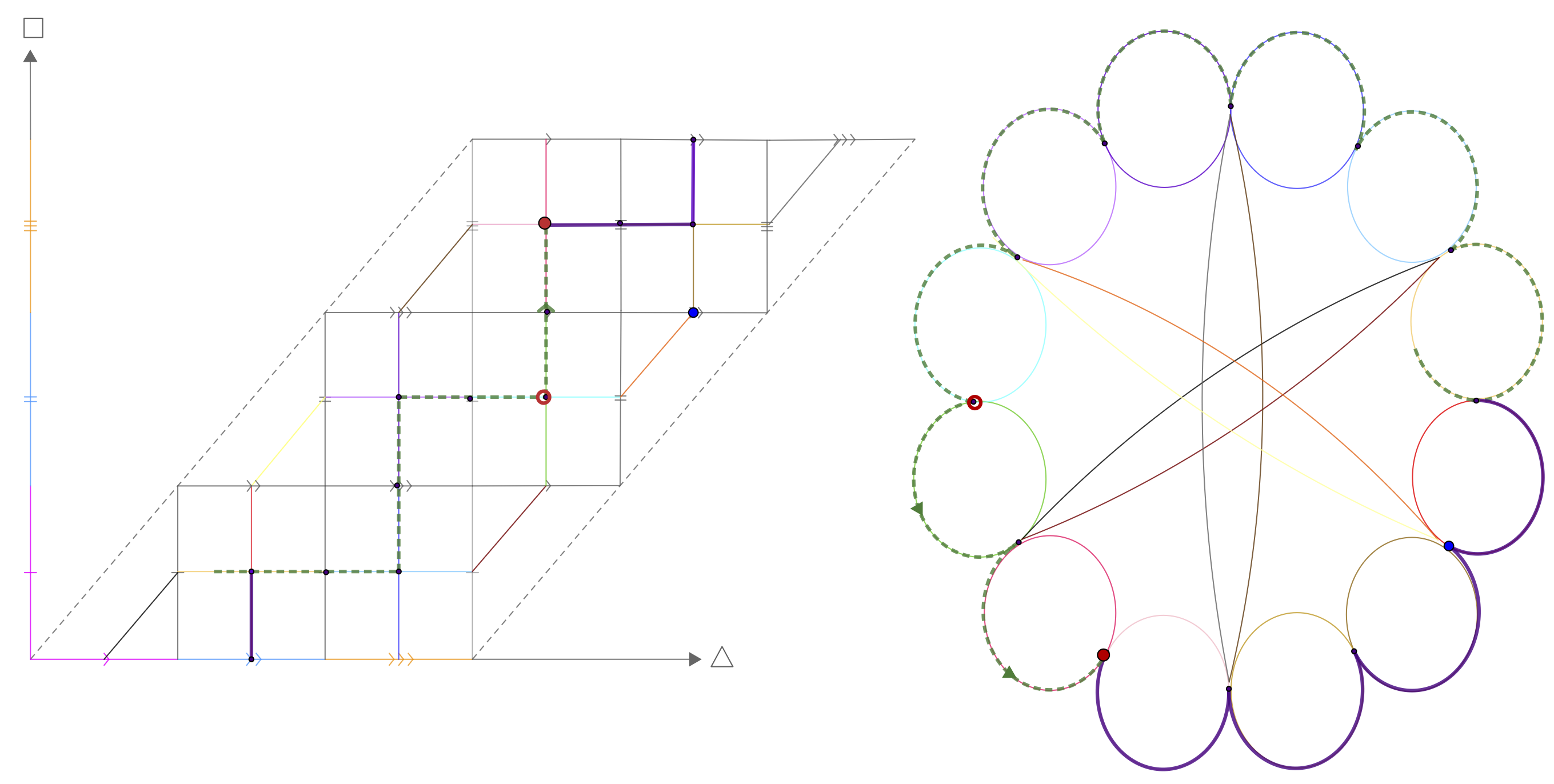}
\includegraphics[height=6cm]{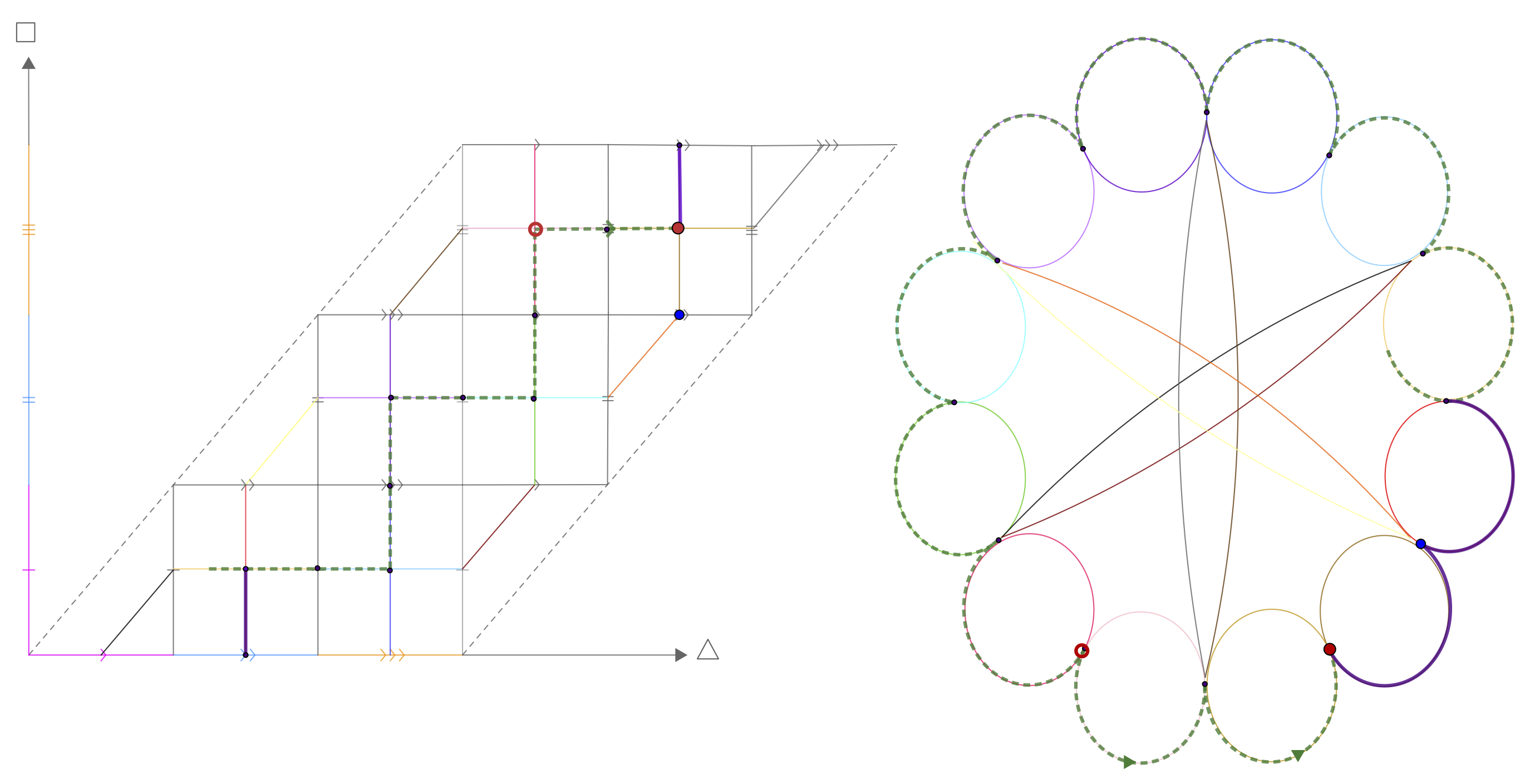}
\includegraphics[height=6cm]{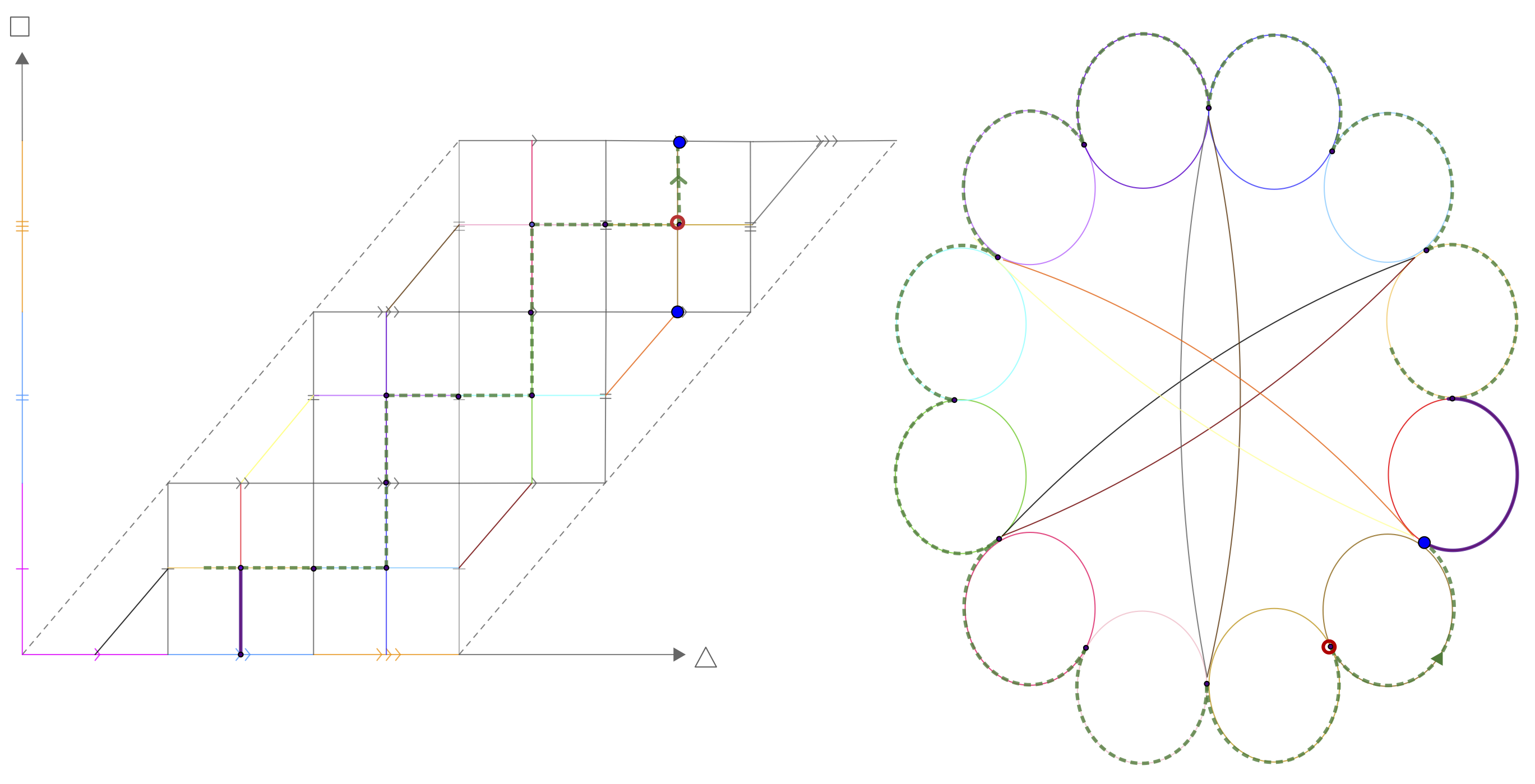}
\end{figure}
\end{center}

\begin{center}
\begin{figure}[H]
\caption{Second Step in $\Gamma$}
\includegraphics[height=4cm]{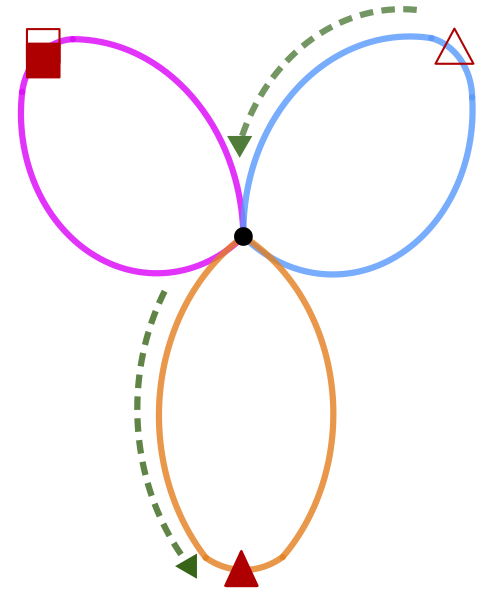}
\includegraphics[height=4cm]{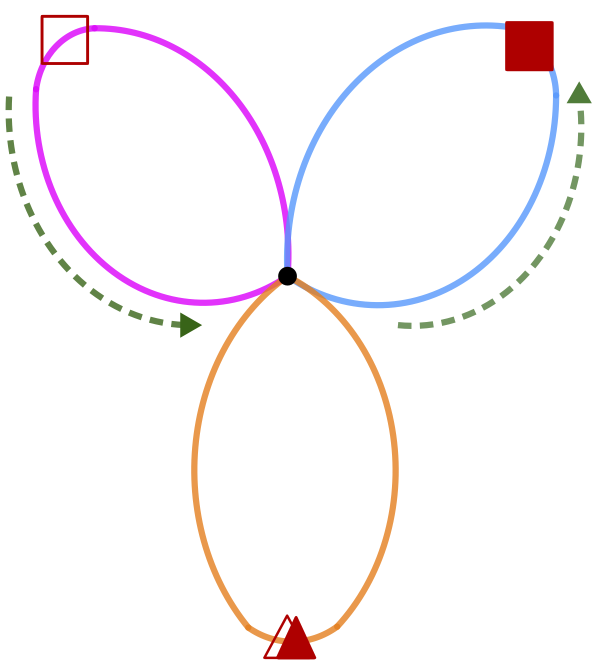}
\includegraphics[height=4cm]{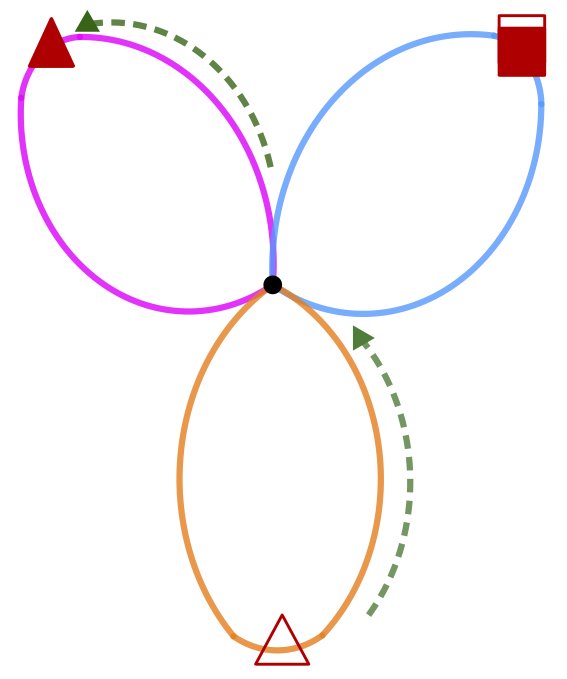}\\
\includegraphics[height=4cm]{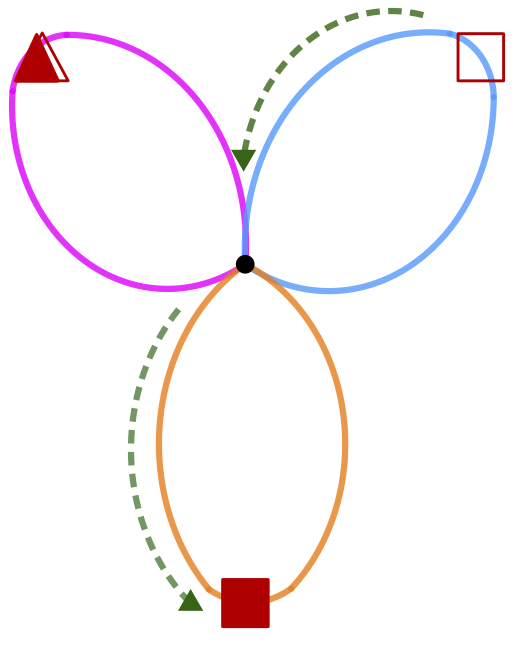}
\includegraphics[height=4cm]{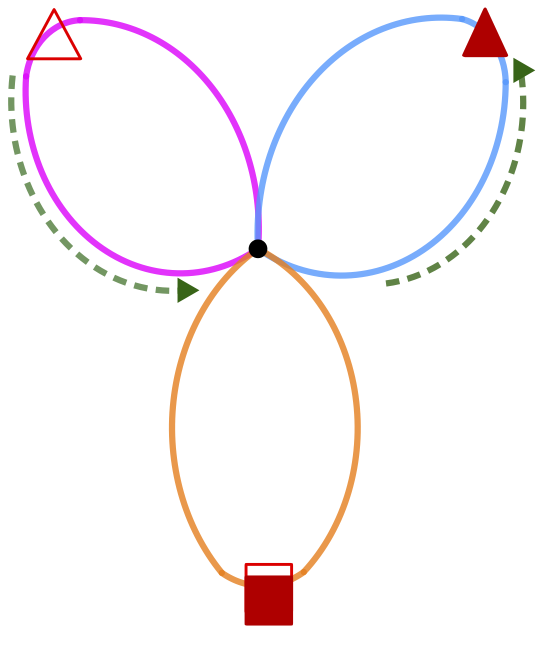}
\includegraphics[height=4cm]{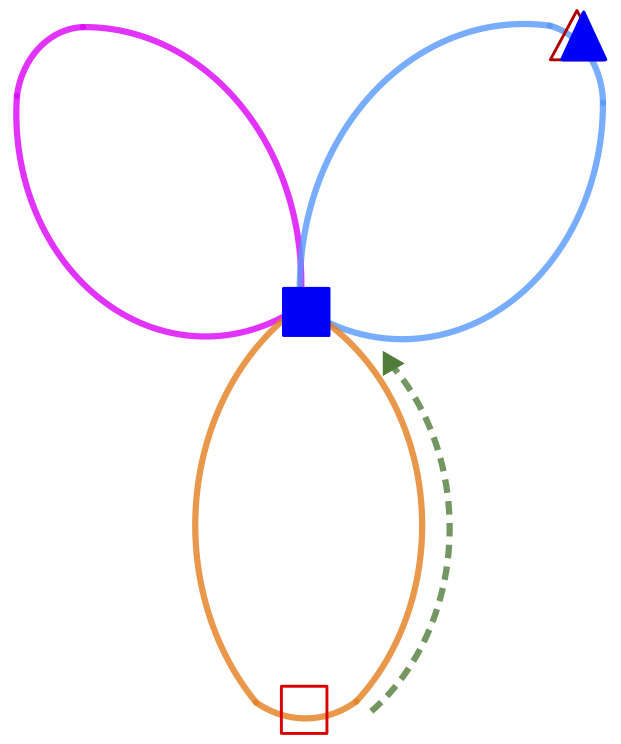}
\end{figure}
\end{center}

\begin{center}
\begin{figure}[H]
\caption{Third Step in $N$ and $C$: $ {CN}_f \rightarrow P_f^*$}
\centering
\includegraphics[height=6cm]{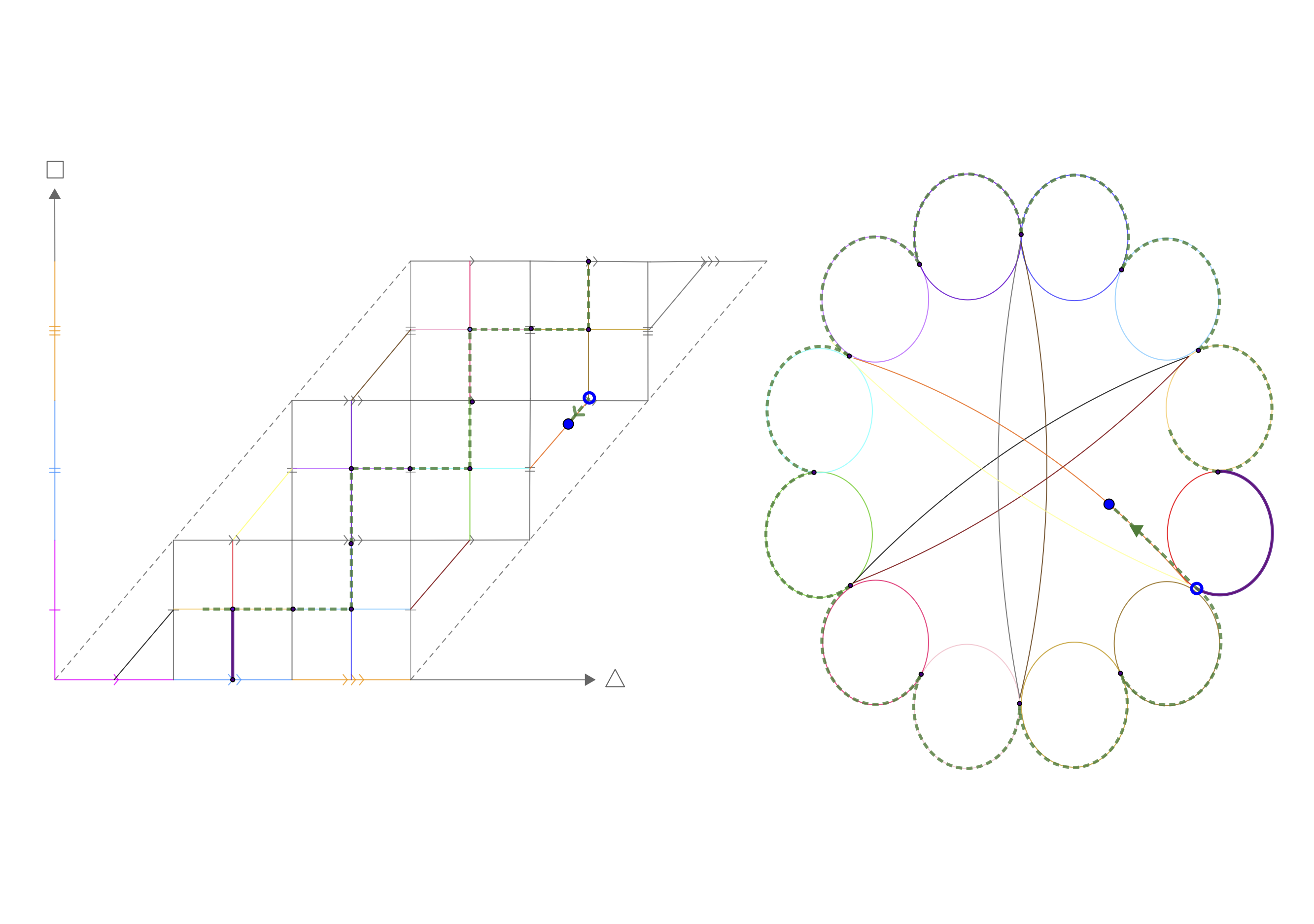}
\end{figure}
\end{center}

\begin{center}
\begin{figure}[H]
\caption{Third Step in $\Gamma$}

\centering
\includegraphics[height=5cm]{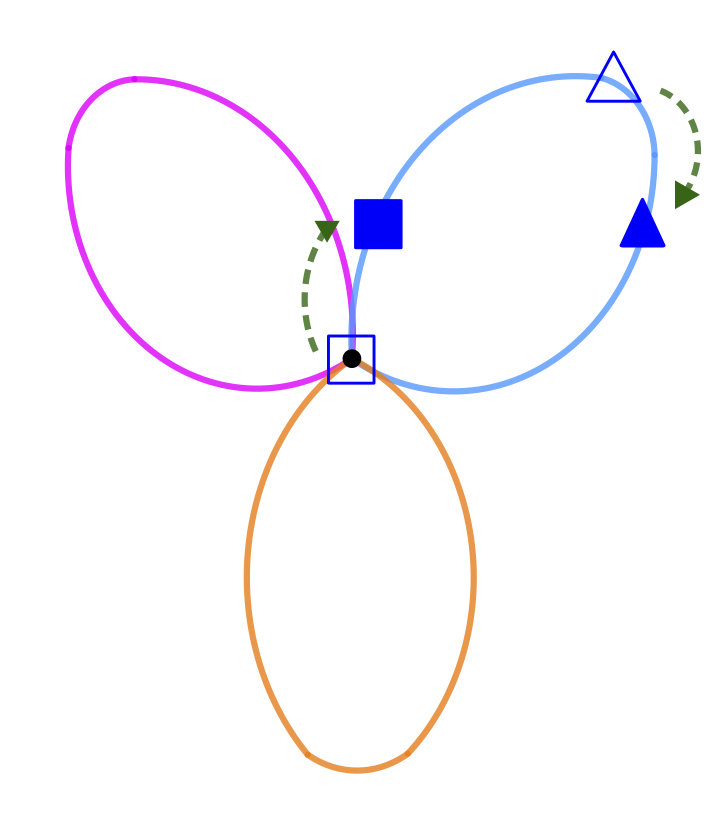}
\end{figure}
\end{center}

\begin{figure}[H]
\caption{Final Step in $N$ and $C$: ${CN}_f\rightarrow P_{f}^*$}
\centering
\includegraphics[height=6cm]{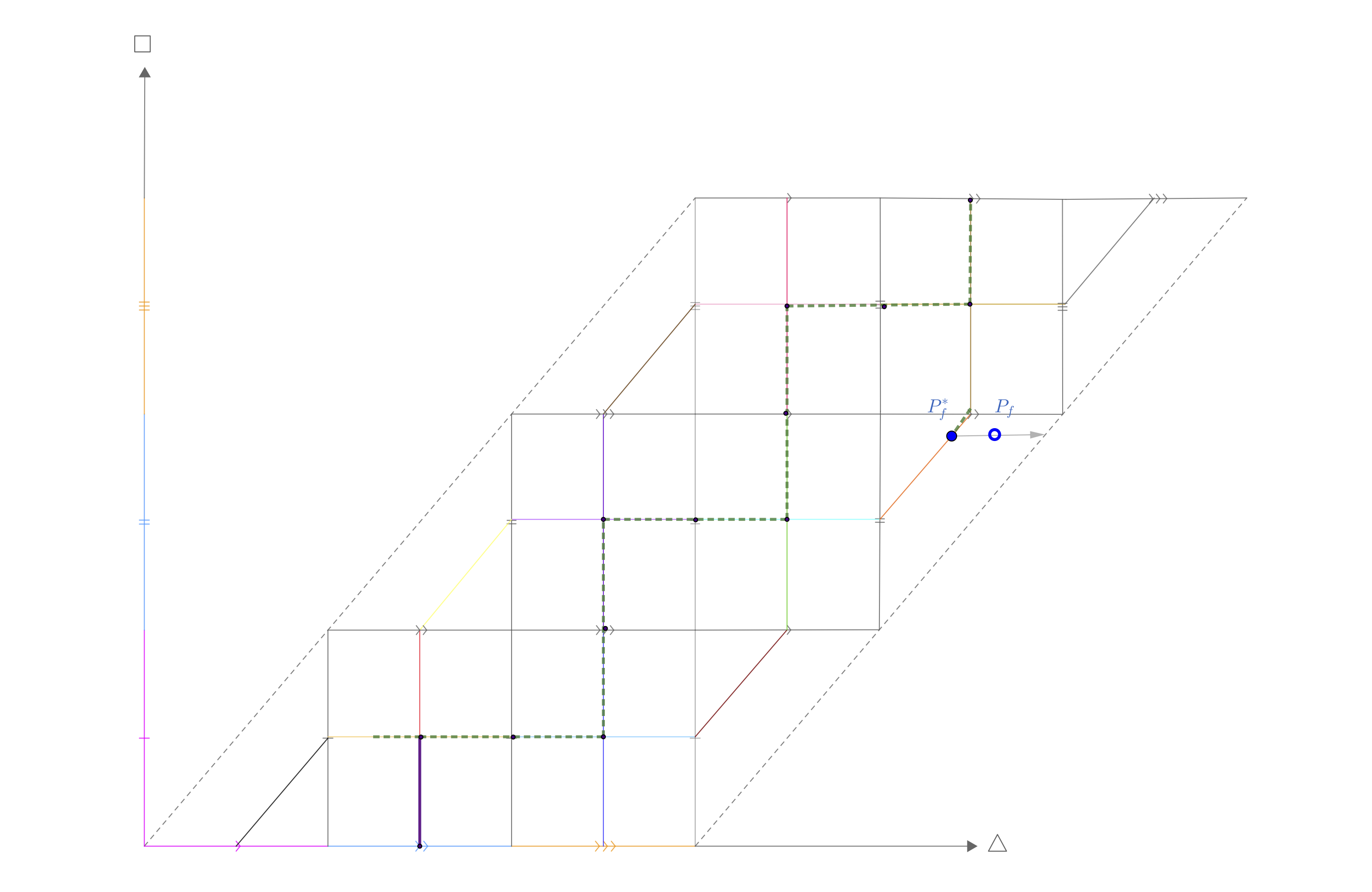}
\end{figure}

\begin{figure}[H]
\caption{Final Step in $\Gamma$}
\centering
\includegraphics[height=5cm]{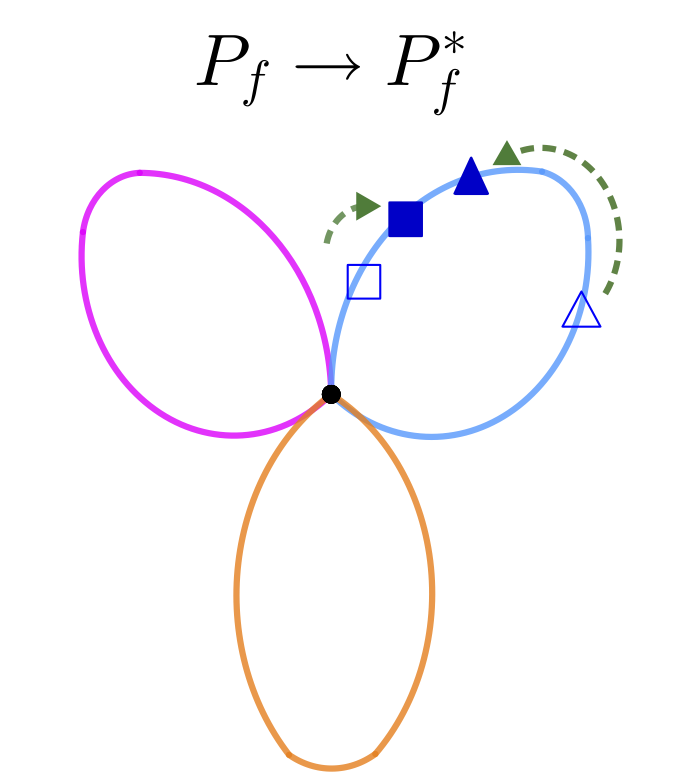}
\end{figure}


\begin{thebibliography}{}

\bibitem{survey}{M. Grant, G. Lupton and L. Vandembroucq}
\emph{Topological Complexity and Related Topics}, Contemp. Math., Amer. Math. Soc., vol. 702, Providence, RI, 2018. 



\bibitem{Farber1} {M. Farber.} \emph{Topological complexity of motion
planning}, {Discrete Comput. Geom.} \textbf{29} (2003),
211--221.

\bibitem{Farber2} M. Farber,  \emph{Instabilities of robot motion}, Topology Appl. \textbf{140} (2004), 245--266.


\bibitem{Ghrist} R. Ghrist, \emph{Configuration spaces and braid groups on graphs in robotics}, in: \emph{Knots, braids, and mapping class groups--papers dedicated to Joan S. Birman (New York, 1998)}, pp. 29--40, AMS/IP Stud. Adv. Math., vol. 24, Amer. Math. Soc., Providence, 2001.


\end{thebibliography}
\end{document}